\documentclass[12pt]{article}

\usepackage[utf8]{inputenc} % allow utf-8 input
\usepackage[T1]{fontenc}    % use 8-bit T1 fonts      % hyperlinks
\usepackage{url}            % simple URL typesetting
\usepackage{booktabs}       % professional-quality tables
\usepackage{amsfonts}       % blackboard math symbols
\usepackage{nicefrac}       % compact symbols for 1/2, etc.
\usepackage{microtype}      % microtypography
\usepackage{xcolor}         % colors

\usepackage{amsmath, amssymb} 
\usepackage{graphics, graphicx} % graphics packages
\usepackage[margin=1in]{geometry} % more reasonable margins
\usepackage{booktabs} % prettier tables
\usepackage{units} % prettier fractions (?)
\usepackage{multicol}
\usepackage{bm}
\usepackage{mathtools}
\usepackage{natbib} 
\usepackage{tikz} % make graphics in latex
\usetikzlibrary{shapes,decorations}
\usetikzlibrary{arrows.meta}
\usepackage{pgfplots} % to make plots in latex
\pgfplotsset{compat=newest} % compatibility rihu GB
\usepackage{comment}
\usepackage{fancyhdr} % for changing headers and footers
\usepackage[normalem]{ulem} % underline end emphasize
\usepackage[T1]{fontenc} 
\usepackage{comment}
\usepackage[shortlabels]{enumitem}
\usepackage[linesnumbered,algoruled,lined]{algorithm2e}
\usepackage[overload]{empheq}
\usepackage{wrapfig}
\definecolor{mydarkblue}{rgb}{0,0.08,0.45}
\usepackage[colorlinks, citecolor=blue, linkcolor=blue]{hyperref}

\newtheorem{definition}{Definition}
\newtheorem{lemma}{Lemma}
\newtheorem{condition}{Condition}
\newtheorem{example}{Example}
\newtheorem{theorem}{Theorem}
\newtheorem{assumption}{Assumption}
\newtheorem{remark}{Remark}

\providecommand{\keywords}[1]
{
  \small	
  \textbf{\textbf{Keywords:}} #1
}

\newcommand{\one}{\ensuremath{\mathbf{1}}}

\newcommand{\setX}{\mathcal{X}}

\newcommand{\setS}{\mathcal{S}}

\newcommand{\setP}{\mathcal{P}}
\newcommand{\setI}{\mathcal{I}}

\newcommand{\bA}{\mathbf{A}}
\newcommand{\bB}{\mathbf{B}}
\newcommand{\bW}{\mathbf{W}}
\newcommand{\bI}{\mathbf{I}}

\DeclareMathOperator*{\argmin}{arg\,min}

\tikzstyle{obsvar}=[draw,circle,orange, text=black, minimum size=6mm]
\tikzstyle{source}=[draw, circle, blue!75, text=black, minimum size=6mm]
\tikzstyle{real}=[draw, circle, text=black, minimum size=8.1mm]
\tikzstyle{real_appendix}=[draw, circle, text=black, minimum size=10mm]

\tikzset{causal/.style={-{Triangle[length=2.5pt, width=3pt]}, line width=1pt},
         exogenous/.style={-{Triangle[length=2.5pt, width=3pt]}, line width=1pt, blue!75},
         difference/.style={-{Triangle[length=2.5pt, width=3pt]}, line width=1pt, red}
}

\title{\textbf{Causal Discovery in Linear Structural Causal Models with Deterministic Relations}}
\usepackage{times}
\author{\normalsize \textbf{Yuqin Yang}\thanks{Correspondence to: \href{mailto:yuqinyang@gatech.edu}{\texttt{yuqinyang@gatech.edu}}. Accepted at 1st Conference on Causal Learning and Reasoning (CLeaR 2022).} 
\\ 
\normalsize Georgia Institute of
\normalsize Technology
\and  
\normalsize
\textbf{Mohamed Nafea} %\thanks{Electrical \& Computer Engineering Department, University of Detroit Mercy.} 
\\ 
\normalsize University of Detroit Mercy
\and
\normalsize
\textbf{AmirEmad Ghassami} %\thanks{Department of Computer Science, Johns Hopkins University.} 
\\ 
\normalsize Johns Hopkins University
\and 
\normalsize
\textbf{Negar Kiyavash} %\thanks{College of Management of Technology, École Polytechnique Fédérale de Lausanne (EPFL).}
\\ \normalsize École Polytechnique Fédérale \\
\normalsize de Lausanne (EPFL)
}

\date{}
\begin{document}

\maketitle

\begin{abstract}%
% \noindent
Linear structural causal models (SCMs)-- in which each observed variable is generated by a subset of the other observed variables as well as a subset of the exogenous sources-- are pervasive in causal inference and casual discovery. However, for the task of causal discovery, existing work almost exclusively focus on the submodel where each observed variable is associated with a distinct source with non-zero variance. This results in the restriction that no observed variable can deterministically depend on other observed variables or latent confounders. In this paper, we extend the results on structure learning by focusing on a subclass of linear SCMs which do not have this property, i.e., models in which observed variables can be causally affected by any subset of the sources, and are allowed to be a deterministic function of other observed variables or latent confounders. This allows for a more realistic modeling of influence or information propagation in systems. We focus on the task of causal discovery form observational data generated from a member of this subclass. We derive a set of necessary and sufficient conditions for unique identifiability of the causal structure. To the best of our knowledge, this is the first work that gives identifiability results for causal discovery under both latent confounding and deterministic relationships. Further, we propose an algorithm for recovering the underlying causal structure when the aforementioned conditions are satisfied. We validate our theoretical results both on synthetic and real datasets. 
\end{abstract}

% \begin{keywords}%
% \end{keywords}

\vspace{-3mm}
\section{Introduction} \label{sec:intro}
\vspace{-2mm}
Causal discovery, which refers to the problem of learning causal relationships among the variables of a system, has received extensive attention with applications ranging from social sciences, economics, all the way to biology. The gold standard of causal discovery is performing interventions (i.e., controlled experiments). But such experiments could be costly, infeasible, or even at times unethical. This necessitates developing statistical methods based on purely observational data, where further assumptions on the data generating process are needed.

\setlength\parskip{5pt}

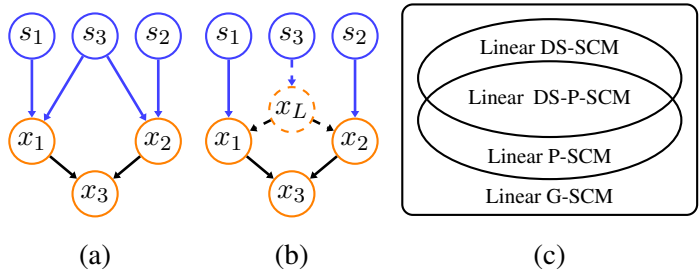
\begin{wrapfigure}{r}{0.62\textwidth}
\centering
\begin{tikzpicture}[thick, scale=.28]
\centering
\foreach \place/\name in {{(-3,1)/x_1}, {(3,1)/x_2}, {(0,-1.5)/x_3}}
    \node[obsvar, label=center:$\name$] (\name) at \place {};
  \foreach \source/\dest in {x_1/x_3, x_2/x_3}
    \path[causal] (\source) edge (\dest);

\foreach \place/\name in {{(-3,6)/s_1}, {(3,6)/s_2}, {(0,6)/s_3}}
    \node[source, label=center:$\name$] (\name) at \place {};
  \foreach \source/\dest in {s_1/x_1, s_2/x_2, s_3/x_1, s_3/x_2}%, s_3/x_3
    \path[exogenous] (\source) edge (\dest);
\node at (0,-4.5) {(a)};
\end{tikzpicture}
\hspace{1mm}
\begin{tikzpicture}[thick, scale=.28]
\centering
\node[obsvar, dashed, label=center:$x_L$] (x_L) at (0,2.5) {};

\foreach \place/\name in {{(-3,1)/x_1}, {(3,1)/x_2}, {(0,-1.5)/x_3}}
    \node[obsvar, label=center:$\name$] (\name) at \place {};
  \foreach \source/\dest in {x_1/x_3, x_2/x_3}
    \path[causal] (\source) edge (\dest);
  \foreach \source/\dest in {x_L/x_1, x_L/x_2}
    \path[causal] (\source) edge[dashed] (\dest);

\foreach \place/\name in {{(-3,6)/s_1}, {(3,6)/s_2}, {(0,6)/s_3}}
    \node[source, label=center:$\name$] (\name) at \place {};
  \foreach \source/\dest in {s_1/x_1, s_2/x_2}%, s_3/x_3
    \path[exogenous] (\source) edge (\dest);
  \foreach \source/\dest in {s_3/x_L}
    \path[exogenous] (\source) edge[dashed] (\dest);
\node at (0,-4.5) {(b)};
\end{tikzpicture}
\hspace{1mm}
\begin{tikzpicture}[thick, scale=.28]
\centering
\tikzstyle{every node}=[font=\small]
\draw[rounded corners] (-7, -5) rectangle (7, 5) {};
\draw (0,1.5) ellipse (6.25cm and 2.8cm);
\draw (0,-0.5) ellipse (6.25cm and 2.8cm);
\node at (0,-4.1) {\scriptsize Linear G-SCM};
\node at (0,3) {\scriptsize Linear DS-SCM};
\node[text width=3cm, text centered] at (0,0.6) {\scriptsize Linear DS-P-SCM}; %\scriptsize
\node at (0,-2.2) {\scriptsize Linear P-SCM};
\node at (0,-7) {(c)};
\end{tikzpicture} 
\caption{
(a) A linear P-SCM with observed variables $x_1$, $x_2$, $x_3$ and independent sources $s_1,s_2,s_3$.
This model cannot be represented as a linear DS-SCM, since $x_3$ is fully determined by $(x_1, x_2)$. (b) An interpretation of the P-SCM in (a), where $x_1$, $x_2$ are both influenced by a latent confounder $x_L$.
(c) Relations among linear causal models. The intersection DS-P-SCM represents the linear causal models under distinct source assumption, jointly independent sources and linear latent confounding. \vspace{-3mm}
%(Left) A linear SCM with a latent confounder, $x_L$. Here, $x_3$ is fully determined by $(x_1, x_2, x_L)$. This model cannot be represented as a linear SCM when all exogenous noises have non-zero variance. (Right) An equivalent linear P-SCM: The latent confounder $x_L$ (now presented as $s_3$), and the exogenous noises $n_1$ and $n_2$ (now presented as $s_1$ and $s_2$) are all considered as sources.
}
\label{fig:example_SCM_fails}
\end{wrapfigure}
Linear structural causal models (SCMs) have been extensively considered in the literature perhaps as the most pervasive causal data generating model \citep{pearl2009causality,spirtes2000causation,peters2017elements}. In this model, the system is comprised of a set of observed (endogenous) variables and a set of source (exogenous) variables. Each observed variable $x$ is generated as a linear combination of a subset of the other observed variables $Pa(x)$ (called direct causes of $x$) plus a function of a subset of the source variables $S(x)$. We refer to this model as the linear general SCM (G-SCM).

Literature on linear G-SCM for causal discovery almost exclusively focuses on a subset of G-SCM, in which for each observed variable $x$, at least one source in $S(x)$ is distinct and has non-zero variance \citep{peters2017elements}. Here, a distinct source of observed variable $x$ means a source $s\in S(x)$ that does not belong to any other subsets $S(x')$, $x'\neq x$. % \footnote{A more common representation of this subclass is that each observed variable $x$ is only associated with one source, and the source of each observed variable is not a deterministic function of other sources.} 
Then the non-distinct (i.e., shared) sources can be explained as {\it{latent confounders}} in the system.
% part of the source variable which does not belong to any other source
We refer to this submodel of G-SCM as \emph{distinct source SCM} (DS-SCM). This restriction implies that no observed variable can be a deterministic function of any other observed variables and/or any latent confounders in the system. % other observed variables
However, in a G-SCM, an observed variable is allowed to have no distinct sources  (or equivalently can have a distinct source with zero-variance). Hence, deterministic relations are allowed. Deterministic relations are problematic in causal discovery as they introduce additional dependencies among variables \citep{peters2017elements}. Specifically, when deterministic relations exist, d-separation is not complete anymore (although it remains sound). Such complications are perhaps the reason that deterministic relations are avoided for the most part in the literature.
% deterministic relations lead to complications such as the changes they cause in the concept of d-separation (that is, d-separation is still sound but will not be complete when deterministic relations exist), hence they are only considered in a few works []. 

%Therefore, they have been only considered in a few works, where almost all developments in causal discovery avoid deterministic relations to avoid complications caused by them such as the changes they cause in the concept of d-separation (that is, d-separation is still sound but will not be complete when deterministic relations exist).

Furthermore, G-SCMs without distinct sources can have practical significance in problems beyond causal modelling. This subset of G-SCMs can be used to model influence propagation through networks, where the observed nodes are influenced directly or indirectly by a set of external factors (sources). Examples of such factors include sources of news or rumors, or sources of diseases.
%-- an ``external factor'' can be either a latent confounder or an exogenous source associated with one observed variable. 
Specifically, G-SCMs without distinct sources can be used to model scenarios with less factors than observed nodes; 
% this situation can happen when one or more observed variables deterministically depend on other observed variables in the system. For instance, in a setting of spread of news or rumors, an individual may have their news exclusively from a certain set of news sources \citep{guille2013information, quinn2015directed}. 
this situation arises when a few sources influence a group of observed variables in the system. For instance, in a setting of spread of news, a large group of individuals may have their news exclusively from a limited set of news sources \citep{guille2013information, quinn2015directed}. 
Other examples include spread of epidemics \citep{nowzari2016analysis}, political campaigns \citep{gonzalez2016networked}, message passing in human brain networks \citep{karwowski2019application}.

% why people dont work with the rest of G-SCM
In this paper, we focus on another subclass of linear G-SCMs which relaxes the distinct source assumption of DS-SCM and hence, allows for deterministic relations among the observed variables. In our proposed model, we consider a set of jointly independent source variables. Each observed variable is generated as a linear combination of the other observed variables plus a linear combination of the sources.
 We relax the distinct source assumption of DS-SCM by assuming instead that each observed variable contains strictly more sources than any of its direct causes. Figure \ref{fig:example_SCM_fails}(a) depicts an example which satisfies our assumption but violates the assumption of DS-SCM.
%We also assume that each observed variable contains strictly more sources than any of its direct causes, which relaxes the distinct source assumption of DS-SCM. 
As previously mentioned, sharing sources introduces latent confounding among the variables. Hence our introduced model allows for both latent confounders as well as deterministic relations. We refer to this model as the linear \emph{propagation SCM} (P-SCM). 
Majority of works in the literature on linear causal models in fact consider an intersection of P-SCM and DS-SCM, %in which all three assumptions of linear latent confounding, jointly independent sources (not necessarily independent source mixtures), and Assumption \ref{assumption:distinct_source} hold (i.e., $\mathcal{M}_4$ in Figure \ref{fig:example_SCM_fails})
which we refer to as DS-P-SCM.
Therefore, our work strictly expands the considered model space compared to those works. See Figure \ref{fig:example_SCM_fails}(c) for a schematic representation of the relations among the linear models. 

% correlation among sources is caused by not observing the full model
%In this paper, we focus on another subclass of linear G-SCMs which relaxed the distinct source assumption of DS-SCM and hence, allows for deterministic relations among the variables.
%Specifically, we consider the subclass of linear G-SCMs which satisfies the assumptions that (i) any correlation among sources can be explained by linear latent confounding (or equivalently, the sources are jointly independent), and (ii) any observed variable contains strictly more sources than any of its direct causes. We refer to this model as the linear propagation SCM (P-SCM). 
%Note that condition (i) above is not required by DS-SCM, but condition (ii) relaxed the requirement of DS-SCM. Therefore, in general, P-SCM is neither a subset of a superset of DS-SCM.
%(ii) above is a relxes the of the requirement in DS-SCM. 
% Under the aforementioned assumptions, linear P-SCM is strictly larger than linear DS-SCM since it allows for deterministic relations. Figure \ref{fig:example_SCM_fails} shows an example which cannot be explained by a linear DS-SCM, yet can be modeled as a linear P-SCM. 

We study the problem of causal discovery when the data generating model is a linear P-SCM, that is, to learn the directions and strengths of the causal influences among the observed variables, as well as the direct connections from the sources to the observed variables. We derive a set of necessary and sufficient conditions for unique identifiability of the causal structure from observational data under the assumptions of faithfulness and separability of the sources.
% By ``causal structure'' here we refer to the directions and strengths of  causal influences among the observed variables, and the direct connections from the exogenous sources to the observed variables.
To the best of our knowledge, this is the first work that gives identifiability results for causal discovery under both latent confounding and deterministic relationships.
We further propose an algorithm which uniquely identifies the causal structure when the conditions are satisfied. Also, we derive an equivalent condition when the generating model is a linear DS-P-SCM.
We examine our algorithm on synthetic and real datasets, and compare with existing causal discovery methods developed for linear DS-SCMs with or without latent confounders. The results show that our proposed algorithm outperforms those methods in recovering the underlying structure.%class of linear P-SCMs.

{\bf{Related work.}} Under causal sufficiency, \citet{shimizu2006linear} showed that a linear non-Gaussian additive model (LiNGAM), in which source variables have non-Gaussian distributions, allows for unique identifiability of the causal structure from observations. Causal sufficiency is a common assumption in causal discovery which means there exist no latent confounders for any pair of observed variables. Yet, without considering the effect of latent confounders, one may infer wrong causal relations among observed variables. \citet{hoyer2008estimation} and \citet{salehkaleybar2020learning} considered an extension of LiNGAM in the presence of latent variables (lvLiNGAM). Our model generalizes lvLiNGAM to the case when observed variables are allowed to deterministically depend on other observed variables or latent confounders. We further comment on the connection of our model to lvLiNGAM in Appendix \ref{sec:application_to_our_model}. 

SCMs with deterministic relations have been considered in a few work. In \citep{geiger1990identifying, spirtes2000causation}, D-separation condition (with capital letter D)
% \footnote{This is different from the classical d-separation (with small letter d) described in \citep[Section 1.2]{pearl2009causality}.} 
is proposed for graphically determining conditional independencies when deterministic relations are allowed. Yet, it remained unclear whether D-separation condition can capture {\it{all}} conditional independencies induced from the distribution. \citet{daniusis2012inferring} and \citet{janzing2012information} considered a system with only two variables, one variable deterministically causing the other, and showed that the correct causal direction can be learnt if the deterministic function contains no information about the cause variable and vice versa.
Their analysis does not hold for linear relations though. \citet{scheines1996tetrad} considered recovering a reduced model in which all deterministic variables are removed. \citet{luo2006learning, mabrouk2014efficient, lemeire2012conservative} adapted conventional causal discovery methods such as PC and greedy-search algorithms \citep{pearl2009causality, tsamardinos2006max,chickering2002optimal}, where deterministic relations are detected either by additional independence tests, or by calculating conditional entropy among variables.
The aforementioned methods suffer from identifiability problem. 
% {\color {red}
% In particular, without prior assumption on the distribution of the source variables, the underlying structure can only be identified up to certain {\it{equivalence class}}.
% They do not derive any identifiability result.
% }
In particular, majority of these works do not discuss the capability of identifying the underlying structure by their proposed algorithms (e.g., the equivalence classes of the recovery result).
% In particular, under the assumption of gaussian source variables, the best that can be identified is the underlying structure under certain equivalence classes
% they only allow for estimating the underlying causal structure from data up to a certain {\it{equivalence class}}.
Besides, these methods do not consider the presence of latent confounders in the system.
Please refer to Appendix \ref{app:related_work} for a detailed background on causal discovery methods.
%Due to these reasons, we do not compare our proposed algorithm for unique identifiability with these methods. 

%the use of word ''all'', ''the''
% Independent components in Section 3 vs Unique components in Section 5
%%$$  w_{i,j}.  W[i,j]

\vspace{-5mm}
\section{System model}
\label{sec:system_model}
\vspace{-3mm}
{\it Notation.} We use upper-case letters for vectors and bold upper-case letters for matrices. We use $[n]\triangleq \{1,2,\cdots,n\}$. If $X=[x_1,\cdots,x_p]$, $X_{[i:j]}$ is the sub-vector $[x_i,x_{i+1},\cdots,x_j]$; $1\leq i < j \leq p$. We use $x_i \rightsquigarrow x_j$ to refer to a directed path from node $x_i$ to node $x_j$ (in the corresponding graph). 
\vspace{-4mm}
\subsection{Generating model}\label{Generating Model} \vspace{-2mm}
As we mentioned in the introduction, linear G-SCM is one of the main data generating models assumed in the literature. A G-SCM \citep[Definition 7.1.1]{pearl2009causality} is a 4-tuple $\langle \setS, \setX, \mathcal{F}, P(\setS)\rangle$. $\setS$ is a set of source variables, $\setX$ is a set of observed variables, $\mathcal{F}=\{f_x\}_{x\in \setX}$ is a set of functions such that for each $x\in \setX$, $x=f_x(Pa(x),S(x))$, $Pa(x)\subseteq \setX \setminus\{x\}$ and $S(x)\subseteq \setS$, and $P(\setS)$ is a joint probability distribution over $\setS$.  In the case of linear functions, the model is restricted as follows:

\vspace{-2mm}
\begin{definition}[Linear G-SCM]
A linear G-SCM is the subclass of G-SCM in which each observed variable $x$ is generated 
as $x=f_x(Pa(x))+g_x(S(x))$, where $f_x$ is a linear function and $g_x$ is a function indicating the effects from the sources. We refer to $g_x(S(x))$ as the source mixture corresponding to observed variable $x$.
\vspace{-2mm}
\end{definition}

%{\color{red}  the causal effect between observed variables are assumed to be linear,  

Previous works on causal discovery mostly focus on a subclass of linear G-SCM
which satisfies the distinct source assumption defined below.
\vspace{-2mm}
\begin{assumption}[Distinct source assumption]
\label{assumption:distinct_source}
A linear G-SCM satisfies the distinct source assumption if each $S(x)$ contains at least one distinct source, that is, a source with non-zero variance that does not appear in any other $S(x')$ for $x'\neq x$.
% A linear G-SCM satisfies the distinct source assumption if for each observed variable $x$, there exists a source such that the information regarding that source in source mixture of $x$ does not fully exist in the source mixtures of the rest of the variables.
\end{assumption}
\vspace{-1mm}
We refer to the subclass of linear G-SCMs satisfying Assumption \ref{assumption:distinct_source} as linear DS-SCM.
The non-distinct (i.e., shared) sources represent latent confounders in the system. 
%}
Assumption \ref{assumption:distinct_source} can be interpreted as:
% A: for any variable $x$
% there exists a source such that the information regarding that source in source mixture of $x$ does not fully exist in the source mixtures of the rest of the variables.
% explaining of what will happen with 
The source mixture of one variable cannot be a deterministic function of the components in the source mixtures of the rest of the variables.\footnote{
In fact, Assumption \ref{assumption:distinct_source} is stronger than what is needed for structure learning methods developed for linear DS-SCMs. We provide a weaker version of Assumption \ref{assumption:distinct_source} in Appendix \ref{app:distinct_source}, under which these methods still work.}
% Note that for a linear G-SCM, whether each observed variable has distinct sources depends on the set of functions $\{g_x\}_{x\in \setX}$. Please refer to Appendix \ref{app:distinct_source} for an example.
% Note that even if $\setS$, $\setX$ and the causal structure of a linear G-SCM are given, it is still unclear whether each observed variable has distinct sources, depending on the functions $\{g_x\}$. Please refer to Appendix \ref{app:distinct_source} for an example.
Equivalently, no observed variable can be a deterministic function of any other observed variables and/or any latent confounders in the system.

In this paper, we introduce another subclass of linear G-SCMs, which we refer to as the linear propagation SCM (P-SCM). Consider a set of observed variables $\setX=\{x_1,\cdots,x_p\}$ arranged in a causal order (no latter variable causes any earlier variable), and another set of jointly independent source variables $\setS=\{s_1,\cdots,s_m\}$. In the generating model of a linear P-SCM, each observed variable consists of a linear combination of earlier observed variables plus another linear combination of the sources. Specifically,
\vspace{-2mm}
\begin{align}
\label{eq:system_model_1}
x_i &= {{\sum}}_{j=1}^{i-1}\;a_{ij} x_j + \tilde{s}_i,\qquad \tilde{s}_i=g_{x_i}(S(x_i))={{\sum}}_{j=1}^m b_{ij}s_j.
\end{align}
$a_{ij}$ is the strength of the direct causal effect from $x_j$ to $x_i$. $\tilde{s}_i$, $i\in[p]$ represents the mixture of sources exogenously causing $x_i$. $b_{ij}$ is the strength of the direct exogenous effect from $s_j$ to $x_i$. We refer to $\{\tilde{s}_1,\cdots, \tilde{s}_p\}$ as {\it{source mixtures}} or simply {\it{mixtures}}. %Note that the linear combination for $\tilde{s}_i$ in  \eqref{eq:system_model_1} leads to linear latent confounding. 
Note that since \eqref{eq:system_model_1} allows observed variables to share sources, our model allows for latent confounding.\footnote{Note that prohibiting latent confounders, i.e., causal sufficiency, requires independent source mixtures.} However, the latent confounding can only be linear, i.e., it is due to the source mixtures being linear combinations of elements in $\setS$.
In our model we require the following assumption on the generating model which is weaker than Assumption \ref{assumption:distinct_source} of DS-SCM.
\vspace{-2mm}
\begin{assumption} \label{assumption:P-SCM_connection}
If $x_i$ is a direct cause of $x_j$, then there is at least one source in $x_j$ that is not in $x_i$.
\end{assumption}
\vspace{-1mm}
Unlike the distinct source assumption in DS-SCMs, Assumption \ref{assumption:P-SCM_connection} does not necessarily require each observed variable to be associated with a distinct source (see Figure \ref{fig:example_SCM_fails}(a) for an example). Therefore, our model allows for deterministic relations.\footnote{If the linear P-SCM only consists of two observed variables $x_i, x_j$, then Assumption \ref{assumption:P-SCM_connection} implies that $x_j$ cannot be a deterministic function of $x_i$ and shared sources. This example is however a degenerate case. In particular, given a deterministic relation between $x_i$ and $x_j$, it is not possible to recover their causal direction based on observations.} We provide a detailed comparison of our model with DS-SCM in Section \ref{sec:comparison}.

% {\color {red} For the linear P-SCM in \eqref{eq:system_model_1}, correlation among the source mixtures is explained by linear latent confounding, which is not necessarily the case for linear G-SCMs. Hence linear P-SCMs are a subclass of linear G-SCMs. However, a linear P-SCM allows for deterministic relations, since there is no restriction on the linear combinations that generates the source mixtures. %That is, if we consider the restriction that each source mixture $\tilde{s}_i$ has a distinct source (with non-zero variance), then no deterministic relations are allowed.
% } 

We define the causal diagram of the model as a directed graph in which the vertex set consists of the observed variables $\mathcal{X}$ and the source variables $\mathcal{S}$. There is a directed edge from observed variable $x_i$ to $x_j$ if and only if $a_{ji}\neq 0$ (we refer to such an edge as a causal connection), and a directed edge from source $s_i$ to observed variable $x_j$ if and only if $b_{ji}\neq 0$ (we refer to such an edge as an exogenous connection).
Due to the assumed causal order, the causal diagram is acyclic, and hence the model is a directed acyclic graph (DAG).
% See Figure \ref{fig:causal_diagram} for an example of the causal diagram.
The model in \eqref{eq:system_model_1} can be written in the matrix form as 
\vspace{-2mm}
\begin{align}
\label{eq:system_model_2}
X=\bA X+  \tilde{S};\qquad \tilde{S} =\bB S, \vspace{-2mm}
\end{align}
where $X\triangleq[x_1,\cdots,x_p]^\top$, $\tilde{S}\triangleq [\tilde{s}_1,\cdots,\tilde{s}_p]^\top$, $S \triangleq [s_1,\cdots,s_m]^\top$. $\bA$ is a $p\times p$ strictly lower triangular matrix with the coefficient $a_{ij}$ on the $(i,j)$-th entry. $\bB$ is a $p\times m$ matrix with the coefficient $ b_{ij}$ on the $(i,j)$-th entry. The strictly lower triangular matrix $\bA$ attests the existence of a causal order among the observed variables (the generating process is recursive \citep{bollen1989structural}). Mixtures $\tilde{s}_i$, $i\in[p]$ represent {\it{additive exogenous noises}}. 
% We refer to the model in (\ref{eq:system_model_2}) as the linear propagation SCM (P-SCM).
\vspace{-2mm}
\subsection{Identifiability assumptions} \label{sec:model_assumptions} \vspace{-2mm}
In \eqref{eq:system_model_2}, each $x_i$ can be written as a linear combination of the sources.
% $\{s_1,\cdots,s_m\}$.
Let $\mathbf{W}$ be a mixing matrix s.t.
\vspace{-2mm}
\begin{align}
\label{eq:system_model_4}
X=\bW S,\qquad \text{where}\quad \bW = (\bI - \bA)^{-1} \bB,\vspace{-2mm}
\end{align}
Define the component set of a variable $x_i=\sum_{j} w_{ij} s_j$ (or a mixture $\tilde{s}_i=\sum_{j} b_{ij} s_j$) as
$ {\rm{Comp}}(x_i) = \left\{s_j\in \mathcal{S}~| w_{ij}\neq 0 \;(\text{or } b_{ij}\neq 0)\right\}.$
%In addition to the description in \eqref{eq:system_model_1}, the data generating process possesses the following properties:
We require the following assumptions for identifiability.
% \vspace{-1mm}
\begin{assumption}[P-SCM faithfulness] \label{assumption:faithfulness}
% \vspace{-3mm}
\begin{enumerate}[(a)]
\itemsep0em
\item If $x_i$ is an ancestor of $x_j$, then all source components in $x_i$ must also appear in $x_j$. 
\vspace{-1mm}
\item Let $\bB_p$ be an arbitrarily permuted version of matrix $\bB$ (either row or column permutations). Any submatrix of $\bB_p$ (with non-zero rows or columns) is of full rank.
\end{enumerate}
% \vspace{-2mm}
\end{assumption}
\vspace{-1mm}
%{\textbf{1. P-SCM faithfulness:}}

% Conditions (a) and (b) can be combined into a single condition: $x_i \rightsquigarrow x_j \Longrightarrow {{\rm{Comp}}}(x_i) \subsetneq {\rm{Comp}}(x_j)$. \sout{Conditions (a) and (c) are satisfied almost surely if all model coefficients $\{a_{ij}\}$ and $\{b_{ij}\}$ are drawn randomly and independently from continuous distributions, see} \citep{meek2013strong}.

\vspace{-3mm}
\begin{assumption}[Separability]\label{assumption:separability}
We assume the linear P-SCM is separable-- its corresponding mixing matrix $\bf{W}$, cf. \eqref{eq:system_model_4}, can be correctly recovered using observations $X$, up to scaling and permutation of its columns.
\end{assumption}
\vspace{-2mm}

P-SCM faithfulness assumption is extended from the common faithfulness assumption in studying causal models \citep{spirtes2000causation,pearl2009causality}. This assumption is satisfied {\it{almost surely}} if all model coefficients $\{a_{ij}\}$ and $\{b_{ij}\}$ are drawn randomly and independently from continuous distributions \citep{meek2013strong}. %[2.3]Condition (b) is relaxed from the distinct source assumption in a linear DS-SCM.} 
Separability, on the other hand, can be achieved using blind source separation (BSS) methods \citep{comon2010handbook}, given certain assumption on the sources. For example, when the sources are non-Gaussian random variables, the mixing matrix $\bf{W}$ can be recovered using Independent Component Analysis (ICA) or overcomplete ICA methods \citep{comon1994independent, hyvarinen2002independent, lewicki2000learning}. Other BSS methods include Statistical Blind Source Separation Regression (SBSSR) model and Non-negative Matrix Factorization (NMF).
See Appendices \ref{app:assumptions} and \ref{app:BSS} for detailed explanation about both assumptions, and how BSS methods are applied to our model.
Note that the result of BSS may suffer from permutation and scalability indeterminacies, due to lack of prior information about the sources. In Appendix \ref{sec:application_to_our_model} we show that these indeterminacies do not affect the recovery performance of our algorithm. Specifically, we show that, even in the presence of these indeterminacies, our recovered model is identical to the true model. 
% \vspace{-4mm}

\section{Conditions for unique identifiability} \vspace{-2mm} \label{sec:definition_conditions}
% In this section, we first define the ``possible parent set'' for each observed variable as a superset of its ancestors which can be recovered from the mixing matrix (from observational data). This set is needed to explain the necessary and sufficient conditions for unique identifiability, which consists of two parts, the unique components condition and marriage condition discussed in Sections \ref{sec:unique_component}, \ref{sec:marriage_condition}. 
In this section, we present the necessary and sufficient conditions for unique identifiability of a linear P-SCM, which consists of two parts, the unique components condition and marriage condition. Both conditions are imposed on the ``possible parent set'' of each observed variable which is a superset of its ancestors that can be recovered from the mixing matrix (from observational data).
We first define the possible parent set, and then present the combinatorial version of the conditions. Equivalent algebraic (matrix) representations of the conditions are provided in Appendix \ref{Matrix_representation}.

%Unique components are the source components in the ancestors of an observed variable $x$ that are not shared by other ancestors, and hence can be used to compute total causal effects to $x$. The marriage condition handles the cases when some ancestors do not have unique components.
%\noindent {\bf{Possible parent sets.}}
%\label{sec:possible_parent_sets}\
%In earlier methods such as LiNGAM, the learning algorithm relies on deducing the causal order by permuting the recovered mixing matrix. In contrast, 
The necessary and sufficient condition we derive for unique identifiability of a linear P-SCM, as well as our proposed algorithm, require a search over the {\it{possible parent set}} of an observed variable $x_k$ which is defined as follows.
%a certain set that includes the ancestors of a given observed variable $x_k$. We refer to this minimal set as the ``possible parent set'' of $x_k$, and define it as follows: 

\vspace{-2mm}
\begin{definition}\label{def:pp_set} 
The \emph{possible parent set} of an observed variable $x_k$ is defined as the set of observed variables in $\setX$ whose component set is a strict subset of the component set of $x_k$. That is,
$\mathcal{P}_k =\{ x_i\in \setX | {\rm{Comp}}(x_i)\subsetneq {\rm{Comp}}(x_k)\}$.
\end{definition}
\vspace{-2mm}

We refer to the elements of this set as the ``possible parents'' of $x_k$ since each element is a candidate to be a parent of $x_k$, according to P-SCM faithfulness assumption. The possible parent set of $x_k$ (i) can be deduced from the mixing matrix and (ii) can be strictly smaller than the set of observed variables preceding $x_k$ in the causal order; see Appendix \ref{app:possible_parent_set_example} for an example.

In linear DS-SCMs, since each observed variable has a distinct source, $\mathcal{P}_k$ only includes the ancestors of $x_k$ in the generating model (see Section \ref{sec:simplify} for more explanation).
This property allows to identify all the observable ancestors of each variable $x_k$ from the mixing matrix. Subsequently, an algorithm, such as lvLiNGAM in \citep{salehkaleybar2020learning}, can recover the directed causal paths among variables. By contrast, in our linear P-SCM, the possible parent set $\mathcal{P}_k$ can include observed variables which are not ancestors of $x_k$. The reason behind this is that there may exist an observed variable which does not have a distinct source, hence this variable can mimic an ancestor of $x_k$ by its exogenous connections to the sources. 

From \eqref{eq:system_model_4}, the observed variables are given by $X\!=\!(\mathbf{I}\!-\! \bA)^{-1}\bB S \!=\! (\mathbf{I}\!-\! \bA)^{-1} \tilde{\bB} S'$, where $\tilde{\bB}$ is a version of $\bB$ with the columns arbitrarily permuted and rescaled such that $\tilde{\bB} S'=\bB S$. Note that the $(i,j)$-th entry of $(\bI-\bA)^{-1}$ represents the {\it{total causal effect}} from observed variable $x_j$ to observed variable $x_i$ \citep{spirtes2000causation}.
Our task is to uniquely recover $(\bI-\bA)^{-1}$, i.e., the total causal effects among observed variables, and to recover $\bB$ up to permutation and scaling of its columns, i.e., the exogenous connections from the sources to the observed variables.
Note that the permutation indeterminacy of $\bB$ is immaterial since there exists no ordering between the exogenous connections, and we are not concerned about the scalability indeterminacy in recovering their effect. Next, we introduce the necessary and sufficient conditions for recovering $(\bI-\bA)^{-1}$ and $\tilde{\bB}$. 

\vspace{-4mm}
\subsection{Unique components condition} \label{sec:unique_component} \vspace{-1mm}
% observed variables in $\mathcal{P}_k$
When recovering the total causal effect from possible parents of $x_k$ (i.e., variables in $\setP_k$) to $x_k$, the question is whether each existing component in $x_k$ (i.e., $s\in \bigcup_{x_i\in \mathcal{P}_k} {\rm{Comp}}(x_i)$) comes from a possible parent ($x_i\in\mathcal{P}_k$), or comes from the exogenous connections to $x_k$ (i.e., $\tilde{s}_k=\sum_{l=1}^m\; b_{kl} s_l$), or maybe both. To address this question, we propose the concept of {\it{unique components}}. For a possible parent $x_i\in\mathcal{P}_k$, 
if $\tilde{s}_i$ has a component that is not shared by the source mixture of any other possible parent in $\mathcal{P}_k$ (i.e., any $\tilde{s}_j$; $j\neq i$, $x_j\in\mathcal{P}_k$), then we can use this {\it{unique component}} to compute the total causal effect from $x_i$ to $x_k$ (by dividing the coefficient of this unique component in $x_k$ by its coefficient in $x_i$, see Remark \ref{remark:sufficiency_calculation} in the sufficiency proof for more details). 

%The aforementioned description of unique components involve all observed variables in $\mathcal{P}_k$. 
%We can improve the recovery of the total causal effects 
We can recursively use the unique components idea to find the total causal effects from all possible parents in $\mathcal{P}_k$ to $x_k$: After using the unique components to recover total causal effects from the corresponding possible parents to $x_k$, we remove those variables and search for unique components with respect to the remainder subset of $\mathcal{P}_k$. Thus, we extend the concept of unique components to subsets of $\mathcal{P}_k$ by considering only the possible parents in these subsets. We define the unique components over the mixtures $\tilde{\setS}_k=\{\tilde{s}_i: x_i\in \mathcal{P}_k\}$ in an iterative manner as follows.
\begin{definition} \label{def:unique}
%Fix an observed variable $x_k$. Let $U_k(i)$ denote the unique component set of observed variable $x_i$ in the possible parent set $\mathcal{P}_k$. We define $U_k(i)$ for each $x_i\in\mathcal{P}_k$ as follows:\\
Fix an observed variable $x_k$. For each possible parent of $x_k$, i.e., $x_i\in\mathcal{P}_k$, we define the \emph{unique component set} of $x_i$, denoted by $U_k(i)$, as follows.\\
(i) Let $J^{(0)}=\{i: x_i\in \mathcal{P}_k \}$, i.e., the index set of the possible parent set of $x_k$.\\
(ii) For $n=1,2,\cdots$ ($n$ is the iteration index): \\
\hspace*{4pt} - For $i\in J^{(n-1)}$, define 
$\tilde{U}_{k}^{(n)}(i) = {\rm{Comp}}\left(\tilde{s}_{i}\right) \big\backslash \bigcup_{j \in J^{(n-1)}, j \neq i} {\rm{Comp}}\left(\tilde{s}_{j}\right).$ \\
% \end{align}
\hspace*{4pt} - Let $J^{(n)} = \{i\in J^{(n-1)} |~ \tilde{U}^{(n)}_{k}(i)=\emptyset \}$. \\
\hspace*{4pt} - Repeat until $|J^{(N+1)}|=|J^{(N)}|$ for some $N\in\mathbb{N}$. Define $\mathcal{I}_k=\{x_l:  l\in J^{(N)}\}$.\\
%{\color {red}{Denote the set of observed variables with indices in $J^{(N)}$ by $\mathcal{I}_k$.}}\\
For each $x_i\in \mathcal{P}_k$, $U_k(i) = \tilde{U}_{k}^{(n)}(i),~ i \in J^{(n-1)} \backslash J^{(n)},~n = 1,\cdots, N$; $U_k(i) = \emptyset,~i \in J^{(N)}$.
\end{definition}
\vspace{-2mm}
In Definition \ref{def:unique}, when $n=1$, $\tilde{U}_{k}^{(1)}(i)$ represents the components in $\tilde{s}_i$ but not in any other $\tilde{s}_j$, where $x_j\in\mathcal{P}_k$ (or equivalently $j\in J^{(0)}$). $J^{(1)}$ represents the index set of $\mathcal{P}_k^{(1)}$ which is the subset of $\mathcal{P}_k$ composed of observed variables with no unique components in the first iteration. In the next iteration, $\tilde{U}_k^{(2)}(i)$ is defined similarly but only for $i\in J^{(1)}$. $J^{(2)}$ is the index set of the subset of $\mathcal{P}_k^{(1)}$ with no unique components. This procedure is repeated until no more unique components are found; $\mathcal{I}_k \subseteq \mathcal{P}_k$ is the remaining subset of possible parents with no unique components after the procedure terminates. Appendix \ref{app:example_unique_component} provides an example to demonstrate this iterative definition. 

We use Definition \ref{def:unique} as follows. We (i) identify the possible parents in $\mathcal{P}_k$ with unique components, (ii) compute their total causal effects to $x_k$ using these unique components, and (iii) subtract the learnt causal effects from $x_k$. Next, we consider only the possible parents with no unique components in the first iteration ($x_i\in\mathcal{P}_k^{(1)}$), and find which of these have unique components among variables in $\mathcal{P}_k^{(1)}$. We then use these unique components to compute total causal effects from the corresponding variables in $\mathcal{P}_k^{(1)}$ to $x_k$. Based on Definition \ref{def:unique}, we state the unique components condition as follows.

%Definition \ref{def:unique} can be used as follows. We first identify the observed variables in $\mathcal{P}_k$ which have unique components and use these unique components to compute the total causal effects from the corresponding observed variables to $x_k$. Subsequently, we consider only the remaining subset of $\mathcal{P}_k$ composed of observed variables with no unique components, denoted by $\mathcal{X}_{k}^{\mathcal{J}}$. For each $x_i \in \mathcal{X}_{k}^{\mathcal{J}}$, we find whether $\tilde{s}_i$ has a unique component that is not shared by any other mixture $\tilde{s}_j$, where $x_j\in \mathcal{X}_{k}^{\mathcal{J}}$. Since the total causal effects from the observed variables in  $\mathcal{P}_k\setminus\mathcal{X}_{k}^{\mathcal{J}}$ are computed in the first iteration, we can use the unique components within $\mathcal{X}_{k}^{\mathcal{J}}$ to compute the total causal effects from the corresponding observed variables in $\mathcal{X}_{k}^{\mathcal{J}}$ to $x_k$. This procedure can be repeated until no further unique components are found. In Appendix \ref{app:example_unique_component}, we provide an example to demonstrate the iterative definition of unique components. Using Definition \ref{def:unique}, we state the unique components condition as:
\vspace{-2mm}
\begin{condition}[Unique components condition]\label{condition:unique_component}
(i) For each possible parent $x_i$ of $x_k$ (i.e., $x_i \in \mathcal{P}_k)$ with a non-empty unique component set (i.e., $|U_{k}(i)|\neq 0$), the unique components of $x_i$ are not exogenously connected to $x_k$. That is, $ {\rm{Comp}}(\tilde{s}_k) \cap  U_k(i) =\emptyset,\; \forall x_i\in \mathcal{P}_k \text{ s.t. } U_k (i) \neq \emptyset.$\\ (ii) Further, for any $x_i\in \mathcal{P}_k$ with an empty unique component set ($|U_{k}(i)|= 0$), the exogenous connections to $x_i$ and $x_k$ are disjoint. That is,  ${\rm{Comp}}(\tilde{s}_k) \cap  {\rm{Comp}}(\tilde{s}_{i}) = \emptyset,\;  \forall x_i\in \mathcal{P}_k \text{ s.t. } U_k (i) = \emptyset.$
\end{condition}
\vspace{-2mm}
Condition \ref{condition:unique_component} restricts certain exogenous connections from sources to observed variables in the ground-truth structure. The first part of Condition \ref{condition:unique_component} states that if a source $s$ is a unique component of any possible parent of $x_k$, then $s$ should not have an exogenous connection to $x_k$. The second part states that any possible parent of $x_k$ with no unique components should not overlap in any exogenous connections with $x_k$.\footnote{When an observed variable has multiple unique components, we only need one unique component not to be exogenously connected to $x_k$ to infer the total causal effect. However, there is no prior information about which of these unique components is not exogenously connected to $x_k$, hence none of the unique components can be exogenously connected to $x_k$. For more details, see the necessity proof in Appendix \ref{app:proof_necessity}.}
In general, 
%the condition requires that an observed variable 
$x_k$ can only be exogenously connected to (i) source components that are shared only among its possible parents that have unique component(s), (ii) new source components that are not connected to any of its possible parents. Please refer to Example \ref{example:unique_components} in Appendix \ref{app:example_conditions} to see why this condition is necessary for unique identifiability.

\begin{algorithm}[t]
\SetAlgoVlined
\LinesNumbered
% \DontPrintSemicolon
\KwIn{Recovered mixing matrix $\tilde{\bW}$ from BSS.\quad  \textbf{Initialize:} $\tilde{\bA} = \mathbf{I}$, $\bB=\mathbf{0}$.} 
Repermute $\tilde{\bW}$ such that the number of non-zero entries in each row is in an increasing order (rows with equal number of non-zero entries are permuted at random) \label{alg:permute}\;
\For{$k=1:p$}{
Find possible parent set $\mathcal{P}_k$ using $\tilde{\bW}$; Select the row $W_k = \tilde{\bW}[k,:]$; Initialize $\bar{\mathcal{P}}_k=\mathcal{P}_k$ \;
Find the set $\mathcal{U}$ of possible parents in $\bar{\mathcal{P}}_k$ that have unique components \label{alg:find_unique}\;
\eIf{$|\mathcal{U}| \neq 0$}{
	Compute the total causal effect $\tilde{a}_{ki}$ for each $x_i\in \mathcal{U}$ using its unique components \label{alg:total_effect_unique} \;
	% Remove the causal effect of $\tilde{s}_i$ from $x_k$:
	$W_k \leftarrow W_k - \sum_{x_i\in \mathcal{U}} \tilde{a}_{ki} \bB[i,:]$;~ $\bar{\mathcal{P}}_k \leftarrow \bar{\mathcal{P}}_k \setminus \mathcal{U}$;~ 
	% Remove the possible parents in $U$ from possible parent set:  $. 
	Go back to Step \ref{alg:find_unique} using updated $\bar{\mathcal{P}}_k$ \label{alg:subtraction} \;
}{Select the rows of $\bB$ corresponding to the possible parents in $\bar{\mathcal{P}}_k$, denote by $\bB_I$\; 
Compute $\tilde{a}_{ki}$ by solving an overdetermined linear system, using the non-zero columns of $\bB_I$ and the corresponding entries in $W_k$; Set the selected entries of $W_k$ as 0 \label{alg:total_effect_nonunique}\;
}
$\bB[k,:] = W_k$\;
}
$\bA = \mathbf{I} - \tilde{\bA}^{-1}$\;
\KwOut{Repermuted matrices $\bA$, $\bB$ according to the reversed order from Step \ref{alg:permute}.}
\caption{P-SCM Recovery} 
\label{alg:recovery} 
\end{algorithm}

% \vspace{-3mm}
\subsection{The marriage condition} \label{sec:marriage_condition} 
\vspace{-1mm}
The second part of the necessary and sufficient conditions for unique identifiability of a linear P-SCM is identical to the marriage condition of Hall's marriage theorem \citep{cameron1994combinatorics}. The combinatorial formulation of this theorem considers a collection of finite sets and provides a necessary and sufficient condition (the marriage condition) for the feasibility of selecting a distinct element from each set.  
\begin{lemma}[Hall's marriage theorem and marriage condition] \label{Lemma : Marriage Theorem}
Let $X$ be a collection of finite subsets of a set $S$. The members of $X$ can be counted with multiplicity (two or more members are identical). Define the mapping $f(x): X\mapsto S$ s.t. $f$ selects a distinct element from each set $x\in X$. The collection $X$ satisfies the marriage condition if for every sub-collection $C \subseteq X$,
$|C| \leq |\bigcup_{x\in C}  x |$, i.e., every $C \subseteq X$ covers at least $|C|$ different elements of $S$. Hall's Marriage Theorem states that, the mapping $f$ exists if and only if $X$ satisfies the marriage condition.
\end{lemma}
In our framework, the collection $X$ corresponds to the family of component sets of possible parents of a variable $x_k$. Specifically, members of the collection $X$ are the finite subsets ${\rm{Comp}}(x_i)\subseteq \setS$,  $x_i\in\mathcal{P}_k$, $\setS$ is the set of sources. 
The marriage condition translates to the following:
\begin{condition}[The marriage condition] \label{condition:marriage}
For each observed variable $x_k$ with possible parent set $\mathcal{P}_k$, every subset of the possible parent set, $X_C \subseteq \mathcal{P}_k$, satisfies $|X_C| \leq \big| \bigcup_{x_i\in X_C} {\rm{Comp}}(x_i) \big|$. That is, every subset $X_C\subseteq \mathcal{P}_k$ includes at least $|X_C|$ different source variables. 
\end{condition}
{Condition \ref{condition:marriage} implies that the possible parent set of every observed variable must have a sufficient number of exogenous connections (from source variables) so that it allows for unique identifiability of the causal effects to this observed variable.} Since each variable $x_k$ can be written as a linear combination of the source mixtures in its possible parent set ($\tilde{\setS}_k=\{\tilde{s}_i: x_i\in \mathcal{P}_k\}$) plus $\tilde{s}_k$, Condition \ref{condition:marriage} can be equivalently imposed on the collection $\tilde{\setS}_k$:
%That is, the marriage condition is equivalent to: 
For $\tilde{S}_C \subseteq \tilde{\setS}_k$, $|\tilde{S}_C| \leq \left| \bigcup_{\tilde{s}_i\in \tilde{S}_C} {\rm{Comp}}(\tilde{s}_i) \right|$. 
See Appendix \ref{app:proof_marriage} for the proof. 
Recall that Condition \ref{condition:unique_component} is also imposed on $\tilde{\mathcal{S}}_k$ (the unique components are defined over the source mixtures in $\mathcal{P}_k$). Please refer to Example \ref{ex:marriage_condition} in Appendix \ref{app:example_conditions} to see the necessity of the marriage condition for unique identifiability of a linear P-SCM.

% {\color {red}Notice that we presented a combinatorial version of the conditions in this section. Equivalent algebraic (matrix) representation of the conditions is provided in Appendix \ref{Matrix_representation}.  In particular, both conditions are imposed on the submatrix $\bB$ that corresponds to the rows of the possible parent sets, while the marriage condition can be equivalently imposed on the submatrix of $\bW$.}

\vspace{-2mm}
\subsection{Main result} \label{sec:main_result}
\vspace{-2mm}
Our main result states the necessary and sufficient conditions for unique identifiability of the total causal effects among the observed variables and the exogenous connections from the sources.
% \subsection{Main Theorem} \label{sec:main_result}
\vspace{-1mm}
\begin{theorem} \label{thm:main}
Let $\mathcal{A}=\{a_{ij},b_{ik}: i,j\in\{1,\cdots,p\}, k\in\{1,\cdots,m\}\}$. $\mathcal{A}$ fully parameterizes a linear P-SCM. Let $\pi$ denote the Lebesgue measure over $\mathcal{A}$. The causal effects and exogenous connections of a linear P-SCM are uniquely identifiable from observations {\it{almost surely}} (w.r.t. the measure $\pi$) if and only if, for any observed variable $x_k$, both Conditions \ref{condition:unique_component} and \ref{condition:marriage} are satisfied. 
%\begin{enumerate}[(1)]
%\item For each possible parent $x_i$ of $x_k$ (i.e., $x_i \in \mathcal{P}_k)$ with a non-empty unique component set (i.e., $|U_{k}(i)|\neq 0$), the unique components of $x_i$ are not exogenously connected to $x_k$. That is:\\  $ {\rm{Comp}}(\tilde{s}_k) \cap  U_k(i) =\emptyset,\; \forall x_i\in \mathcal{P}_k \text{ s.t. } U_k (i) \neq \emptyset.$
%\begin{align}
%\label{eq:UC_condition_1}
%& {\rm{Comp}}(\tilde{s}_k) \cap  U_k(i) =\emptyset,\qquad \forall x_i\in \mathcal{P}_k \text{ s.t. } U_k (i) \neq \emptyset.
%\end{align}

%Further, for any $x_i\in \mathcal{P}_k$ with an empty unique component set, $|U_{k}(i)|= 0$, the exogenous connections to $x_i$ and $x_k$ are disjoint:  ${\rm{Comp}}(\tilde{s}_k) \cap  {\rm{Comp}}(\tilde{s}_{i}) = \emptyset,\;  \forall x_i\in \mathcal{P}_k \text{ s.t. } U_k (i) = \emptyset.$
%\begin{align}
%\label{eq:UC_condition_2}
%& {\rm{Comp}}(\tilde{s}_k)  \cap  {\rm{Comp}}(\tilde{s}_{i}) = \emptyset,\qquad  \forall x_i\in \mathcal{P}_k \text{ s.t. } U_k (i) = \emptyset.
%\end{align}

%\item The collection of component sets of the possible parents of $x_k$, i.e., $\{{\rm{Comp}}(x_i)\}_{x_i\in \mathcal{P}_k}$, satisfies marriage condition: For any subset $X_C \subseteq \mathcal{P}_k$, $|X_C| \leq \big| \bigcup_{x_i\in X_C} {\rm{Comp}}(x_i) \big|$.
%\end{align}$
%\begin{align}
%label{eq:Marriage_condition}
%|C| \leq \Big| \bigcup_{x_i\in C} %{\rm{Comp}}(x_i) \Big|.
%\end{align}
%\end{enumerate}
\end{theorem}
\vspace{-1mm}

The proof of Theorem \ref{thm:main} appears in Appendix \ref{app:proof_main_theorem}. Necessity of the conditions is proved using similar reasoning as in Examples \ref{example:unique_components} and \ref{ex:marriage_condition}: If either condition is not satisfied, it is impossible to distinguish the true generating model from certain different generating model(s).
For sufficiency of the conditions, we propose the P-SCM Recovery algorithm (Algorithm \ref{alg:recovery}) and prove its correctness, i.e., the algorithm recovers the true causal structure when the conditions are satisfied. For each observed variable $x_k$ with possible parent set $\mathcal{P}_k$, Algorithm \ref{alg:recovery} computes the total causal effects from $x_i\in \mathcal{P}_k$ to $x_k$ using the unique components in $U_k(i)$ in the same iterative manner as described in Definition \ref{def:unique}, until no more unique components can be found in the last subset $\mathcal{I}_k$. The algorithm computes the total causal effect from all $x_i\in \mathcal{I}_k$ to $x_k$ by solving an overdetermined linear system. 
%Here, the constraints are the sources exogenously connected to $x_i$, for all $x_i\in \mathcal{I}_k$. 
After computing total causal effects from all possible parents to $x_k$, the remaining parts in $x_k$ are considered as exogenous connections.

% , since these sources cannot be exogenously connected to $x_k$ according to the condition

% Given the recovered mixing matrix $\tilde{\bW}$, the algorithm first reveals the correct causal order by row permutations. Then for each observed variable $x_k$ with the possible parent set $\mathcal{P}_k$ deduced from $\tilde{\bW}$, the algorithm iteratively calculates the total causal effect using the unique components in $U_k(i)$. For the last subset $\mathcal{I}_k$ where no possible parent have unique components, solve for the overdetermined system to find the causal effect. The remaining parts in the row of $\tilde{\bW}$ corresponding to $x_k$ are the direct connections $\tilde{s}_k$.
%\section{Reduction of our main results to linear DS-SCMs}

\vspace{-3mm}
\section{Comparison with linear DS-SCM} \label{sec:comparison} \vspace{-2mm}
There are two main differences between our proposed linear P-SCM and linear DS-SCM. The first difference regards the type of latent confounders they allow. From \eqref{eq:system_model_1}, in linear P-SCM, the source mixture associated with each observed variable is comprised of a linear combination of jointly independent sources, which can lead to {\it{linear}} latent confounding. That is, the dependency between two source mixtures is due to the fact that they are both linear combinations of the same set of sources.
%explained by a linear function of some latent confounders. 
This is not necessarily the case for linear DS-SCM, i.e., the source mixtures can be nonlinear combinations of the sources.
%where the correlation among the source mixtures is not explained by a linear relation. 
The second difference between P-SCM and DS-SCM regards the restriction of distinct sources. Linear P-SCM relaxes the distinct source assumption to the requirement in Assumption \ref{assumption:P-SCM_connection} which assumes that each observed variable contains strictly more sources than any of its direct causes. This means that unlike DS-SCM, in our model, an observed variable can be a deterministic function of its direct causes and the latent confounders.

Note that all the works in the literature on linear DS-SCM that we are aware of-- with the exception of \citep{wang2020causal}-- in fact consider an intersection of P-SCM and DS-SCM, in which all three assumptions of linear latent confounding, jointly independent sources (not necessarily independent source mixtures), and Assumption \ref{assumption:distinct_source} hold \citep{hoyer2008estimation,entner2010discovering,tashiro2014parcelingam,salehkaleybar2020learning}. In this case, the subclass of P-SCMs representing this subclass are those with matrix $\bB$ in \eqref{eq:system_model_2} of the form $\bB =[\bI \;\; \bB_{s}]$. We refer to this subclass of linear G-SCM as linear DS-P-SCM. Therefore, our work strictly expands the considered model space compared to those works. %In Appendix \ref{app:proof_equiv_nondeter}, we show that the submodel of linear P-SCM satisfying distinct source assumption is equivalent to the submodel of linear DS-SCM under linear latent confounding and jointly independent sources. 
% In Section \ref{sec:simplify} we will discuss how our conditions can be reduced for the submodels for linear DS-P-SCM.
To demonstrate the practical significance of linear P-SCM over linear DS-P-SCM in causal discovery, we show that there are models that can be correctly explained by a linear P-SCM, while linear DS-P-SCMs either are inapplicable or give misleading results (hence the model cannot be recovered using causal discovery methods for linear DS-P-SCMs). These examples appear in Appendix \ref{app:example_equivalence}.

\vspace{-3mm}
\subsection{Reduction of the conditions for linear DS-P-SCM} \label{sec:simplify}
\vspace{-2mm}
To show how our conditions for unique identifiability of linear P-SCMs can be reduced for linear DS-P-SCMs, consider the setting in which each observed variable is associated with a distinct source. Under this setting, the model can be equivalently written as a linear DS-SCM with linear latent confounding and jointly independent sources (see Appendix \ref{app:proof_equiv_nondeter} for more details).
This setting reduces Conditions \ref{condition:unique_component} and \ref{condition:marriage} in Theorem \ref{thm:main} as follows. First, all possible parents of an observed variable $x_k$ have unique components, i.e., $U_k(i)\neq \emptyset$ for all $x_i\in\mathcal{P}_k$. This further implies that the marriage condition is automatically satisfied. Second, for each $x_k$, the possible parent set $\mathcal{P}_k$ is equivalent to the ancestor set of $x_k$. This follows because, for each $x_i\in \mathcal{P}_k$ with a distinct source $s_j$, since $s_j$ is not connected to any other observed variables including $x_k$, the $s_j$ component in $x_k$ must result from $x_i$. Hence, there must be a causal connection/path from $x_i$ to $x_k$. 
% This property allows us to impose the reduced conditions on the generating model.
Theorem \ref{thm:simplify} below provides the equivalent necessary and sufficient conditions for this setting. See Appendix \ref{app:proof_simplify} for the proof, as well as a graphical representation of the condition. 
% Note that in this setting, the marriage condition described in Theorem \ref{thm:main} is automatically satisfied.
\vspace{-1mm}
\begin{theorem} \label{thm:simplify}
Suppose every observed variable in the linear P-SCM is associated with a distinct source. The causal effects and exogenous connections of a linear P-SCM are uniquely identifiable from observations almost surely (in the sense described in Theorem \ref{thm:main}) if and only if the following condition holds: For a source $s_j$ that is exogenously connected to an observed variable $x_k$, either there are no causal paths from any observed variable containing $s_j$ to $x_k$, or there are at least two distinct causal paths from the observed variable(s) with $s_j$ in their source mixtures to $x_k$.
\end{theorem}
\vspace{-2mm}
% In the following, we state a few remarks about Theorem \ref{thm:simplify}.

\begin{figure}[t]
% \begin{wrapfigure}{r}{0.6\textwidth}
\centering
\begin{tikzpicture}[scale=.3]
\centering
\node at (-19.5,11) { };
% \tikzstyle{every node}=[font=\small]
\draw[rounded corners] (-18, -12) rectangle (24, 13) {};
% \node at (0,16.5) {\Large Linear P-SCM}; %outside
\node at (16.5,11) {\large Linear P-SCM}; % inside

\draw (-3,0.7) ellipse (13.5cm and 11.2cm);
% \node at (-3,13.5) {\Large Linear DS-P-SCM}; % outside
\node at (-3,9.6) {Linear DS-P-SCM}; % inside

\draw (-2,-0.07) ellipse (11cm and 8.7cm); 
% \node[text width=5cm, text centered] at (-3,10) {Uniquely identifiable in Linear DS-P-SCM}; % outside
\node[text width=4cm, text centered] at (-2, 5) {\small Uniquely identifiable in linear DS-P-SCM}; % inside
\draw[dashed] (-0.9,-2.6) ellipse (8.5cm and 5.2cm); % Saber's result

\draw[fill=gray!30] (-0.2,-3) ellipse (7cm and 4.2cm); % Uniquely identifiable as a -P_SCM
\node[text width=4cm, text centered] at (-0.2,-2.8) {\small Uniquely identifiable in linear P-SCM (A)};

\draw[fill=gray!30] (16.5,-7) ellipse (7cm and 4cm); % Uniquely identifiable as a -P_SCM
\node[text width=4cm, text centered] at (16.5,-7) {\small Uniquely identifiable in linear P-SCM (B)}; %12.5,-11
\end{tikzpicture} 
\caption{Theoretical results derived in this paper. Dashed circle represent the condition in \citep{salehkaleybar2020learning}. Theorem \ref{thm:main} translates to: A linear P-SCM $\mathcal{G}\in A\cup B \Leftrightarrow \mathcal{G}$ satisfies Conditions \ref{condition:unique_component} and \ref{condition:marriage}. Theorem \ref{thm:simplify} translates to: A linear DS-P-SCM $\mathcal{G}\in A \Leftrightarrow \mathcal{G}$ satisfies the condition in Theorem \ref{thm:simplify}.}
\label{fig:summary}
\end{figure}
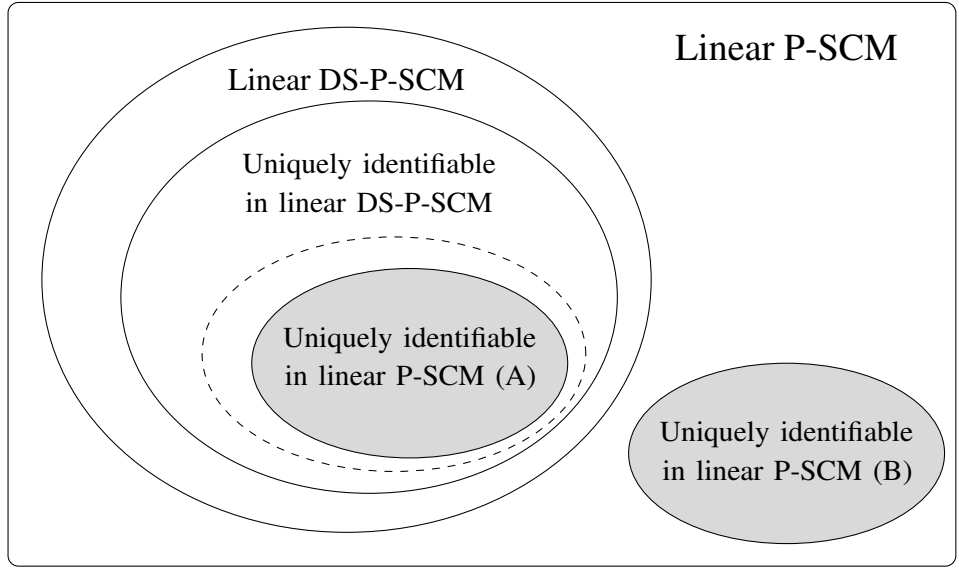

\vspace{-2mm}
\begin{remark} \label{remark:relation_to_amirs_work}
The condition in Theorem \ref{thm:simplify} for a linear P-SCM with distinct source implies the condition for unique identifiability of a linear DS-P-SCM in \citep[Theorem 16]{salehkaleybar2020learning}, but the reverse is not true. Thus, a linear DS-P-SCM can be uniquely identifiable in \citep{salehkaleybar2020learning} but not uniquely identifiable (as a linear P-SCM) in our setting. The reason is that 
the search space in Theorem \ref{thm:simplify}, which is the whole class of linear P-SCMs, is strictly larger than the search space in \citep{salehkaleybar2020learning}, which is linear DS-P-SCMs. 
%Note that Theorem 2 reduces the class of generating models but not the search space used to derive the condition. 
% reduced condition in Theorem \ref{thm:simplify} remains to be for unique identifiability over the whole class of linear P-SCMs, which is a strictly larger class (search space) than linear SCMs. %Note that reducing the conditions in Theorem \ref{thm:simplify} does not reduce the search space
See Appendix \ref{app:amirs_work} for the proof, and Figure \ref{fig:summary} for a schematic representation of the conditions. 
\vspace{-1mm}
%, while the condition in Theorem \ref{thm:simplify} is reduced to the case when each observe is for unique identifiability over all linear P-SCMs, which is a strictly larger class the linear SCMs. 
%while the result in \citep{salehkaleybar2020learning} is over only linear SCMs, which is a strict subclass of linear P-SCMs.
\end{remark}

\begin{figure*}[t]
\centering
\begin{minipage}[c]{0.66\textwidth}
\includegraphics[width=\textwidth]{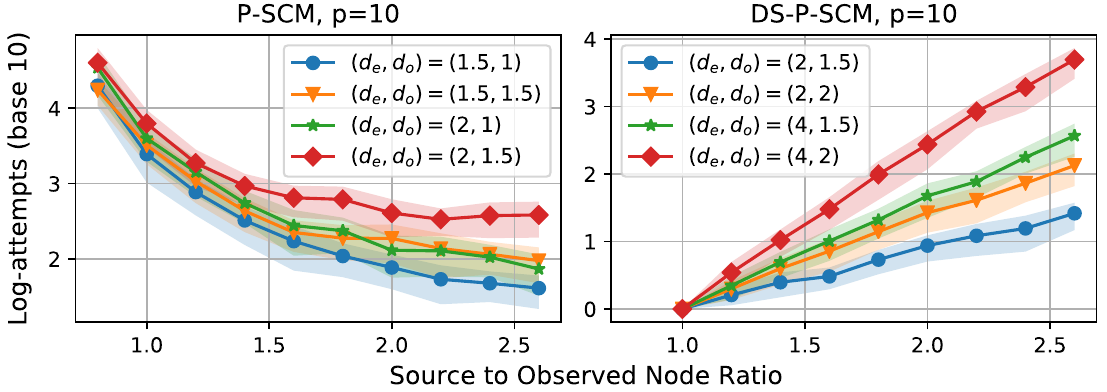}
\end{minipage}\hfill
\begin{minipage}[c]{0.32\textwidth}
\begin{tikzpicture}[thick, scale=0.4]
% \node at (-5,0) {}
% \tikzstyle{every node}=[font=\scriptsize]
\foreach \place/\name in {{(-3.5,0)/N100}, {(3.5,0)/N225}, {(-1.75,-4)/SSEC}}
    \node[real, label=center:{\scriptsize \name}] (\name) at \place {};
\foreach \place/\name in {{(1.75,-4)/HSI}, {(0,2.5)/DJI}}
    \node[real, label=center:{\small \name}] (\name) at \place {};
\foreach \source/\dest in {DJI/N100, DJI/SSEC, DJI/HSI, N100/SSEC, N100/HSI, N225/SSEC, N225/HSI}
    \path[causal] (\source) edge (\dest);
\end{tikzpicture}
\end{minipage}
\caption{(Left) Satisfiability of our conditions on models generated according to a linear P-SCM or DS-P-SCM.  (Right) Recovered causal graphs among five world stock indices.}
%The pair in each legend represents the expected number of causal connections $d_e$ and exogenous connections $d_o$.
\label{fig:nips2021_exp1}
\end{figure*}
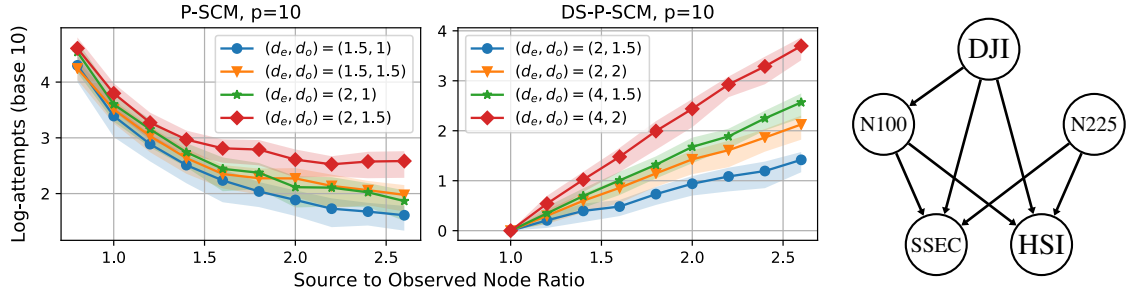

\vspace{-4mm}
\subsection{Choice of the representation of linear P-SCM}
\vspace{-2mm}
Note that equivalently we can say P-SCM also requires distinct sources but those sources can have zero variance (i.e., zero). In this case, as shown in Appendix \ref{app:proof_equiv_deter}, instead of the form in \eqref{eq:system_model_2}, our model can be equivalently written 
as 
\vspace{-3mm}
\begin{equation}\label{eq:ds_scm_deterministic}
\begin{aligned}
X=\bA_{ol}X_l+\bA X+S_d,
\end{aligned} 
\vspace{-3mm}
\end{equation}
where
$X_l$ is the vector of latent confounders; $S_d$ is the vector of (possibly zero) jointly independent distinct sources; $\bA_{ol}$ represents the latent confounding.
% ; $\bB_{oo}$ is a square matrix that indicates the assignment of distinct exogenous sources to non-deterministic variables\footnote{Note that we can remove $\bB_{oo}$ and allow for zero entries in $S_o$ (i.e., $X=\bA_{ol}X_l+\bA X+S$) but it does not matter in the following arguments.}.
% A_ol
This representation is perhaps more familiar to the reader (as it is considered in most works on linear DS-P-SCM), yet we claim that the representation in \eqref{eq:system_model_2} is more suitable for the task of identification when zero variance sources exist in the model:
%especially when deterministic relations and latent variables are present.
% First, by allowing zero variances, the DS-SCM can contain deterministic relationships not only among the observed variables, but also among latent variables and across latent and observed variables (in its canonical form) which to the best of our knowledge has not been studied before. All such relations can be easily coded into the design of the connections in a P-SCM. Second, P-SCM is more suitable for the task of identification, as we do not need to make the distinction between latent confounder and distinct sources. 
First, in order to fully learn the structure of the model in \eqref{eq:ds_scm_deterministic}, the number of latent confounders must be known in advance \citep{hoyer2008estimation,salehkaleybar2020learning}, which cannot be deduced from the mixing matrix $\bW$ when deterministic relations are present. In this case, we can only learn the sum of the number of latent confounders (i.e., cardinality of $X_l$) and non-zero distinct sources (i.e., support of $S_d$). On the contrary, by considering both latent confounders and non-zero distinct sources as $S$ in \eqref{eq:system_model_2}, we do not need to know the number of latent confounders or the number of non-zero distinct sources in the recovery. Second, our conditions for identifiability of linear P-SCM depend on the exogenous connection matrix $\bB$, which includes both latent confounding and non-deterministic relationships in \eqref{eq:ds_scm_deterministic} (i.e., matrix $\bA_{ol}$ and the non-zero entries in $S_d$). As a result, the identifiability results derived in this work would be more complicated if translated on the model in \eqref{eq:ds_scm_deterministic}, and we are not aware of a straightforward way to do this translation. %Note that our derived conditions are necessary and sufficient for unique identifiability. Therefore, all our conditions should be translated to a DS-SCM to obtain equivalently strong results.

\vspace{-4mm}
\section{Numerical experiments} \label{sec:simulations}
\vspace{-2mm}
We first provide estimates for the likelihood of satisfiability of the conditions in Theorem \ref{thm:main}. Specifically, we randomly generate linear P-SCMs and linear DS-P-SCMs with fixed expected degrees among observed variables and sources, and investigate what portion of them satisfy the conditions. We refer to the generating model as a linear DS-P-SCM when each observed variable in the generated P-SCM has a distinct source. Next, we evaluate the performance of Algorithm \ref{alg:recovery} for P-SCM recovery on synthetic models under different settings. Finally, for real data, we consider the closing prices of five world stock indices, and model the corresponding returns as a linear P-SCM.\footnote{Our code is available at:
\url{https://github.com/Yuqin-Yang/Propagation-SCM}.}

\begin{figure}%[tbp]
\centering
\includegraphics[width=\textwidth]{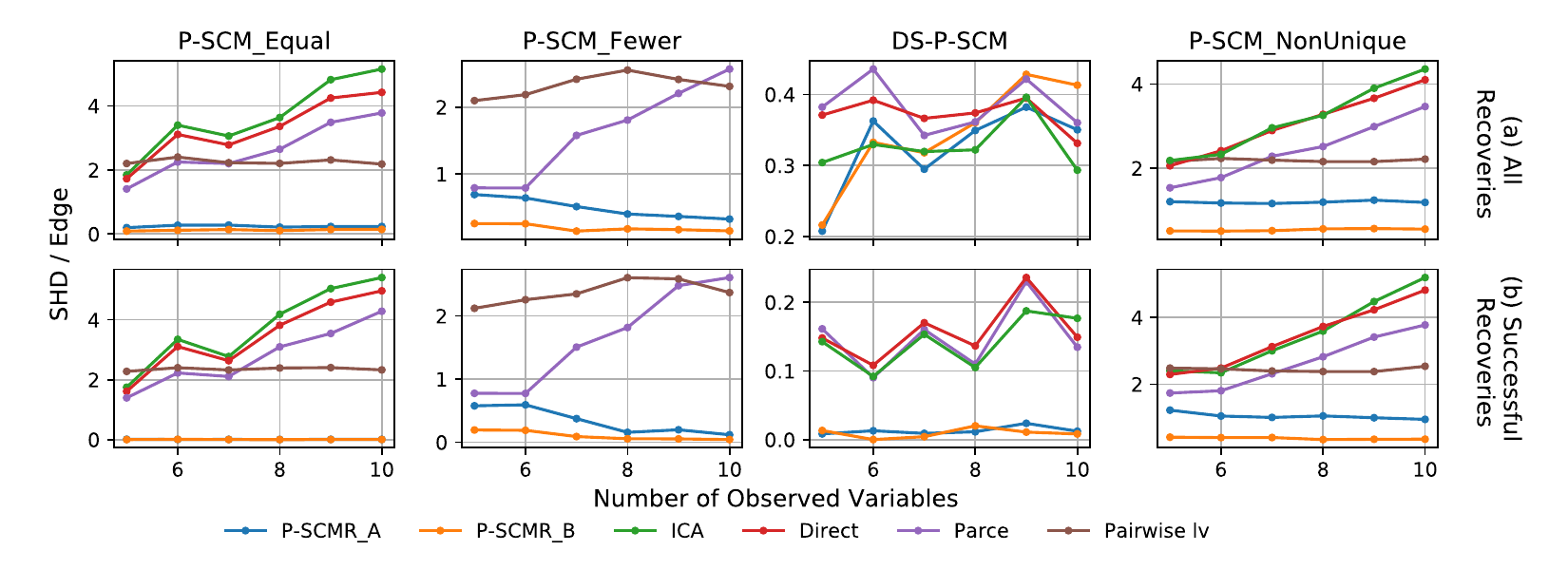}
\caption{Performance of algorithms in terms of SHD/Edge metric (lower value means better performance). P-SCMR\_A, P-SCMR\_B represent the recoveries of $\bA$, $\bB$ by our algorithm.} %Rows: Whether considering all ICA recoveries or successful recoveries only. Columns: Settings (1)-(4).
\label{fig:nips2021_exp2} 
\end{figure}
{\bf Satisfiability of the conditions.}
We test the satisfiability of our derived conditions on linear P-SCM and linear DS-P-SCM with different pairs of $(d_e,d_o)$, where $d_e (d_o)$ is the expected number of causal (exogenous) connections for each observed (source) variable. 
Figure \ref{fig:nips2021_exp1} (left) shows the average number of generated models (attempts) required to obtain a model that satisfies our conditions, plus/minus half the standard deviations w.r.t. the source to observed node ratio (the number of sources divided by the number of observed variables). 
The result shows that our conditions are likely to be satisfied on (1) linear P-SCMs when the number of sources is large and the average degrees are small, and (2) linear DS-P-SCMs when both the number of sources and average degrees are small. We believe that satisfiability of our conditions depends on (1) the existence and similarity of exogenous connections among observed variables, (2) the ratio of overlap among exogenous connections (how many sources are shared among observed variables) and (3) the structural complexity of the causal connections among observed variables. See Appendix \ref{app:simulation_satisfiability} for more details. 

{\textbf{Performance of recovery.}}
We evaluate the performance of our recovery algorithm on models generated under each of the following settings: 
(1) P-SCM with equal number of observed variables and sources (\texttt{P-SCM\_Equal}); 
(2) P-SCM with fewer sources than observed variables (\texttt{P-SCM\_Fewer}) (this indicates that deterministic relation must exist); 
(3) DS-P-SCM with more sources than observed variables (\texttt{DS-P-SCM});
(4) Same as Setting (1) but the model does not satisfy our conditions (\texttt{P-SCM\_NonUnique}).
We only select the generating models which satisfy Conditions \ref{condition:unique_component} and \ref{condition:marriage} in the first three settings. The sources are independently drawn from uniform distributions; we use FastICA \citep{hyvarinen1999fast} and ReconstructionICA \citep{le2011ica} for BSS. 

We compare our proposed algorithm with ICA-LiNGAM \citep{shimizu2006linear}, DirectLiNGAM \citep{shimizu2011directlingam}, ParceLiNGAM \citep{tashiro2014parcelingam}, Pairwise lvLiNGAM \citep{entner2010discovering}, in terms of the differences between recovered and true adjacency matrix $\bA$, and recovered and true exogenous connection matrix $\bB$ (after normalization and repermutation).\footnote{A well-known class of methods for causal structure learning in the presence of latent confounders is FCI algorithm \citep{spirtes2000causation} and its variants. Yet, due to the non-compatibility of the requirements of these methods with our model, we did not compare our approach with these methods (see Appendix \ref{app:simulation_recovery} for more details).} Evaluating $\bB$ is only applicable to our algorithm since other methods do not return $\bB$. We report the results for the normalized Structural Hamming Distance over true edges (SHD/Edges); comparisons with additional metrics appear in Appendix \ref{app:simulation_recovery}.
% using the following four test metrics: (1) normalized Structural Hamming Distance over the number of edges in the true model (SHD / Edge); (2) Frobenius norm; (3) Precision; (4) Recall.
% We repeat the simulation until 50 models can be successfully recovered by ICA, and report the four metrics averaged over all generated models. 
We compare the performance under (i) all ICA recoveries and (ii) only the successful recoveries ($\tilde{\bW}$ recovered by ICA is highly accurate). Figure \ref{fig:nips2021_exp2} shows the performance of recoveries for all algorithms (whenever applicable), averaged over models randomly generated under Settings (1)-(4). The number of observed nodes ranges from 5 to 10.

We observe that when the generating model is a P-SCM, our algorithm can learn the model more accurately than existing algorithms, even if the conditions are not satisfied. This is because existing algorithms may misinterpret the shared sources among observed variables as confounding or direct causal connections, which results in adding additional edges to the recovered graph. When the generating model is a linear DS-P-SCM, the performance of our method is also comparable to 
existing algorithms. Besides, our algorithm performs significantly better than others when the ICA recovery is successful. Our algorithm assumes that the true mixing matrix $\bW$ can be recovered up to permutation and scaling of its columns, hence depends heavily on the accuracy of BSS recovery. 

% \end{wrapfigure}

{\textbf{Performance on real data.}} 
We test the performance of our recovery algorithm on the daily closing prices of the following five world stock indices: DJI, N225, N100, HSI, SSEC \citep{hyvarinen2010estimation,salehkaleybar2020learning}.
% Let $c_i(t)$ be the closing price of the $i$-th index on day $t$. We consider the corresponding return of index $i$ at day $t$ $R_{i}(t)=\left(c_{i}(t)-c_{i-1}(t)\right) / c_{i-1}(t)$, for all $i=2,3,\cdots, T-1$. 
We consider the corresponding return (relative daily change) of each index as an observed variable, and model these observed variables to be generated by a linear P-SCM with five sources, where the sources are non-Gaussian random variables. 
% We apply FastICA for source separation from observations with bootstrapping. 
Using P-SCM Recovery algorithm, the directed graph among these observed variables can be recovered as the right plot in Figure \ref{fig:nips2021_exp1}. We observe that DJI is a root node of this DAG. Further, DJI, N225 N100 all have causal effects on HSI. Both observations are known to be true either from common belief in economy, or from previous results in \citep{hyvarinen2010estimation}.

% \vspace{-4mm}
\section{Conclusion} \label{sec:conclusion}
\vspace{-2mm}
We proposed the linear propagation SCM (P-SCM), which establishes a way of allowing observed variables in a linear structural causal model to deterministically depend on other observed variables or latent confounders. We considered the problem of finding the necessary and sufficient conditions to uniquely identify a linear P-SCM based on observational data. We proposed an algorithm to recover the underlying causal model when a linear P-SCM is assumed to be the data generating process. As mentioned earlier, the performance of our proposed algorithm relies on the accuracy of blind source separation (BSS) recovery. Future directions include investigating algorithms that are robust to BSS errors. Further, it is of interest to investigate whether relaxing the unique identifiability requirement could result in simpler necessary and sufficient conditions that can be efficiently verified.

% Acknowledgments---Will not appear in anonymized version
% \acks{This research was in part supported by the Swiss National Science Foundation under NCCR Automation, grant agreement 51NF40\_180545 and Swiss SNF project 200021\_204355 /1.}
\vspace{-4mm}
\section*{Acknowledgments}
\vspace{-2mm}
This research was in part supported by the Swiss National Science Foundation under NCCR Automation, grant agreement 51NF40\_180545 and Swiss SNF project 200021\_204355 /1.

\small
\bibliographystyle{apalike}   %apalike
\bibliography{Mylib.bib}

\appendix

\section{Related work: A review of causal discovery methods} \label{app:related_work}
Conventional methods for recovering causal directions from observational data often make use of Bayesian graphical models \citep{pearl2009causality,spirtes2000causation}. Bayesian networks allow for the use of efficient algorithms, such as constraint-based PC and Fast Causal Inference (FCI) algorithms \citep{pearl2009causality,spirtes2000causation}, and score-based Greedy Equivalent Search (GES) algorithm \citep{chickering2002optimal}. However, these methods suffer from an identifiability problem. In particular, they only allow for estimating the underlying causal structure from data up to its {\it{Markov equivalence class}}, i.e., the set of structures representing the same conditional independence assertions under causal sufficiency. It turns out the Markov equivalence of the causal structure is a fundamental limit of identification for an underlying data generating process that is a linear DS-SCM with additive Gaussian noises.  

Further assumptions on the data generating process can narrow down the set of possible models compatible with the observed data. These assumptions can be made on (i) the functional form relating the observed variables and/or (ii) the distributions of the sources (noises). For example, \citet{shimizu2006linear} showed that assuming a linear non-Gaussian additive model (LiNGAM), in which the noises have arbitrary non-Gaussian distributions, enables unique identifiability of the causal structure under causal sufficiency. Alternative assumptions may also lead to unique identifiability: \citep{peters2014identifiability} showed that if all variances of the noises in a linear Gaussian DS-SCM are equal, then one can uniquely identify the causal structure.  \citet{hoyer2009nonlinear} assumed (i) specific types of nonlinear relationships among observed variables and (ii) additive exogenous noises, and showed that, for the case of two observed variables, nonlinearity allows for identifying the causal structure. 

These developments, including LiNGAM algorithm \citep{shimizu2006linear} and its variant DirectLiNGAM \citep{shimizu2011directlingam} assume causal sufficiency. Without considering the effect of latent confounders, one may infer wrong causal relations among observed variables. Studying causal models with latent confounders might be essential to infer the correct causal structure. \citet{hoyer2008estimation} and \citet{salehkaleybar2020learning} considered an extension of LiNGAM in the presence of latent confounders (lvLiNGAM). Please refer to Appendix \ref{Sec:ICA} for detailed discussion about both methods.

LiNGAM algorithm is partly based on independent component analysis (ICA) \citep{comon1994independent,hyvarinen2002independent}. The identifiability proof and the recovery algorithm both begin with using ICA to estimate the source mixing matrix from observations. In lvLiNGAM, the number of sources is larger than the number of observed variables since the latent confounders are associated with additional sources. This results in an overcomplete basis for the mixing matrix, and hence lvLiNGAM is based on overcomplete ICA \citep{lewicki2000learning,eriksson2004identifiability}. 

Our linear P-SCM generalizes lvLiNGAM to the case when 
% observed variables are allowed to deterministically depend on other observed variables. In this case, 
the number of sources can be equal or less than the number of observed variables, since each observed variable is not necessarily associated with a distinct source. Instead, each observed variable is exogenously connected to a subset of the jointly independent set of sources, which allows the observed variable to deterministically depend on other observed variables and latent confounders. We further assume {\it{separability}} of the sources from observations, which means all sources composing each observed variable can be correctly estimated, up to scaling and permutation of the coefficients. 

% As mentioned in Section \ref{sec:intro}, our linear P-SCM is equivalent to a linear DS-SCM with linear latent confounding and jointly independent sources if deterministic relations are allowed in the DS-SCM (sources are allowed to have zero variance). 
SCMs with deterministic relations have been previously considered in the literature. In \citep{geiger1990identifying, spirtes2000causation}, D-separation\footnote{This is different from the classical d-separation (with small letter d) described in \citep[Section 1.2]{pearl2009causality}.} condition is proposed for graphically determining conditional independencies when deterministic relations are allowed. Yet, it remains unclear whether D-separation condition can capture all conditional independencies induced from the distribution.
Further, a few approaches have been proposed for causal discovery. When the system only consists of two variables with one variable deterministically causing the other, \citet{daniusis2012inferring} and \citet{janzing2012information} showed that the correct causal direction can be learnt if there is no correlation between the density of the cause variable and the slope of the deterministic function (w.r.t. a reference measure). Their analysis however does not hold for linear relations. \citet{janzing2010telling} and \citet{chen2013nonlinear} considered the deterministic relation between two high-dimensional observed variables, and showed that the correct causal direction can be recovered using the trace of the covariance matrix of the cause variable and the deterministic function. The trace method in \citep{chen2013nonlinear} was extended by \citep{zeng2021nonlinear} to the case of multi-variable causal discovery. However, all these methods assume that  
there are only deterministic relations among observed variables.

In a system where both deterministic and non-deterministic relations are present, \citet{scheines1996tetrad} considered recovering the reduced model where all deterministic variables are removed. \citet{luo2006learning} adapted the conventional constraint-based IC algorithm \citep{pearl2009causality} and added new rules in the independence tests to detect deterministic relations. \citet{mabrouk2014efficient} combined constraint-based methods with greedy search \citep{tsamardinos2006max}, where the deterministic relation is detected by calculating the conditional entropy among variables. \citet{lemeire2012conservative} introduced information equivalence, where two variables are information equivalent (w.r.t. a third variable) if knowing one variable is equivalent to knowing the other, from the viewpoint of the third variable. Information equivalence is thus a generalization of deterministic relations. \citet{lemeire2012conservative} adapted the PC algorithm with additional tests to detect information equivalence among variables.
The aforementioned methods all suffer from the same identifiability problem as in conventional causal discovery methods, where the underlying causal structure can only be estimated up to certain equivalence classes. In fact, majority of these works do not discuss the capability of identifying the underlying structure by their proposed algorithms (i.e., the equivalence class of the recovery result). Further, these methods do not consider the presence of latent confounders in the system. %Due to these reasons, we do not compare our proposed algorithm for unique identifiability with these methods. 

\section{Detailed descriptions of model assumptions} \label{app:assumptions}
\subsection{P-SCM faithfulness assumption}
For a G-SCM, faithfulness assumption means that all marginal and conditional independencies in the underlying causal graph are captured (required) by the Markov condition.\footnote{The Markov condition implies a Markov factorization of the underlying probability distribution such that each variable is conditionally independent on its non-descendant variables given its parents.} For linear DS-SCMs, it is often assumed that any variable is marginally dependent on all its descendants \citep{salehkaleybar2020learning}. Thus, the faithfulness assumption translates to the linear coefficients not canceling out causal effects among variables. That is, when multiple causal effects exist from one observed variable to another, their combined effect is not exactly equal to zero. 
%In the language of linear causal models, faithfulness translates to the linear coefficients in $\bA$ not canceling out causal effects among variables. 

P-SCM faithfulness assumption extends this idea to linear P-SCMs. Specifically, Assumption \ref{assumption:faithfulness}(a) states that when multiple distinct causal paths exist from some observed variables-- containing source a $s_j$-- to an observed variable $x_i$, their combined effect do not cancel out one another, or cancel out the exogenous connection from $s_j$ to $x_i$ (i.e., $s_j$ component in $x_i$ is not equal to zero).
% when multiple causal effects exist from one source to one observed variable, their combined effect does not exactly equal to zero. 
This is equivalent to the faithfulness assumption in \citep{salehkaleybar2020learning} for linear DS-SCMs, and is less restrictive than the conventional faithfulness assumption for G-SCMs.
%
% , and both are equivalent when linear P-SCMs are reduced to linear DS-SCMs (see Section \ref{sec:main_result}).
Assumption \ref{assumption:faithfulness}(b) considers the coefficients of matrix $\bB$, and can be explained as: Any source mixture $\tilde{s}_i$ cannot be written as a linear combination of other source mixtures. Assumption \ref{assumption:faithfulness}(b) is to prevent the case when the coefficients of the exogenous connections from a subset of two or more sources (say $\tilde{S}$) to an observed variable are proportional to the coefficients of the exogenous connections from $\tilde{S}$ to another observed variable. For example, suppose $x_1=b_{11}s_1 + b_{21}s_2$ and $x_2=b_{12}s_1 + b_{22}s_2$. Assumption \ref{assumption:faithfulness}(b) prevents $(b_{11}, b_{21})$ from being proportional to $(b_{12}, b_{22})$.

Note that when all model coefficients $\{a_{ij}\}$ and $\{b_{ij}\}$ are independently drawn from continuous distributions, Assumptions (a) and (b) are satisfied almost surely \citep{meek2013strong}, hence P-SCM faithfulness is a reasonable assumption.

\subsection{Separability assumption}
Separability can be achieved using blind source separation (BSS) methods, which is to separate a set of sources from a set of their mixtures given little to no information about the sources and the mixing process. For example, when the sources are known to be non-Gaussian random variables, the mixing matrix $\bW$ can be recovered using Independent Component Analysis (ICA) or overcomplete ICA methods, {depending on the number of observed variables and sources} \citep{comon1994independent, hyvarinen2002independent, lewicki2000learning,eriksson2004identifiability}. Separability can also be achieved when the sources are piecewise constant functionals satisfying a set of mild conditions \citep{behr2018multiscale}. We discuss BSS in more details in Appendix \ref{app:BSS}.

To satisfy the separability property, we assume the availability of a large number of data vectors $\{X^{(i)}: i\in[n]\}$, and that each is generated according to the described process, and using the same coefficients $\{a_{ij}\}$ and $\{b_{ij}\}$.

% Separability can be achieved, for instance, using multi-scale blind source separation when the exogenous sources are piecewise constant functionals satisfying a set of mild conditions, see \citep{behr2018multiscale}. When the sources are non-Gaussian random variables, the mixing matrix $\bf{W}$ can be recovered using Independent Component Analysis (ICA) or overcomplete ICA methods \citep{comon1994independent,hyvarinen2002independent,lewicki2000learning,eriksson2004identifiability}. To satisfy the separability property, we assume the availability of a large number of data vectors $\{X^{(i)}: i\in[n]\}$, and that each is generated according to the described process, and using the same coefficients $\{a_{ij}\}$ and $\{b_{ij}\}$. In the next section, we describe these methods in more details.

\section{Blind source separation methods} \label{app:BSS}
Blind source separation (BSS) is the problem of separating a set of source signals (sources) from a set of mixed signals (mixtures) given little to no information about the sources and the mixing process \citep{comon2010handbook}. In the linear setting where the mixtures are linear combinations of the sources, the blindness often refers to not knowing the source realizations nor the mixing weights. Without any further knowledge about the sources, the problem is infeasible, and hence further assumptions on the sources are needed to facilitate source separation. An example of BSS is the independent component analysis framework in which the sources are assumed to be drawn from independent non-Gaussian distributions. Other examples include the statistical blind source separation regression (SBSSR) model in \citep{behr2018multiscale} and non-negative matrix factorization in \citep{donoho2004does}. In this section, we provide a brief description of these methods, and present how to utilize them as seeds to our causal structure learning framework.     

\subsection{Independent component analysis}\label{Sec:ICA}
Independent component analysis \citep{comon1994independent} is a statistical technique to separate independent, non-Gaussian random variables (sources) from their observed linear mixtures. Specifically, consider a vector of $p$ observed variables (mixtures) $X=[x_1,\cdots, x_p]^{\top}$ which is generated as
\begin{align}
X = \bW S,
\label{eq:ICA}
\end{align}
where $S=[s_1,\cdots,s_m]^{\top}$ are $m$ unknown real-valued independent non-degenerate random variables (sources). $\bW$ is a constant $p\times m$ unknown mixing matrix. $(\bW,S)$ is called the {\it{representation}} of $X$.

If the observed data is invertible mixtures of non-Gaussian components (i.e., the mixing matrix $\bW$ is invertible), \citet{comon1994independent} showed that the representation of $X$ can be uniquely identified up to scaling and permutation of the columns of $\bW$ (given enough data vectors $X$). Further, if $\bW$ is of full column rank and at most one source in $S$ is Gaussian, the representation of $X$ is still identifiable up to scaling and permutation indeterminacies \citep{eriksson2004identifiability}. For detailed explanation about ICA, the reader can refer to \citep{hyvarinen2000independent,comon2010handbook}.

When the number of observed variables is less than the number of sources, \citet{eriksson2004identifiability} showed that if all the $m$ sources are non-Gaussian and the representation of $X$ is irreducible, then $\bW$ can be identified up to scaling and permutation of its columns. Irreducibility means that the columns of $\bW$ are pairwise linearly independent. If two columns in $\bW$ are linearly dependent, with corresponding sources $s_i$ and $s_j$, then it is impossible to to distinguish between the true representation of $X$ and another representation with $m-1$ sources, where the two linearly dependent columns are merged into a single column.

We now describe LiNGAM algorithm, which uniquely recovers the causal structure of the underlying model when the mixing matrix is invertible and the sources are non-Gaussian \citep{shimizu2006linear}. The underlying DS-SCM can be written as
\begin{align}
\label{eq:LiNGAM_1}
    X= \bA X+ S= \bW S
\end{align}
where $S$ consists of $m=p$ jointly independent non-Gaussian sources, and $\bW=(\bI-\bA)^{-1}$. 

Note that \eqref{eq:LiNGAM_1} fits into the standard ICA framework in \eqref{eq:ICA}. Thus, $\bW$ can be identified up to scaling and permutation of its columns. Without prior knowledge about the scale and ordering of the sources, the following general ICA model for $X$ holds:
\begin{align}
\label{eq:BSS_output}
X=\tilde{\bW} S'.
\end{align}
$\tilde{\mathbf{W}}$ is the recovered mixing matrix, with scaling and permutation indeterminacies, and is given by
%with the indeterminacies of scaling and permutation of its columns. That is,
\begin{align}
\label{eq:W_hat}
\tilde{\bW}=\bW \mathbf{P} \mathbf{\Gamma},
\end{align}
where $\mathbf{P}$ is a permutation matrix and $\mathbf{\Gamma}$ is a diagonal scaling matrix. $S'$ contains the corresponding set of sources with reordering and rescaling that correspond to $\mathbf{P}$ and $\mathbf{\Gamma}$. 

The corresponding causal model, represented by matrix $\bA$, can be uniquely identified due to the acyclicity assumption. Specifically, by some scaling and permutation of its columns and rows, $\tilde{\bW}^{-1}$ can be converted uniquely to a lower triangular matrix with all ones on its main diagonal. 

In the presence of latent variables, \citet{hoyer2008estimation} and \citet{salehkaleybar2020learning} utilized overcomplete ICA to extend LiNGAM algorithm. Given the output of the overcomplete ICA, the recovered mixing matrix $\tilde{\bW}$ is unique up to scaling and permutation of its columns, under the assumption of irreducibility. The number of latent variables in the system is assumed to be known. The task is to identify the columns of $\tilde{\bW}$ which correspond to exogenous noises associated with observed variables, and subsequently apply LiNGAM algorithm after eliminating the columns corresponding to latent variables. \citet{hoyer2008estimation} proposed an algorithm which brute-forces all possible classifications of the columns of $\tilde{\bW}$ (whether each of the columns corresponds to distinct source associated with an observed variable, or to a latent confounder). The algorithm then identifies the possible selections that are compatible with faithfulness and acyclicity assumptions. More recently, \citet{salehkaleybar2020learning} proposed a more efficient algorithm that identifies the set of columns corresponding to observed variables. The idea is to learn the skeleton of the underlying graph via pairwise comparisons of the observed variables (based on their corresponding rows in $\tilde{\bW}$). This is possible due to faithfulness and acyclicity assumptions. Using the learnt skeleton, the algorithm selects only the columns whose non-zero entries are compatible with the observable descendants of each observed variable.

\subsection{Statistical blind source separation regression model}\label{Sec:SBSSR}
Statistical blind source separation regression (SBSSR) model is a finite alphabet BSS method proposed in \citep{behr2018multiscale} to separate piecewise constant source functions from their linear mixtures. Suppose we have a set of $m$ source functions $\setS=\{s_1,s_2,\cdots,s_m\}$, each of which consists of an array of constant segments (i.e., a step function) on $[0,1]$. The function values are selected from a known finite alphabet, yet the jump sizes, numbers, and locations of change points for each function are unknown. Specifically, let $\mathcal{U}=\{u_1,u_2,\cdots,u_k\}\subset \mathbb{R}$ be a known finite and totally ordered alphabet, i.e., $u_1<u_2< \cdots <u_k$. Each source function belongs to the class 
\begin{equation*}
\setS =
\left\{\sum_{j=0}^{K} \theta_{j} \one_{\left[\tau_{j}, \tau_{j+1}\right)}: \theta_{j} \in \mathcal{U}, 0=\tau_{0}<\cdots<\tau_{K}<\tau_{K+1}=1, K \in \mathbb{N}\right\}.
\end{equation*}
$K$ is the unknown number of changes, which is assumed to be finite.  $\theta_1,\cdots, \theta_K$ are the function values selected from the alphabet $\mathcal{U}$ such that $\theta_j\neq \theta_{j+1}$ for $j=0,1,\cdots,K$. $\tau_1,\cdots, \tau_K$ are the change points. We observe samples of $p$ linear mixtures $\{x_k(t),~ k=1,\cdots,p\}$ at $n$ uniformly selected location points $t_i$ in $[0,1]$: 
\begin{equation}
x_k(t_i) = \sum_{j=1}^m w_{kj} s_j(t_i) + \sigma \epsilon_{ki} =  W_k^{\top} S(t_i) + \sigma \epsilon_{ki}; \qquad i=1,\cdots,n;~ k=1,\cdots,p.
\end{equation}
$W_k=[w_{k1},\cdots,w_{km}]^{\top}$ is the vector of mixing weights for mixture $x_k$ with
\begin{equation*}
W_k \geq 0,~ \sum_{j=1}^m w_{kj}=1,\qquad k=1,\cdots, p.
\end{equation*}
$S(t_i)=[s_1(t_i),\cdots,s_m(t_i)]^{\top}$ is the vector of values of the source functions at point $t_i$ for all $i\in [n]$. $\epsilon_{ki}$ is an additive Gaussian noise with zero mean and unit variance, and $\sigma>0$ is the known noise variance. \citet{behr2018multiscale} showed that, under mild separability conditions among weights and source functions, both the source functions and the mixing weights can be identified up to permutations, including the number of change points, change point locations, and the function values on each segment for all source functions. 

Unlike ICA, the mixing matrix $\bW$ in SBSSR does not necessarily need to satisfy the irreducibility assumption, due to the prior knowledge about the sources alphabet. In fact, the recovered mixing matrix $\tilde{\bW}$ using SBSSR can include two linearly independent columns, while the method is able to separate the corresponding sources. However, even with the ability to separate such sources, we require to merge them into one combined source, since the two corresponding generating models are observationally equivalent, as we will shortly explain. 

\subsection{Non-negative matrix factorization}\label{Sec:NMF}
Non-negative matrix factorization (NMF) is another BSS method to decompose large matrices into the multiplication of low-rank, non-negative smaller matrices \citep{lee1999learning}. Suppose each observed variable is an $n_1 \times n_2$ matrix $\mathbf{X}_k$ with rank $r_k$. Using NMF, $\mathbf{X}_k$ can be decomposed into
\begin{equation}
\mathbf{X}_k = \mathbf{W}_k \mathbf{F}_k^{\top} = \sum_{i=1}^{r_k} W_{ki} F_{ki}^{\top}.
\label{eq:nnmf}
\end{equation}
$\mathbf{W}_k$ is a non-negative $n_1\times r_k$ weight matrix. $\mathbf{F}_k$ is a non-negative $n_2 \times r_k$ feature matrix. $W_{ki}$, $F_{ki}$ are the column vectors of $\mathbf{W}_k$, $\mathbf{F}_k$ for all $i=1,\cdots, r_k$. If we consider the feature vectors $F_{ki}$ as sources, then we can write each observed matrix $\mathbf{X}_k $ into a linear combination of sources (here, both sources and weights are column vectors). \citet{donoho2004does} showed that under certain conditions on $\mathbf{W}_k$ and $\mathbf{F}_k$, the factorization (\ref{eq:nnmf}) is unique up to permutation and scaling of the feature column vectors, i.e., the sources.

\subsection{Application to our model}\label{sec:application_to_our_model} 
Under different assumptions on the sources and the mixing matrix, we can use the aforementioned BSS methods to estimate the correct representation of the observed variables.
%i.e., 
%estimate the mixing matrix $\bW$ and the corresponding sources $S$.
In the BSS methods described in Appendices \ref{Sec:ICA}-\ref{Sec:NMF}, the permutation and scalability indeterminacies always exist, except for SBSSR, where we do not have scalability indeterminacy due to the prior information about the sources alphabets and the mixing weights. In general, more prior information about the sources may eliminate the scalability indeterminacy. Another example where the scalability indeterminacy relaxes is when the sources in the ICA framework are all known to have unit variance. In the same spirit, having prior information about the ordering may eliminate the permutation indeterminacy. 

The output of BSS is the factorization of $X$ as in \eqref{eq:BSS_output}, where under permutation and scalability indeterminacies, $\tilde{\bW}$ is given by \eqref{eq:W_hat} , and $S'$ is given by
\begin{equation*}
S' = \mathbf{\Gamma}^{-1}\mathbf{P}^{\top} S.
% \label{eq:S_dash}
\end{equation*}
$\mathbf{P}$ is the $m\times m$ permutation matrix with $\mathbf{P}\mathbf{P}^{\top}=\bI$, and $\mathbf{\Gamma}$ is the diagonal scaling matrix. Recall that in LiNGAM, the mixing matrix $\mathbf{W}=(\bI - \bA)^{-1}$ is lower triangular with all ones on the main diagonal, since $\bA$ is strictly lower triangular due to the DAG-ness assumption of the underlying graph. By contrast, in our model, the mixing matrix in (\ref{eq:system_model_4}) does not possess these structural restrictions since $\mathbf{W}=(\bI - \bA)^{-1}\bB$, where $\bB$ can have an arbitrary structure. 

As in LiNGAM, our recovery algorithm is robust to permutation and scaling ambiguities. From \eqref{eq:system_model_4} and \eqref{eq:W_hat}, the recovered mixing matrix in our model can be written as 
\begin{align}
\tilde{\bW} =\bW \mathbf{P} \mathbf{\Gamma}=(\bI - \bA)^{-1} \bB \mathbf{P} \mathbf{\Gamma}.\nonumber
\label{eq:bssrecovery0}
\end{align}
By writing $\tilde{\bB}$ as 
\begin{align}
\label{eq:B_tilde}
\tilde{\bB} = \bB \mathbf{P} \mathbf{\Gamma},
\end{align} 
we have 
\begin{align}
\tilde{\bW} =(\bI - \bA)^{-1} \tilde{\bB}.
\label{eq:bssrecovery1}
\end{align}
Our recovery algorithm is in fact able to recover $(\bI-\bA)^{-1}$ and $\tilde{\bB}$ using $\tilde{\bW}$, where $\tilde{\bB}$ is a column-repermuted and rescaled version of $\bB$. The permutation indeterminacy of $\bB$ is immaterial because there exists no ordering among the exogenous connections from the sources, and we are not concerned about the scalability indeterminacy in recovering the strengths of these exogenous connections. Therefore, permutation and scalability indeterminacies do not affect the recovery performance of our algorithm. In the following, we provide an example for the aforementioned indeterminacies.  

\begin{example}\label{example_0}
Consider the generating model in Figure \ref{fig:example_1}. In the matrix form, the generating model is 
\begin{figure}
\centering
\begin{tikzpicture}[thick, scale=.4]
\foreach \place/\name in {{(-2.5,0)/x_1}, {(2.5,0)/x_2}, {(0,-1.5)/x_3}}
    \node[obsvar, label=center:$\name$] (\name) at \place {};
  \foreach \source/\dest in {x_1/x_3, x_2/x_3}
    \path[causal] (\source) edge (\dest);

\foreach \place/\name in {{(-2.5,3.5)/s_1}, {(0,3.5)/s_2}, {(2.5,3.5)/s_3}}
    \node[source, label=center:$\name$] (\name) at \place {};
  \foreach \source/\dest in {s_1/x_1, s_2/x_1, s_1/x_2, s_1/x_3, s_3/x_2}
    \path[exogenous] (\source) edge (\dest);
\end{tikzpicture}
\caption{Generating model for Example \ref{example_0}.}
\label{fig:example_1}
\end{figure}
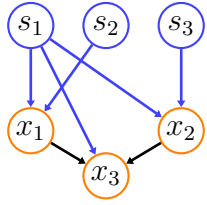

\begin{align} \label{eq:example1_1}
\begin{bmatrix}
x_1 \\
x_2 \\
x_3 \\
\end{bmatrix}
 = 
\begin{bmatrix}
0 & 0 & 0 \\
0 & 0 & 0 \\
a_{31} & a_{32} & 0 \\
\end{bmatrix}
\begin{bmatrix}
x_1 \\
x_2 \\
x_3 \\
\end{bmatrix}
+
\begin{bmatrix}
b_{11} & b_{12} & 0 \\
b_{21} & 0 & b_{23} \\
b_{31} & 0 & 0 \\
\end{bmatrix}
\begin{bmatrix}
s_1 \\
s_2 \\
s_3 \\
\end{bmatrix}
.
\end{align}
The observed variables can be represented as linear combinations of the sources as follows: 
\begin{align} \nonumber
X = \bW S
=
\begin{bmatrix}
b_{11} & b_{12} & 0 \\
b_{21} & 0 & b_{23} \\
w_{31} & a_{31}b_{12} & a_{32}b_{23} \\
\end{bmatrix}
\begin{bmatrix}
s_1 \\
s_2 \\
s_3 \\
\end{bmatrix},
\end{align}
where $w_{31}=a_{31}b_{11}+ a_{32}b_{21} + b_{31}$.

Given samples of the observed variables $X$, we can apply BSS methods to learn the representation of $X$ up to permutation and scaling of the columns of $\bW$. Suppose the recovered mixing matrix is
\begin{align} \nonumber
\tilde{\bW} 
= \bW \mathbf{P} \mathbf{\Gamma}
=
\begin{bmatrix}
\alpha b_{12} & \gamma b_{11} & 0 \\
0 & \gamma b_{21} & \beta b_{23} \\
\alpha a_{31}b_{12} & \gamma w_{31} & \beta a_{32}b_{23} \\
\end{bmatrix}.
\end{align}
The permutation and scaling matrices are given by
\begin{align} \nonumber
\mathbf{P}
=
\begin{bmatrix}
0 & 1 & 0 \\
1 & 0 & 0 \\
0 & 0 & 1 \\
\end{bmatrix}
, \quad
\mathbf{\Gamma}
=
\begin{bmatrix}
\alpha & 0 & 0 \\
0 & \beta & 0 \\
0 & 0 & \gamma \\
\end{bmatrix}
,
\end{align}
and the corresponding sources are
\begin{align} \nonumber
S' = 
\begin{bmatrix}
s'_1 \\
s'_2 \\
s'_3 \\
\end{bmatrix}
=
\begin{bmatrix}
\alpha^{-1}s_2 \\
\gamma^{-1}s_1 \\
\beta^{-1}s_3 \\
\end{bmatrix}.
\label{eq:ex1_s_dash}
\end{align}
$\alpha, \beta, \gamma \neq 0$ are the scaling coefficients. 

Given the recovered mixing matrix $\tilde{\bW}$, our P-SCM recovery algorithm uniquely recovers $\bA$ and $\tilde{\bB}=\bB \mathbf{P} \mathbf{\Gamma}$, as explained in Section \ref{sec:main_result}. Thus the recovered model is 
% which we shall see how in detail in Section \ref{sec:main_result}

\begin{align}
\label{eq:example1_2}
X = \bA X+\tilde{\bB} S'
=
\begin{bmatrix}
0 & 0 & 0 \\
0 & 0 & 0 \\
a_{31} & a_{32} & 0 \\
\end{bmatrix}
\begin{bmatrix}
x_1 \\
x_2 \\
x_3 \\
\end{bmatrix}
+
\begin{bmatrix}
\alpha b_{12} & \gamma b_{11} & 0 \\
0 & \gamma b_{21} & \beta b_{23} \\
0 & \gamma b_{31} & 0 \\
\end{bmatrix}
\begin{bmatrix}
s'_1 \\
s'_2 \\
s'_3 \\
\end{bmatrix}
.
\end{align}
Note that $\tilde{\bB}S'=\bB S$, hence the recovered model in \eqref{eq:example1_2} is identical to the generating model in \eqref{eq:example1_1}. However, we do not learn the representation $\bB S$ uniquely, instead we learn $\tilde{\bB} S'$. 
%which is irrelevant as we explained above. 
\end{example}

%It is worth noting that 
The irreducibility assumption in our linear P-SCM is without loss of generality. Under P-SCM faithfulness assumption, reducibility
%, i.e., existence of a pair of columns $W_i,~W_j,~ i\neq j$ in the mixing matrix $\bW$ that are linearly dependent,
occurs if and only if two sources $s_i$ and $s_j$ are exogenously connected to a common observed variable, and are not exogenously connected to any other observed variables. 
%This follows because 
Specifically, we can write the corresponding columns in $\bW=(\bI-\bA)^{-1}\bB$ (cf. \eqref{eq:system_model_4}) as
\begin{equation*}
[W_i\quad W_j]=(\bI-\bA)^{-1} [B_i\quad B_j],
\end{equation*}
where $B_i$ and $B_j$ are the columns in $\bB$ corresponding to sources $s_i$ and $s_j$. Since $(\bI-\bA)^{-1}$ is an invertible square matrix, $W_i$ and $W_j$ are linearly dependent if and only if $B_i$ and $B_j$ are linearly dependent. Under P-SCM faithfulness assumption (b), this happens if and only if $[B_i~~ B_j]$ have only one non-zero row, which means $s_i$ and $s_j$ are exogenously connected to a common observed variable, and to this variable only. Therefore, when $W_i$ and $W_j$ are linearly dependent, i.e., $W_i=\alpha W_j, \alpha\neq 0$, the corresponding sources $s_i,s_j$ appear at each node $x_k$ as $w_{kj}(s_j+\alpha s_i)$, where $w_{kj}\in \mathbb{R}$. Since $s_i$ and $s_j$ are connected to the same variable, replacing the exogenous connections from $s_i,s_j$ by a combined source $s_i'=s_j+\alpha s_i$ in the generating model results in a model that has the same distribution as the original generating model (i.e., an observationally equivalent model). 

In conclusion, under different problem settings and assumptions, given observations of variables $X$, we can use BSS methods to recover the mixing matrix $\mathbf{W}$ up to permutation and/or scalability indeterminacies, which have no impact in recovering the true generating model. We call these models separable, as mentioned in Section \ref{sec:model_assumptions}. 

{\bf Connection to LiNGAM.}
According to the above analysis and the comparison between DS-SCM and P-SCM in Section \ref{sec:comparison}, both LiNGAM and lvLiNGAM models are included in our linear P-SCM. Specifically, our model reduces to LiNGAM when each observed variable is directly influenced by a single source, and the sources are jointly independent non-Gaussian random variables with non-zero variances. Here, separability holds by using ICA. Similarly, lvLiNGAM corresponds to the case when the sources are non-Gaussian, and each of the source mixtures $\tilde{s}_i$ has at least one distinct source. As we previously mentioned, for lvLiNGAM, separability hold using overcomplete ICA. 
%Note that, in lvLiNGAM, the number of source variables is larger than the number of observed variables since the latent confounders are associated with additional distinct sources. This results in an overcomplete basis for the mixing matrix, and hence separability holds using overcomplete ICA. Further, a sufficient condition for unique identifiability in lvLiNGAM is that the mixing matrix $\bW$ in \eqref{eq:system_model_4} is irreducible in the sense that its columns are pairwise linearly independent. This is in fact a sufficient condition for overcomplete ICA to return a unique solution up to scaling and permutation of the columns \citep{eriksson2004identifiability}.
In this work, we impose a more general assumption on the {\it{separability}} of the source variables from observations. This framework includes ICA and overcomplete ICA as special cases, but is not restricted to these two methods.

\section{Examples}
\subsection{Generalized distinct source assumption} \label{app:distinct_source}
As mentioned in Section \ref{Generating Model}, Assumption \ref{assumption:distinct_source} is stronger than what is needed for structure learning methods developed for linear DS-SCMs. Here, we provide a weaker version of the distinct source assumption, under which methods developed for linear DS-SCMs still work.

\begin{assumption}[Generalized distinct source assumption]
\label{assumption:distinct_source_generalized}
A linear G-SCM satisfies the distinct source assumption if for each observed variable $x$, there exists a source such that the information regarding that source in the source mixture of $x$ cannot be fully described by the source mixtures of the rest of the variables.
\end{assumption}

%What we mean by ``information regarding that source in the source mixture of $x$" can be explained by the following example. If source $s$

%Regarding the notion of information of a source in the source mixture of an observed variable, 

The intuition behind Assumption \ref{assumption:distinct_source_generalized} can be explained as follows. If, for example, source $s$ is shared between two variables $x_1$ and $x_2$, yet $x_2$ is only a function of a coarsened version of $s$, then the remaining information of $s$ could be considered as a distinct source for $x_1$. In the following we provide a concrete example where depending on the set of functions $\{g_x\}_{x\in \setX}$, the model with the same causal structure may or may not satisfy Assumption \ref{assumption:distinct_source_generalized}.

%Here, we show an example in which whether or not the distinct source assumption is satisfied depends on the set of functions $\{g_x\}_{x\in \setX}$. 

\begin{example}
Consider a linear G-SCM with two observed variables $x_1,x_2$ and two independent source variables $s_1,s_2$. Suppose $s_2$ is discrete with support $\{1,2,3,4\}$, and the generating models of $x_1$ and $x_2$ are
\begin{align*}
x_1 &= g_{x_1}(s_1, s_2)=g_{x_1,1}(s_1) + g_{x_1,2} (s_2), \\
x_2 &= a_{21} x_1 + g_{x_2} (s_2),
\end{align*}
where $g_{x_1,1}$, $g_{x_1,2}$ and $g_{x_2}$ are possibly nonlinear functions.

% $g_{12}(s_4)$

% $s_4=f_1I(s_2 \in \{1,3\})+f_2I(s_2 \in \{2,4\})$

% $s_3\in\{f_3,f_4\}$

% $s_2=1 iff s_4=f_1 and s_3=f_3$\\
% $s_2=3 iff s_4=f_1 and s_3=f_4$\\
% $s_2=2 iff s_4=f_2 and s_3=f_3$\\
% $s_2=4 iff s_4=f_2 and s_3=f_4$

\begin{itemize}
    \item If all three functions are linear, then the information regarding $s_2$ in the mixture of $x_2$, i.e., $g_{x_2}(s_2)$, can be written as a deterministic function of $g_{x_1,2}(s_2)$, the information regarding $s_2$ in the mixture of $x_1$. That is, $x_2$ does not have a distinct source and hence, the model is not a linear DS-SCM.
    
    \item If $g_{x_2} (s_2)$ is a linear function of $s_2$, but $g_{x_1,2} (s_2)$ only depends on whether $s_2$ is an even number (e.g., $g_{x_1,2} (s_2)=c_1\mathbf{1}[s_2\in \{1,3\}] + c_2\mathbf{1}[s_2\in \{2,4\}]$ for some constants $c_1$, $c_2$); then we can write $s_2$ as the combination of two separate sources $s_3$ and $s_4$, where $s_3, s_4\in \{0,1\}$, and 
    \begin{align*}
    &s_2=1 \textit{~~iff~~} s_4=0 \textit{~and~} s_3=0;\\
    &s_2=2 \textit{~~iff~~} s_4=1 \textit{~and~} s_3=0;\\
    &s_2=3 \textit{~~iff~~} s_4=0 \textit{~and~} s_3=1;\\
    &s_2=4 \textit{~~iff~~} s_4=1 \textit{~and~} s_3=1.
    \end{align*}
    Note that $s_2 = 2s_3+s_4+1$, and $g_{x_1,2} (s_2)$ is only a function of $s_4$. We can replace $s_2$ in the model by the combination of $s_3$ and $s_4$, and rewrite the model as
    \begin{align*}
    x_1 &= \tilde{g}_{x_1}(s_1, s_4)=g_{x_1,1}(s_1) + f(s_4), \\
    x_2 &= a_{21} x_1 + g_{x_2} (2s_3 + s_4+1),
    \end{align*}
    where $f(s_4) = c_1\mathbf{1}[s_4=0] + c_2\mathbf{1}[s_4=1]$.
    Therefore, because of $s_3$, the information regarding $s_2$ in the mixture of $x_2$ (i.e., $g_{x_2} (2s_3 + s_4+1)$) cannot be written as a deterministic function of the information regarding $s_2$ in the mixture of $x_1$ (i.e., $f(x_4)$). Further, we can consider $s_3$ as a distinct source of $x_2$, and $x_4$ as a latent confounder between $x_1$ and $x_2$. Therefore, both variables have distinct sources, and the model is a linear DS-SCM.
\end{itemize}

\end{example}

\subsection{Examples showing the distinction between linear P-SCM and linear DS-P-SCM} \label{app:example_equivalence}
In this subsection, we demonstrate the practical significance of linear P-SCM over a linear DS-P-SCM. Specifically, we answer the following question: Using the more general linear P-SCM, can we correctly identify models that are not identifiable using linear DS-P-SCMs? In the following, we provide two examples where linear P-SCMs can explain the underlying causal model, while linear DS-P-SCMs, with or without latent confounders, either are inapplicable or give misleading results. Therefore, the model cannot be recovered using causal discovery methods for linear DS-P-SCMs, but can be correctly recovered by our P-SCM Recovery algorithm.

\begin{example} \label{example_1}
In this example, we begin with the ground truth (i.e., the underlying model), and show that linear DS-P-SCMs fail to explain the model, while our linear P-SCM explains the model. Consider the graph on the left in Figure \ref{fig:example1}, which represents the true generating model for this example. We have five observed variables $x_1,\cdots, x_5$ and four jointly independent sources $s_1,\cdots, s_4$. 
% Solid arrows represent causal connections between observed variables, and dashed arrows represent direct connections from exogenous sources to observed variables. 
Since there are less sources than observed variables, the model cannot be represented by a linear DS-P-SCM with or without latent confounders. %Recall, in linear SCMs, each observed variable must have an independent source component.
This is because each observed variable in linear DS-P-SCMs must be associated with an distinct source that is not shared with other observed variables. Hence the number of independent sources has to be greater than or equal to the number of observed variables in linear DS-P-SCMs.

% Recall that each observed variable  must be associated with an exogenous noise, and even if the noises can be correlated (in the presence of latent confounders), an exogenous noise cannot be deterministically determined by other exogenous noises. 

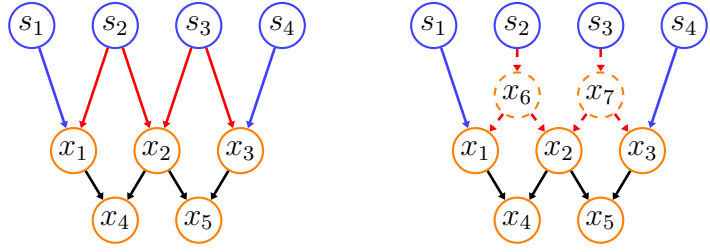
\begin{figure}[t]
\centering
\begin{tikzpicture}[thick, scale=.37]
\foreach \place/\name in {{(-3,0)/x_1}, {(0,0)/x_2}, {(3,0)/x_3}, {(-1.5,-2.5)/x_4}, {(1.5,-2.5)/x_5}}
    \node[obsvar, label=center:$\name$] (\name) at \place {};
  \foreach \source/\dest in {x_1/x_4, x_2/x_4, x_2/x_5, x_3/x_5}
    \path[causal] (\source) edge (\dest);

\foreach \place/\name in {{(-4.5,4.5)/s_1}, {(-1.5,4.5)/s_2}, {(1.5,4.5)/s_3}, {(4.5,4.5)/s_4}}
    \node[source, label=center:$\name$] (\name) at \place {};
  \foreach \source/\dest in {s_1/x_1, s_4/x_3}
    \path[exogenous] (\source) edge (\dest);
  \foreach \source/\dest in {s_2/x_1, s_2/x_2, s_3/x_2, s_3/x_3}
    \path[difference] (\source) edge (\dest);

\end{tikzpicture}
\quad
\quad
\quad
\begin{tikzpicture}[thick, scale=.37]
\foreach \place/\name in {{(-3,0)/x_1}, {(0,0)/x_2}, {(3,0)/x_3}, {(-1.5,-2.5)/x_4}, {(1.5,-2.5)/x_5}}
    \node[obsvar, label=center:$\name$] (\name) at \place {};
  \foreach \place/\name in {{(-1.5,2)/x_6}, {(1.5,2)/x_7}}
    \node[obsvar, dashed, label=center:$\name$] (\name) at \place {};
  \foreach \source/\dest in {x_1/x_4, x_2/x_4, x_2/x_5, x_3/x_5}
    \path[causal] (\source) edge (\dest);
  \foreach \source/\dest in {x_6/x_1, x_6/x_2, x_7/x_2, x_7/x_3}
    \path[difference, dashed] (\source) edge (\dest);

\foreach \place/\name in {{(-4.5,4.5)/s_1}, {(-1.5,4.5)/s_2}, {(1.5,4.5)/s_3}, {(4.5,4.5)/s_4}}
    \node[source, label=center:$\name$] (\name) at \place {};
  \foreach \source/\dest in {s_1/x_1, s_4/x_3}
    \path[exogenous] (\source) edge (\dest);
  \foreach \source/\dest in {s_2/x_6, s_3/x_7}
    \path[difference, dashed] (\source) edge (\dest);
\end{tikzpicture}

\caption{(Left) The ground truth (the true generating model) for Example \ref{example_1}. (Right) The ground truth cannot be modeled as a linear DS-P-SCM with latent confounders. The exogenous connections shared among $x_1$, $x_2$ and $x_3$ in the ground truth are explained here by two additional latent confounders $x_6$ and $x_7$. However, $x_2$, $x_4$, $x_5$ all have no distinct sources, hence deterministically depend on their direct causes and the latent confounder. Red arrows represent the differences.
}
\label{fig:example1}
\end{figure}

% Discussion about the hierarchical structure
On the other hand, it is easy to see that the ground truth in Figure \ref{fig:example1} can be modeled by our linear P-SCM, which allows for observed variables not to have exogenous connections, and explains ``latent confounding'' simply by ``shared'' exogenous connections from the sources (for instance, $s_2$ has exogenous connections to both $x_1$ and $x_2$). 

Lastly, it is worth mentioning that the generating model in this example has a hierarchical structure that entails practical significance: $x_1$, $x_2$ and $x_3$ are exogenously connected to different pairs of sources, while $x_4$ and $x_5$ are composed of different triplets of sources (through different pairs of causal connections from $x_1$, $x_2$ and $x_3$). We thus can treat $\{x_1, x_2,x_3\}$ as one layer, and $\{x_4,x_5\}$ as another layer below $\{x_1,x_2,x_3\}$. This hierarchical structure may arise naturally in influence propagation problems through networks of nodes. For instance, one layer of twitter users tweet news directly from the news sources (NY-times, Washington-Post, etc), then another layer of users retweet from the first layer, and so on. In conclusion, this example show that our linear P-SCM is more suited to model such a setting than a linear DS-P-SCM. 

%Lastly, to show the hierarchical structure of the generating model and its practical applications, notice that $x_1$, $x_2$ and $x_3$ are connected to different pairs of sources. Meanwhile, $x_4$ and $x_5$ are connected to different triplets of sources through different pairs of causal connections among $x_1$, $x_2$ and $x_3$. Therefore, this generating model reflects a hierarchical structure where $x_1$, $x_2$ and $x_3$ are in the same layer, and $x_4$, $x_5$ are in another layer below. This model can be used in analyzing influence propagation through networks, and we have shown that linear P-SCMs explain this model better than linear SCMs.

\end{example}

\begin{example} \label{example_2}
We begin with a ground truth graph and show that-- as in the previous example-- linear DS-P-SCMs with latent confounders fail to explain the model. Further, we show that linear DS-P-SCMs without latent confounders can explain the model, yet the recovery under this assumption results in misinterpreted edges (low accuracy). A linear P-SCM, on the other hand, explains the model and results in the recovery of a correct network.

Consider the graph on the left in Figure \ref{fig:example2}, which represents the true generating model. We have four observed variables $x_1,\cdots, x_4$ and four jointly independent sources $s_1,\cdots, s_4$. %Solid and dashed arrows represent the same type of connections as in Example \ref{example_1}.
Notice that $s_3$ is exogenously connected to two observed variables. To explain the model by a linear DS-P-SCM, a latent confounder $x_5$ must be added to the model as shown in the middle in Figure \ref{fig:example2} (to explain the shared sources), which leaves observed variable $x_3$ having no distinct sources. Therefore the ground truth cannot be explained by a DS-P-SCM with latent confounders. On the contrary, the ground truth can be explained by a linear P-SCM, which allows each observed variable to be exogenously connected to a subset of the sources.

Note that, since we have four observed variables and four sources, the model can be modeled by a linear DS-P-SCM without latent confounders (i.e., both models have the same joint distribution over the observed variables), where each observed variable is exogenously connected to one distinct source, and to this source only. This however leads to incorrect recovery as explained in the plot on the right in Figure \ref{fig:example2}. Specifically, in the true generating model, $s_3$ is exogenously connected to both $x_3$ and $x_4$, while $x_4$ has a distinct source $s_4$. Thus, assuming a linear DS-P-SCM as the generating model, $s_3$ will be the distinct source of $x_3$, and the exogenous connection from $s_3$ to $x_4$ will be interpreted as a causal connection from $x_3$ to $x_4$. This will in turn necessitate the erroneous causal connection from $x_1$ to $x_4$ to cancel out the causal effect of $x_1$ on $x_4$ through $x_3$. Notice here that these additional connections break the marginal independency between $x_1$ and $x_4$, therefore the recovered model is unfaithful to the true generating model. 

% Further, with the same observation, the ground-truth can be explained by a linear SCM without latent confounder, as shown in the right plot of of Figure \ref{fig:example2}. However, this representation includes additional connections among observed variables, and is no longer faithful to the generating model. Notice that under the assumption that each observed variable can only be exogenously connected to one distinct source, $s_3$ is connected to $x_3$, and 
% the direct connection from $s_3$ to $x_4$ in the generating model is replaced by a causal connection from $x_3$ to $x_4$. In particular, in order to cancel out the causal effect from $x_1$ to $x_4$ through $x_3$, an extra causal connection from $x_1$ to $x_4$ must be added to the SCM. This additional connection breaks the marginal independency between $x_1$ and $x_4$, which is misleading in causal recovery. % which does not make sense

\begin{figure}[t]
\centering

\begin{tikzpicture}[thick, scale=.4]
\foreach \place/\name in {{(-3,0)/x_1}, {(3,0)/x_2}, {(-1.5,-2.5)/x_3}, {(1.5,-2.5)/x_4}}
    \node[obsvar, label=center:$\name$] (\name) at \place {};
  \foreach \source/\dest in {x_1/x_3, x_2/x_4}
    \path[causal] (\source) edge (\dest);

\foreach \place/\name in {{(-3,2.5)/s_1}, {(0,2.5)/s_3}, {(3,2.5)/s_2}, {(1.5,-5)/s_4}}
    \node[source, label=center:$\name$] (\name) at \place {};
  \foreach \source/\dest in {s_1/x_1, s_2/x_2, s_3/x_3, s_4/x_4}
    \path[exogenous] (\source) edge (\dest);
  \foreach \source/\dest in {s_3/x_4}
    \path[difference] (\source) edge (\dest);

\end{tikzpicture}
\hspace{3em}
\begin{tikzpicture}[thick, scale=.4]
\foreach \place/\name in {{(-3,0)/x_1}, {(3,0)/x_2}, {(-1.5,-2.5)/x_3}, {(1.5,-2.5)/x_4}}
    \node[obsvar, label=center:$\name$] (\name) at \place {};
    \node[obsvar, dashed, label=center:$x_5$] (x_5) at (0,0) {};
  \foreach \source/\dest in {x_1/x_3, x_2/x_4}
    \path[causal] (\source) edge (\dest);

\foreach \place/\name in {{(-3,2.5)/s_1}, {(0,2.5)/s_3}, {(3,2.5)/s_2}, {(1.5,-5)/s_4}}
    \node[source, label=center:$\name$] (\name) at \place {};
  \foreach \source/\dest in {s_1/x_1, s_2/x_2, s_4/x_4}
    \path[exogenous] (\source) edge (\dest);
  \foreach \source/\dest in {s_3/x_5, x_5/x_4}
    \path[difference, dashed] (\source) edge (\dest);

\path[causal, dashed] (x_5) edge (x_3);

\end{tikzpicture}
\hspace{3em}
\begin{tikzpicture}[thick, scale=.4]
\foreach \place/\name in {{(-3,0)/x_1}, {(3,0)/x_2}, {(-1.5,-2.5)/x_3}, {(1.5,-2.5)/x_4}}
    \node[obsvar, label=center:$\name$] (\name) at \place {};
  \foreach \source/\dest in {x_1/x_3, x_2/x_4}
    \path[causal] (\source) edge (\dest);
  \foreach \source/\dest in {x_1/x_4, x_3/x_4}
    \path[difference] (\source) edge (\dest);

\foreach \place/\name in {{(-3,2.5)/s_1}, {(0,2.5)/s_3}, {(3,2.5)/s_2}, {(1.5,-5)/s_4}}
    \node[source, label=center:$\name$] (\name) at \place {};
  \foreach \source/\dest in {s_1/x_1, s_2/x_2, s_3/x_3, s_4/x_4}
    \path[exogenous] (\source) edge (\dest);

\end{tikzpicture}
\caption{(Left) The ground truth (the true generating model) for Example \ref{example_2}. (Middle) The ground truth can not be modeled as a linear DS-P-SCM with latent confounders: $s_3$ is exogenously connected to two observed variables $x_3$ and $x_4$. A latent confounder $x_5$ is added to the model, which leaves observed variable $x_3$ with no distinct sources. (Right) The ground truth is modeled incorrectly as a linear DS-P-SCM without latent confounders.}
\label{fig:example2}
\end{figure}
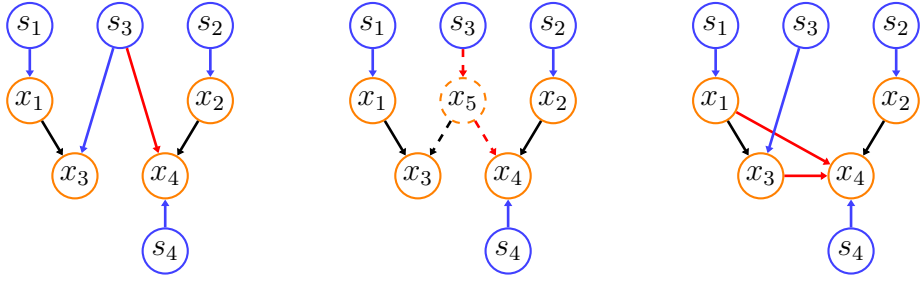
\end{example}

\subsection{Example explaining the possible parent set}\label{app:possible_parent_set_example}
\begin{example}
Suppose we have 3 observed variables $x_1,x_2,x_3$ and 4 source variables $s_1,s_2,s_3,s_4$. The generating model is given by $x_1=b_{11} s_1+ b_{12} s_2$; $x_2=b_{21} s_1 +b_{23} s_3$; $x_3=a_{31} x_1+b_{34}s_4$. The observed variables can be written as a linear combination of the sources as follows:
\begin{align}\nonumber
X =\bW S=
\begin{bmatrix}
b_{11} & b_{12} & 0 & 0 \\
b_{21} & 0 & b_{23} & 0 \\
a_{31}b_{11} & a_{31}b_{12} & 0 & b_{43} \\
\end{bmatrix}
\begin{bmatrix}
s_1 \\
s_2 \\
s_3 \\
s_4 \\
\end{bmatrix}.
\end{align}
The recovered mixing matrix $\tilde{\bW}$ is a column-permuted and rescaled version of $\bW$. Using $\tilde{\bW}$, one possible causal order of the observed variables is $(x_1,x_2,x_3)$, where 
one may further conclude that 
$x_1$ and $x_2$ have no parents and $x_3$ has $x_1$ as a parent. Note that both $x_1$ and $x_2$ precede $x_3$ in this causal order, 
hence both need to be considered when recovering the causal structure of $x_3$. 

However, the possible parent set of $x_3$ is $\mathcal{P}_3=\{x_1\}$ (as defined in Section \ref{sec:definition_conditions}): $x_2$ cannot be a parent of $x_3$ since $x_3$ does not have $s_3$ as a component. To conclude, the possible parent set of $x_3$, i.e., $\mathcal{P}_3$, is a strict subset of the observed variable preceding $x_3$ in the casual order deduced from $\tilde{\bW}$.
\end{example}

\subsection{Example explaining the definition of unique component}\label{app:example_unique_component}
\begin{example}
Consider a generating model with seven observed variables $x_1,\cdots,x_7$ and eight source variables $s_1,\cdots,s_8$. Let us fix $x_7$. Suppose all the other six variables are possible parents of $x_7$, i.e., $\mathcal{P}_7=\{x_1,\cdots, x_6\}$. The exogenous connections from the source variables to the observed variables in $\mathcal{P}_7$ are
\begin{align*}
\tilde{s}_1 &= b_{11}s_1 + b_{13}s_3 + b_{14}s_4 + b_{17}s_7; \\
\tilde{s}_2 &= b_{22}s_2 + b_{23}s_3 + b_{25}s_5; \\
\tilde{s}_3 &= b_{34}s_4 + b_{36}s_6; \\
\tilde{s}_4 &= b_{45}s_5 + b_{46}s_6 + b_{48}s_8; \\
\tilde{s}_5 &= b_{57} s_7 + b_{58} s_8; \\
\tilde{s}_6 &= b_{67} s_7 + b_{68} s_8. \\
\end{align*}
Let us first consider the possible parent set of $x_7$, i.e., $\mathcal{P}_7$. The index set of $\mathcal{P}_7$ is $J^{(0)}=\{1,\cdots,6\}$. According to the equations, $s_1$ is connected only to $x_1$, and $s_2$ is connected only to $x_2$. Thus, the unique component set of $x_1$ is $U_7(1)=\{s_1\}$, and the unique component set of $x_2$ is $U_7(2)=\{s_2\}$. The other observed variables in $\mathcal{P}_7$ do not have unique components. Thus, the subset of $\mathcal{P}_7$ with no unique components is $\mathcal{P}_7^{(1)}=\{x_3,x_4,x_5,x_6\}$. The corresponding index set is $J^{(1)} = \{3, 4, 5, 6\}$.

Next, we consider only the observed variables in $\mathcal{P}_7^{(1)}$ with indices in $J^{(1)}$. $s_4$ is connected only to $x_3$, and hence $U_7(3)=\{s_4\}$. Similarly, $U_7(4)=\{s_5\}$, and $J^{(2)} = \{5, 6\}$.

Finally, each of the observed variables with indices in $J^{(2)}$ ($x_5,x_6$) is connected to both sources $s_7$ and $s_8$. Thus, neither $x_5$ nor $x_6$ has a unique component, and the procedure terminates. To conclude, the set of observed variables with no unique components is $\mathcal{I}_7 = \{x_5, x_6\}$.
\end{example}

\subsection{Examples showing the necessity of the conditions} \label{app:example_conditions}
In this subsection, we provide examples on how Conditions \ref{condition:unique_component} and \ref{condition:marriage} are necessary for the unique identifiability of a linear P-SCM. More specifically, in Example \ref{example:unique_components} below, we show that, for a given observed variable $x_k$, if any source variable that is a unique component of a possible parent $x_i\in \setP_k$ has an exogenous connection to $x_k$, then we can not uniquely identify the total causal effect from $x_i$ to $x_k$ in the corresponding generating model. In other words, for unique identifiability of the causal effects to $x_k$, $\tilde{s}_k$ must not contain a unique component of any $x_i\in\mathcal{P}_k$. 

Notice that since we consider a fixed observed variable, $x_k$, and aim to recover/estimate the total causal effects from the observed variables in $\mathcal{P}_k$, we do not consider the causal connections among these possible parents. 

\begin{example} \label{example:unique_components}
Consider the following generating model:
\begin{align} \label{eq:example3_0}
\begin{bmatrix}
x_1 \\
x_2 \\
x_3 \\
\end{bmatrix}
 = 
\begin{bmatrix}
0 & 0 & 0 \\
0 & 0 & 0 \\
a_{31} & a_{32} & 0 \\
\end{bmatrix}
\begin{bmatrix}
x_1 \\
x_2 \\
x_3 \\
\end{bmatrix}
+
\begin{bmatrix}
b_{11} & b_{12} & 0 \\
b_{21} & 0 & b_{23} \\
0 & b_{32} & 0 \\
\end{bmatrix}
\begin{bmatrix}
s_1 \\
s_2 \\
s_3 \\
\end{bmatrix}
.
\end{align}
The observed variables can be represented as linear combinations of the sources as follows: 
\begin{align}
X = \bW S=
\begin{bmatrix}
b_{11} & b_{12} & 0 \\
b_{21} & 0 & b_{23} \\
a_{31}b_{11} & a_{31}b_{12}+ b_{32} & a_{32}b_{23} \\
\end{bmatrix}
\begin{bmatrix}
s_1 \\
s_2 \\
s_3 \\
\end{bmatrix}.
\label{eq:example3_1}
\end{align}
Given the observations $X=(x_1,x_2,x_3)^{\top}$, the mixing matrix $\bW$ in \eqref{eq:example3_1} can be recovered up to permutation and scaling indeterminacies as described in Appendix \ref{sec:application_to_our_model}. %In this example, we focus on the necessity of the following condition for unique identifiability of causal effects: {\emph{a given observed variable $x_k$ cannot be exogenously connected to a source variable which appears as a unique component in any of the possible parents of $x_k$}}. In order to abstract the idea, 
For simplicity, we assume that the recovered mixing matrix $\tilde{\bW}$ is the same as $\bW$. The intricacies resulting from permutation and scaling indeterminacies is resolved in our proposed P-SCM Recovery algorithm. 
 
We assume that the causal structure of $x_1$ and $x_2$ have been correctly recovered as in \eqref{eq:example3_0}, and consider observed variable $x_3$. From the blind source separation output, i.e., the recovered mixing matrix $\bW$ in \eqref{eq:example3_1}, we deduce that $x_1$ and $x_2$ are possible parents of $x_3$. %As we previously mentioned, we do not consider causal effects among the possible parents of $x_3$. That is, the sources in $x_1$ and $x_2$ result merely from exogenous connections.

We can rewrite the generating model of $x_3$ in \eqref{eq:example3_0} as follows:
\begin{equation}
x_3=a_{31} x_1 + a_{32} x_2 +\tilde{s}_3, \quad \text{where } \tilde{s}_3=b_{32} s_2.
\label{eq:example3_2}
\end{equation}
Notice that $s_2$ is a unique component of $x_1$. The $s_2$ component in $x_3$ (cf. \eqref{eq:example3_1}) consists of two parts: $a_{31} b_{12} s_2$ which results from a causal connection from $x_1$, and $b_{32} s_2$ which results from an exogenous connection from $s_2$. In order to learn the causal structure of $x_3$, we ought to be able to differentiate between these two parts from their mixture in the mixing matrix $\bW$, which is not possible in this example. Specifically, from the recovered mixing matrix $\tilde{\bW}=\bW$, we can not distinguish the true generating model in \eqref{eq:example3_2} from the two following generating models:
\begin{equation*}
\textbf{Model 1}: \quad x_3^{(1)} = \left(a_{31} + \frac{b_{32}}{b_{12}}\right) x_1 + a_{32} x_2 + \tilde{s}_3^{(1)},\quad \text {where } \tilde{s}_3^{(1)} = - \frac{b_{11}b_{32}}{b_{12}} s_1,
\label{eq:ex14}
\end{equation*}
which corresponds to the $s_2$ component in $x_3$ being caused only by a causal connection from $x_1$; and
\begin{equation*}
\textbf{Model 2}: \quad x_3^{(2)} = a_{32} x_2 + \tilde{s}_3^{(2)},\quad \text {where }\tilde{s}_3^{(2)} = a_{31}b_{11}s_1 + (a_{31}b_{12} + b_{32})s_2.
\label{eq:ex15}
\end{equation*}
which corresponds to the $s_2$ component being caused only by an exogenous connection from $s_2$.

Similarly, if the generating model in \eqref{eq:example3_2} is such that $\tilde{s}_3= b_{33} s_3$, the causal effects cannot be uniquely identified, since $s_3$ is a unique component for $x_2$. In conclusion, to uniquely recover the causal structure of $x_3$, neither $s_2$ nor $s_3$ can be included in $\tilde{s}_3$.

% Finally, if \eqref{eq:example3_2} is such that $\tilde{s}_3= b_{31} s_1$, we {\emph{can uniquely identify the casual connections to $x_3$}}. From \eqref{eq:example3_0}, the source $s_1$ is shared between the observed variables $x_1$ and $x_2$, each of which has a unique component. In particular, using the unique component of $x_1$, i.e., $s_2$, we can compute the causal effect from $x_1$ to $x_3$, i.e., $a_{31}$. Similarly, we can compute $a_{32}$ using $s_3$, the unique component of $x_2$. Subsequently, we are able to compute the exogenous connection from $s_1$ to $x_3$ by subtracting the part of $s_1$ which results from the causal connections from $x_1$ and $x_2$. 
\end{example}

Next, we present an example about the necessity of the marriage condition for unique identifiability a in linear P-SCM. Recall that the marriage condition is: For a given observed variable $x_k$ and its possible parent set $\mathcal{P}_k$, every subset of the possible parents $X_C\subseteq \mathcal{P}_k$ must include at least $|X_C|$ different source variables. %That is, for every $X_C \subseteq \mathcal{P}_k$, $|X_C|\leq \left|\bigcup_{x_i \in X_c} {\rm{Comp}}(x_i)\right|$. 

\begin{example}[Marriage condition] \label{ex:marriage_condition}
Consider the following generating model:
\begin{equation}
\begin{aligned}
x_1 &= \tilde{s}_1 =  b_{11} s_1 + b_{12} s_2;\qquad 
x_2 = \tilde{s}_2 = b_{21} s_1 + b_{23} s_3; \\
x_3 &= \tilde{s}_3 = b_{32} s_2 + b_{33} s_3;\qquad 
x_4 = \tilde{s}_4 = b_{42} s_2 + b_{44} s_4;\\
x_5 &= \tilde{s}_5 = b_{53} s_3 + b_{54} s_4;\qquad 
x_6 = a_{62} x_2 + a_{64} x_4 + a_{65} x_5 + \tilde{s}_6,\quad \tilde{s}_6 =b_{65} s_5.
\end{aligned}
\label{eq:exammple5_05}
\end{equation}
The observed variables can be written as follows: 
\begin{align}
\label{eq:exammple5_1}
X = \bW S=
\begin{bmatrix}
b_{11} & b_{12} & 0 & 0 & 0 \\
b_{21} & 0 & b_{23} & 0 & 0 \\
0 & b_{32} & b_{33} & 0 & 0 \\
0 & b_{42}  & 0 & b_{44} & 0 \\
0 & 0  & b_{53} & b_{54} & 0 \\
w_{61} & w_{62} & w_{63} & w_{64} & w_{65} \\
\end{bmatrix}
\begin{bmatrix}
s_1 \\
s_2 \\
s_3 \\
s_4 \\
s_5 \\
\end{bmatrix}.
\end{align}
We recover the mixing matrix $\bW$ using BSS. As in the previous example, we assume that $\tilde{\bW}=\bW$. Consider the observed variable $x_6$. From $\bW$, the possible parents of $x_6$ are $\mathcal{P}_6= \{x_1,\cdots,x_5\}$. Notice that the marriage condition is not satisfied in this example: $\left| \bigcup_{x_i \in \mathcal{P}_6}{\rm{Comp}}(x_i)\right|=4$ while $|\mathcal{P}_6|=5$, therefore, $|\mathcal{P}_6|\nleq \left|\bigcup_{x_i \in \mathcal{P}_6} {\rm{Comp}}(x_i)\right|$. In the following, we show how this renders the unique identifiability to be impossible. 

Consider $X_{[1:4]}=(x_1, x_2, x_3, x_4)^{\top}$. These observed variables only contain the source variables $S_{[1:4]}=(s_1,\cdots, s_4)^{\top}$, and thus can be written as $X_{[1:4]}=\bW_0 S_{[1:4]}$, where $\bW_0$ is the upper left $4\times 4$ submatrix of $\bW$ in \eqref{eq:exammple5_1}. $\bW_0$ is invertible due to P-SCM faithfulness assumption. Now, $x_5$ contains only the sources $(s_3,s_4)$, and hence can be written as a linear combination of $X_{[1:4]}$ as follows:
\begin{align}
x_5 = [0 \;0\;b_{53}\;b_{54}] \bW^{-1}_0 X_{[1:4]}
=\sum_{i=1}^4 c_{i} x_i,
\label{eq:exammple5_2}
\end{align}
where $c_i, i=1,\cdots,4$ can be calculated using the exogenous connections.
% Recall that we are interested in recovering the causal connections to the fixed observed variable $x_6$, and that we are not considering any causal connections among the possible parents of $x_6$. Yet, 
The fact that $x_5$ is linearly dependant on (i.e., can be represented as a linear combination of) $x_1,\cdots,x_4$ results in impossibility of learning the causal connections to $x_6$ from observations. Specifically, it is impossible to distinguish the true generating model of $x_6$ in \eqref{eq:exammple5_05} from the following model, in which $x_5$ in the true generating model is replaced by \eqref{eq:exammple5_2}:
\begin{equation}
x_6 = c_1 a_{65} x_1 + (c_2a_{65} + a_{62}) x_2 + c_3a_{65} x_3 + (c_4a_{65} + a_{64}) x_4 + \tilde{s}_6.
\end{equation}
Therefore, the generating model is not uniquely identifiable. 

As we have seen in this example, unique identifiability of causal effects to $x_6$ fails because, in the representation of $X$ recovered by BSS (cf. \eqref{eq:exammple5_1}),  $x_5$ is linearly dependent on the remaining possible parents of $x_6$, i.e., $X_{[1:4]}$.
In Appendix \ref{app:proof_marriage}, we show that linear independence among possible parents in $\mathcal{P}_k$ holds if and only if the marriage condition is satisfied.
\end{example}

\section{Proof of main theorem}\label{app:proof_main_theorem}
\subsection{Matrix representation of the necessary and sufficient conditions} \label{Matrix_representation}
Let us first define the existing component set $\mathcal{E}_k$ for each observed variable $x_k$ as follows.
\begin{definition} \label{def:existing_component}
The \emph{existing component set} of $x_k$, denoted by $\mathcal{E}_k$, is the set of all components in the possible parents of $x_k$, i.e.,
$\mathcal{E}_k = \bigcup_{x_i\in \mathcal{P}_k} {\rm{Comp}}(x_i)$.
% i.e., components introduced prior to $x_k$. %${\rm{Comp}}(x_k)$. [are not new]
%That is, $\mathcal{E}_k$ represents all the components in $x_k$ that are not introduced to $x_k$ for the first time (following the causal order).
\end{definition}

Recall that matrix $\bB$ in \eqref{eq:system_model_2} consists of all the exogenous connections from sources to observed variables. 
Let $\bB_k$ denote the submatrix of $\bB$ with rows corresponding to the possible parents of $x_k$, i.e., $\mathcal{P}_k$, and columns corresponding to the existing component set of $x_k$, i.e., $\mathcal{E}_k$. That is,
\begin{align}
    \bB_k = \left[\bB_{ij}\right]_{i,j:\;x_i\in \mathcal{P}_k,\;s_j\in \mathcal{E}_k}
\end{align}
We repermute the rows and columns of $\bB_k$, and label the columns, as follows:
\begin{enumerate}[(i)]
    \item We find the columns in $\bB_k$ with only one non-zero entry and label these columns as type-1. We repermute the columns of $\bB_k$ such that all type-1 columns are at the leftmost of $\bB_k$.
    
    \item We repermute the rows of $\bB_k$ such that all nonzero entries in type-1 columns appear in the upper rows. Note that two type-1 columns can have their non-zero entry in the same row. 
    
    \item Among the columns that are not type-1 (have two or more non-zero entries), we find those columns with their non-zero entries appearing in the non-zero part of type-1 columns (i.e., the upper non-zero rows). We label these columns as type-2. We repermute the columns of $\bB_k$ such that type-2 columns are next to type-1 columns. The matrix $\bB_k$ can be now written as
    \begin{align} \label{eq:matrix_representation_1}
    \bB_k = 
    \begin{bmatrix}
    \mathbf{U}_1 & \mathbf{X}_1 & \mathbf{Y}_1 \\
    \mathbf{0} & \mathbf{0} & \mathbf{Z}_1 \\
    \end{bmatrix}.    
    \end{align}
    $[\mathbf{U}_1; \mathbf{0}]$ are the type-1 columns, $[\mathbf{X}_1;\mathbf{0}]$ are the type-2 columns, and $[\mathbf{Y}_1;\mathbf{Z}_1]$ are the remaining columns. Each column of $\mathbf{Z}_1$ has at least one non-zero entry.
    \item We repeat steps (i) to (iii) to permute the columns and rows of $\mathbf{Z}_1$ such that
    \begin{align}
    \mathbf{Z}_1 = 
    \begin{bmatrix}
    \mathbf{U}_2 & \mathbf{X}_2 & \mathbf{Y}_2 \\
    \mathbf{0} & \mathbf{0} & \mathbf{Z}_2 \\
    \end{bmatrix}.    
    \end{align}
    The columns of $\bB_k$ corresponding to the columns of $\mathbf{U}_2$ and $\mathbf{X}_2$ (which were unlabeled in \eqref{eq:matrix_representation_1}) are now labeled as type-1 and type-2 respectively. 
    \item For $\mathbf{Z}_n$, $n=2,3,\cdots$, we repeat steps (i) to (iii) and label the corresponding columns in $\bB_k$ as in (iv); until this factorization does not hold for $\mathbf{Z}_N$ for some $N\in\mathbb{N}$. We label the columns of $\bB_k$ that correspond to the columns of  $\mathbf{Z}_N$ as type-3 columns.
\end{enumerate}

The column and row permutations in (i)-(v) are analogous to the iterative process in Definition \ref{def:unique}. Type-1 columns in (i) correspond to the unique components in the first iteration, i.e., among possible parents in $\mathcal{P}_k$. The rows selected in (ii) correspond to the possible parents in $\mathcal{P}_k\setminus \mathcal{P}_k^{(1)}$. Type-2 columns correspond to the sources that are shared only among the possible parents in $\mathcal{P}_k\setminus \mathcal{P}_k^{(1)}$. $\mathbf{Z}_1$ in \eqref{eq:matrix_representation_1} represents the exogenous connections of observed variables in $\mathcal{P}_k^{(1)}$. For the iteration in (i)-(v), $\mathbf{Z}_n$, $n=1,2,\cdots, N$, can be written as
\begin{equation*}
    \mathbf{Z}_n = [\bB_{ij}]_{i,j: \; x_i\in \mathcal{P}_k^{(n)},\; s_j\in \bigcup_{i\in J^{(n)}} {\rm{Comp}}(\tilde{s}_i)}.
\end{equation*}
Lastly, the rows in $\mathbf{Z}_N$ correspond to the possible parents in $\mathcal{I}_k$, and the sources shared among these variables correspond to type-3 columns.

Given the matrix permutation and labeling in (i)-(v), the unique components condition
translates to: Each observed variable $x_k$ can only be exogenously connected to either source variables corresponding to type-2 columns in $\bB_k$, or new source components that are not in $\mathcal{E}_k$ (i.e., do not correspond to columns in $\bB_k$). Further, the marriage condition translates to the matrix $\bB_k$ being of full row rank, which is equivalent to $\mathbf{Z}_N$ being of full row rank. The equivalence between the marriage condition and the full-row rank of $\bB_k$ is explained in Appendix \ref{app:proof_marriage}.

%\begin{remark} 
%The marriage condition on the collection of the component sets of all possible parents $\{{\rm{Comp}}(x_i)\}_{x_i\in \mathcal{P}_k}$ is equivalent to the marriage condition on the collection of the component sets of all source mixtures of possible parents $\{{\rm{Comp}}(\tilde{s}_i)\}_{x_i\in \mathcal{P}_k}$.
%\end{remark}

\subsection{Proof of sufficiency}

% \begin{algorithm}[tbp]
% \SetAlgoLined
% % \DontPrintSemicolon
% \KwIn{Recovered mixing matrix $\tilde{\bW}$ from BSS. \qquad \textbf{Initialize:} $\tilde{\bA} = \mathbf{I}$, $\bB=\mathbf{0}$.}
% Repermute $\tilde{\bW}$ such that the number of non-zero entries in each row is in an increasing order\;
% \For{$k=1:p$}{
% Find possible parent set $\mathcal{P}_k$; Select the row $W_k = \tilde{\bW}[k,:]$ \;
% Find the set $\mathcal{U}$ of possible parents in $\mathcal{P}_k$ that have unique components\;
% \eIf{$|\mathcal{U}| \neq 0$}{
% 	Calculate the total causal effect $\tilde{a}_{ki}$ for each $x_i\in \mathcal{U}$ using unique components\;
% 	% Remove the causal effect of $\tilde{s}_i$ from $x_k$:
% 	$W_k \leftarrow W_k - \sum_{i\in \mathcal{U}} \tilde{a}_{ki} \bB[i,:]$; $\mathcal{P}_k \leftarrow \mathcal{P}_k \setminus \mathcal{U}$;
% 	% Remove the possible parents in $U$ from possible parent set:  $. 
% 	Repeat from Step 4 on remaining $\mathcal{P}_k$ \;
% }{
%     Select the rows $\bB_I$ from $\bB$ corresponding to $\mathcal{P}_k$\;
%     Calculate $\tilde{a}_{ki}$ by solving a overdetermined linear system, 
%     using the non-zero columns of $\bB_I$ and the corresponding entries in $W_k$; Set the selected entries of $W_k$ as 0\;
% }
% $\bB[k,:] = W_k$\;
% }
% $\bA = \mathbf{I} - \tilde{\bA}^{-1}$\;
% \KwOut{Repermuted matrices $\bA$, $\bB$ according to the reversed order from Step 1.}
% \caption{P-SCM Recovery}
% \label{alg:recovery}
% \end{algorithm}

For any data generating model that satisfies Conditions \ref{condition:unique_component} and \ref{condition:marriage} in Sections \ref{sec:unique_component} and \ref{sec:marriage_condition}, we show that using the recovered mixing matrix $\tilde{\bW}$, we can construct an algorithm that uniquely recovers both the causal effects among observed variables and the exogenous connections from the sources. Specifically, we show that P-SCM Recovery Algorithm (Algorithm \ref{alg:recovery}) recovers the exact matrix of total causal effects, i.e., $(\bI-\bA)^{-1}$, and the exogenous connection matrix $\bB$ up to permutation and scaling, i.e., $\tilde{\bB}$ in \eqref{eq:B_tilde}. The row permutation in Step \ref{alg:permute} of Algorithm \ref{alg:recovery} has the property that for each observed variable $x_k$, the possible parents of $x_k$ are all preceding $x_k$ in the row ordering. This follows because for every $x_i\in\mathcal{P}_k$, ${\rm{Comp}}(x_i)\subsetneq{\rm{Comp}}(x_k)$ according to Definition \ref{def:pp_set}. Without loss of generality, assume that the ordering among observed variables after the permutation in Step \ref{alg:permute} is a natural ordering. In this case, we have $\mathcal{P}_k\subseteq \{x_1,\cdots,x_{k-1}\}$ for all $x_k$.

We provide the proof by induction on the index of the observed variables $k$, where the induction base is $k=1$. For $k=1$, there are no observed variables in $\mathcal{P}_1$, hence all the source components of $x_1$ must result from exogenous connections, which are learnt from the recovered mixing matrix. For the induction hypothesis, we assume that the total causal effects among $\{x_1,\cdots,x_{k-1}\}$ and their exogenous connections from the sources are given. We show that we can recover the total causal effects as well as the exogenous connections to $x_k$.

From the induction hypothesis, we know the exogenous connections (source mixtures) to the possible parents of $x_k$, i.e., $\tilde{\setS}_k = \{\tilde{s}_i:x_i\in \mathcal{P}_k\}$. Our algorithm follows a similar iterative procedure as in Definition \ref{def:unique}, in order to identify the unique components of each $x_i\in\mathcal{P}_k$, while simultaneously solve for the total causal effect from $x_i$ to $x_k$. 

First, the algorithm finds the observed variables with unique components among all possible parents in $\mathcal{P}_k$, i.e.,  $x_i\in \mathcal{P}_k\setminus \mathcal{P}_k^{(1)}$ (see Section \ref{sec:unique_component}). For each $x_i\in \mathcal{P}_k\setminus \mathcal{P}_k^{(1)}$ with a unique component $s_j$, it follows from the unique component condition that $s_j$ is not exogenously connected to $x_k$. Thus, the $s_j$ component in $x_k$ must result from the causal connection/path from $x_i$. The strength of the causal effect can be correctly recovered via dividing $s_j$ component in $x_k$ by $s_j$ component in $\tilde{s}_i$, i.e., $\tilde{w}_{kj}/\tilde{b}_{ij}$. Subsequently, we correctly recover the total causal effect from all $x_i\in \mathcal{P}_k\setminus \mathcal{P}_k^{(1)}$ to $x_k$.

\begin{remark} \label{remark:sufficiency_calculation}
Note that $s_j$ component in $x_i$ only results from an exogenous connection from $s_j$ to $x_i$, since the parents of $x_i$ are also in $\mathcal{P}_k$ and $s_j$ is a unique component. Hence the $s_j$ component in $\tilde{s}_i$ is the same as the $s_j$ component in $x_i$. This explains our use of ``dividing by $s_j$ coefficient in $x_i$'', instead of in $\tilde{s}_i$, in Section \ref{sec:unique_component}.
\end{remark}
 
Next, we find the total causal effects from $x_i\in\mathcal{P}_k^{(1)}$. Similarly, consider $x_i\in \mathcal{P}_k^{(1)}\setminus \mathcal{P}_k^{(2)}$ with a unique component $s_j$. It follows from the unique components condition that $s_j$ is not exogenously connected to $x_k$. Thus,  $s_j$ component in $x_k$ results from the following effects: (i) causal connection/path from $x_i$ and/or (ii) causal connections/paths from observed variables in $\mathcal{P}_k\setminus \mathcal{P}_k^{(1)}$. The latter causal effects are recovered in the last step and subtracted from $x_k$, and hence we calculate the total causal effect from $x_i$ to $x_k$ by the residual of $s_j$ component in $x_k$ divided by $s_j$ component in $\tilde{s}_i$. Therefore, we correctly recover the total causal effects from all $x_i\in \mathcal{P}_k^{(1)}\setminus \mathcal{P}_k^{(2)}$ to $x_k$.

%According to the first condition in Theorem \ref{thm:main}, $x_k$ cannot be exogenously connected to these unique components, then we can always correctly recover the total causal effects based on the residual of the components from previous groups.
Repeat the last step for $x_i\in \mathcal{P}_k^{(n)}\setminus \mathcal{P}_k^{(n+1)}$ for $n=2,\cdots, N-1$, i.e., all observed variables with unique components. Next, we find the total causal effects from  $x_i \in \mathcal{I}_k$ (with no unique components) to $x_k$. The unique components condition implies that $x_k$ cannot be exogenously connected to any source variable that  belongs to $\bigcup_{i\in J^{(N)}} {\rm{Comp}}(\tilde{s}_i)$ (recall that $J^{(N)}$ is the index set of $\mathcal{I}_k$). That is, the exogenous connections to $x_k$ can not overlap with the exogenous connections to any $x_i\in\mathcal{I}_k$. Then, we recover the total causal effects from $x_i\in\mathcal{I}_k$ to $x_k$ using the residuals of the set of sources $\bigcup_{i\in J^{(N)}} {\rm{Comp}}(\tilde{s}_i)$ in $x_k$ (after all subtractions in the previous steps). This results in an overdetermined linear system of the total causal effects:
\begin{align}
\label{eq:proof_sufficiency}
    \mathbf{w} = \mathbf{Z}^{\top}_N \tilde{\mathbf{a}}_k,\quad \mathbf{Z}_N = [\tilde{\bB}_{ij}]_{i,j:\;x_i\in \mathcal{I}_k,\; s_j\in \bigcup_{i\in J^{(N)}} {\rm{Comp}}(\tilde{s}_i) }.
\end{align}
$\mathbf{w}$ is the vector of the residuals of the source components in $x_k$ that correspond to $\bigcup_{i \in J^{(N)}} {\rm{Comp}}(\tilde{s}_i)$; $\tilde{\mathbf{a}}_k$ is the vector of total causal effects from the observed variables in $\mathcal{I}_k$ to $x_k$; $\mathbf{Z}_N$ represents the exogenous connections to observed variables in  $\mathcal{I}_k$ (see Appendix \ref{Matrix_representation}).

According to the marriage condition, $\mathbf{Z}_N$ is of full row rank, see Appendix \ref{Matrix_representation}. Thus, the overdetermined linear system in \eqref{eq:proof_sufficiency} has at most one solution which corresponds to the true total causal effects from each $x_i\in\mathcal{I}_k$. Note that $\tilde{\bB}_k$ has the same rank as $\bB_k$ since permuting the columns does not change the rank. %This completes our recovery of causal effects.
To sum up, Algorithm \ref{alg:recovery} correctly recovers the total causal effects from the possible parents of $x_k$ ($x_i\in\mathcal{P}_k$) to $x_k$. The remaining source components in $x_k$ represent the exogenous connections from the corresponding sources to $x_k$, which can be recovered up to permutation and scalability indeterminacies. The recovered matrix $\tilde{\bB}$ has the same column-permutation and scaling as the output of BSS $\tilde{\bW}$. That is, Algorithm \ref{alg:recovery} is able to recover the exact adjacency matrix $\bA$ using the unique components and/or matrix inversion, by shifting all the indeterminacies into the residual values to recover $\tilde{\bB}$. Recall, for matrix $\mathbf{B}$, the column permutation indeterminacy is irrelevant; the scalability indeterminacy is unavoidable in our setting unless we have prior information about the sources. 

The column permutation is maintained for $\tilde{\bB}$ because the order of the sources, given in $\tilde{\bW}$, is not changed throughout Algorithm \ref{alg:recovery}. 
% For scalability of the causal effects, note that our method for finding the total causal effect $\tilde{a}_{ki}$ (from $x_i$ to $x_k$), with or without unique component, is scale free (cf. Step \ref{alg:total_effect_unique} and Step \ref{alg:total_effect_nonunique}) in Algorithm \ref{alg:recovery}).
% , and the updates on $x_i$ (subtracting $\tilde{a}_{ki} \tilde{s}_i$ in Step 7 or setting as 0 in step 10) are both linear with respect to each source.
% In particular, by assuming that the exogenous connections from each source to all observed variables in $\mathcal{P}_k$ share the same scale as the corresponding column in $\tilde{\bW}$, Algorithm \ref{alg:recovery} first recover the correct $\tilde{a}_{ki}$ using the unique components for $x_i\in \mathcal{P}_{k} \setminus \mathcal{P}_{k}^{(1)}$. Then, each source in the residual after subtracting $\tilde{a}_{ki} \tilde{s}_i$ from $x_k$ (Step \ref{alg:subtraction}) will share the same scale as the corresponding column in $\tilde{\bW}$. We can use the same method to recover the correct $\tilde{a}_{ki}$ for the remaining possible parents. Lastly, the residual still shares the same scale as in $\tilde{\bW}$, which are then considered as the direct causal effects from exogenous sources to $x_k$. 
For scalability of the causal effects, 
% we prove by induction on the index $k$. When $k=1$, then all source components in $x_1$ must result from are exogenous connections, hence $\tilde{s}_1$ share the same scale as $x_1$.
suppose the learnt exogenous connections from a source $s_l$ to all $x_i\in\mathcal{P}_k$ (i.e., the column of $\tilde{\bB}_k$ corresponding to $s_l$) share the same scale as the corresponding column in $\tilde{\bW}$. Algorithm \ref{alg:recovery} first recovers the correct total causal effect $\tilde{a}_{ki}=\tilde{w}_{kj}/\tilde{b}_{ij}$ (cf. Step \ref{alg:total_effect_unique}) using the unique components for $x_i\in \mathcal{P}_{k} \setminus \mathcal{P}_{k}^{(1)}$. The residual of $s_l$ in $x_k$ after subtracting the effect of $x_i$ (Step \ref{alg:subtraction}), which is  $\tilde{w}_{kl}-\tilde{a}_{ki} \tilde{b}_{il}$, shares the same scale as $\tilde{w}_{kl}$. Similarly, for all $x_i$ with unique components, i.e., $x_i\in \setP_k\setminus \setI_k$, Algorithm \ref{alg:recovery} recovers the correct total causal effect $\tilde{a}_{ki}$, and the residual shares the same scale as $\tilde{w}_{kl}$. For $x_i\in \setI_k$, Algorithm \ref{alg:recovery} recovers the correct total causal effect using \eqref{eq:proof_sufficiency} (cf. Step \ref{alg:total_effect_nonunique}), since each column of $\mathbf{Z}_N$ in \eqref{eq:proof_sufficiency} shares the same scale as the corresponding entry of $\mathbf{w}$. Finally, the remaining $s_l$ component in $x_k$ shares the same scale as $\tilde{w}_{kl}$, which is then considered as the exogenous connection $\tilde{b}_{kl}$. Therefore, the learnt exogenous connections from $s_l$ to all observed variables $x_k$, which are in the column of $\tilde{\bB}$ corresponding to $s_l$, share the same scale as the corresponding column in $\tilde{\bW}$.

In conclusion, given the causal structure and exogenous connections to the possible parents of $x_k$, we are able to recover the exact total causal effects to $x_k$ along with its exogenous connections. This establishes the induction step and completes our sufficiency proof. 

\begin{remark}
As we mentioned earlier, for each observed variable $x_k$, the possible parents of $x_k$, i.e., $\mathcal{P}_k$, are all preceding $x_k$ in the row ordering deduced in Step \ref{alg:permute} of Algorithm \ref{alg:recovery}. Since the ancestors of $x_k$ are also included in $\mathcal{P}_k$, this implies that by sorting the observed variables according to the number of source components in the recovered mixing matrix $\tilde{\bW}$, we can find a causal order among the observed variables which is consistent with the correct order. This method is faster than the pairwise comparison in \citep[Lemma 5]{salehkaleybar2020learning}.
\end{remark}

\subsection{Proof of necessity} \label{app:proof_necessity}
We show that for any generating model $\mathcal{G}$ that violates either the unique component condition or the marriage condition, by using the corresponding $\tilde{\bW}$, it is not possible to distinguish between the true generating model $\mathcal{G}$ and another generating model $\mathcal{G}'$. Without loss of generality, assume that the observed variables are ordered such that for any $x_k$, $\mathcal{P}_k\subseteq \{x_1,\cdots,x_{k-1}\}$. This ordering can be achieved by permuting the rows of $\tilde{\bW}$ to the corresponding order.

% Before we start, we can re-permute the rows of observation matrix $\bm{W}$ such that the number of non-zero entries in each row, representing the size of ${\rm{Comp}}(x_k)$ for each $x_k$, is in an increasing order. Since for all $i,j$, we have
% \begin{equation*}
% x_i \text{ is a parent of } x_j \Longrightarrow |{\rm{Comp}}(x_i)| < |{\rm{Comp}}(x_j)|,
% \end{equation*}
% which comes from the faithfulness assumption, this ordering of rows is consistent with the topological ordering of $\mathcal{G}$. Without loss of generality, in the following we assume that the natural order is the one after this permutation.

Now, consider a generating model $\mathcal{G}$ that does not satisfy either or both of the conditions. Suppose $x_k$ has the smallest index among the observed variables such that the conditions are not satisfied. That is, the submodel $\{x_{1},\cdots,x_{k-1}\}$ does satisfy the unique component and marriage conditions.
We can write the generating model $\mathcal{G}$ of $x_k$ as:
\begin{equation}
x_k = \sum_{i: x_i\in \mathcal{P}_k} \tilde{a}_{k i} \tilde{s}_{i} + \tilde{s}_k,\quad \tilde{s}_k = \sum_{s_j\in {\rm{Comp}}(\tilde{s}_k) } b_{kj}s_j.
\label{eq:form}
\end{equation}
where we redefine $\tilde{a}_{ki}$ to be the total causal effect from observed variable $x_i$ to $x_k$, and $b_{kj}$ is the strength of the exogenous connection from $s_j$ to $x_k$. %We now show that, for the same recovered mixing matrix, we can always find another generating model $\mathcal{G}'$ for $x_k$ which is not identical to $\mathcal{G}$. That is, the total causal effects and exogenous connections to $x_k$ in $\mathcal{G}'$ are different from \eqref{eq:form}. 
Non-satisfiability of the theorem conditions can only happen when either of the following cases hold. 
%  we can find another model $\mathcal{G}'$ with the same generating model as $\mathcal{G}$ for nodes $x_i$, $i=1,\cdots,k-1$, but with a different representation of $\tilde{x}_k$ in the form of (\ref{eq:form}), by which we can get a different representation of $\tilde{x}_k$ in the form of (\ref{eq0}).
\begin{enumerate}[(i)]
\item $\exists x_{i_0}\in \mathcal{P}_k$ with a non-empty unique component set, such that {\it{at least one unique component}} of $x_{i_0}$ is exogenously connected to $x_k$.
\item $\exists x_{i_0}\in \mathcal{P}_k$ with an empty unique component set such that {\it{at least one component}} of $\tilde{s}_{i_0}$ is exogenously connected to $x_k$.
\item The collection of component sets of the source mixtures $\tilde{\setS}_k= \{ {\rm{Comp}}(\tilde{s}_i) :x_i\in\mathcal{P}_k\}$ does not satisfy the marriage condition.
\end{enumerate}

\textbf{Case (i). }
Suppose there exists an index $i_0$ such that $U_k(i_0)\cap {\rm{Comp}}(\tilde{s}_k) \neq \emptyset $. Let $s_{j_0}\in U_k(i_0)\cap {\rm{Comp}}(\tilde{s}_k)$, i.e., $s_{j_0}$ is a unique component of $x_{i_0}$ and is also exogenously connected to $x_k$. In the following, we show we cannot determine if the component $s_{j_0}$ in $x_k$ results from a causal connection/path or an exogenous connection to $x_k$, or both.

We first take out the terms that include $\tilde{s}_{i_0}$ and $s_{j_0}$ from $x_k$ in \eqref{eq:form}, and rewrite \eqref{eq:form} as
\begin{align}
x_k = \sum_{\substack{i:x_i\in \mathcal{P}_k\\ i\neq i_0}} \tilde{a}_{k i} \tilde{s}_{i} + \tilde{a}_{k i_0} \tilde{s}_{i_0} + \sum_{\substack{s_j\in {\rm{Comp}}(\tilde{s}_k)\\ j\neq j_0}} b_{kj} s_j + b_{kj_0} s_{j_0} ,\quad b_{kj_0}\neq 0.
\label{eq:case11}
\end{align}
Next, consider the exogenous connections to $x_{i_0}$ (i.e., $\tilde{s}_{i_0}$), which can be expressed as
\begin{align}
\tilde{s}_{i_0} =  \sum_{\substack{s_{j'}\in {\rm{Comp}}(\tilde{s}_{i_0})\\ j'\neq j_0}} b_{i_0j'} s_{j'} + b_{i_0 j_0}s_{j_0},\quad b_{i_0 j_0}\neq 0.
\label{eq:case12}
\end{align}
By substituting \eqref{eq:case12} in \eqref{eq:case11}, we get
\begin{align} 
x_k &= \sum_{\substack{i:x_i\in \mathcal{P}_k\\ i\neq i_0}} \tilde{a}_{k i} \tilde{s}_{i} + \sum_{\substack{s_j\in {\rm{Comp}}(\tilde{s}_k)\\ j\neq j_0}} b_{kj} s_j + b_{kj_0} s_{j_0} + \tilde{a}_{k i_0} \left(\sum_{\substack{s_{j'}\in {\rm{Comp}}(\tilde{s}_{i_0})\\ j'\neq j_0}} b_{i_0j'} s_{j'} + b_{i_0 j_0}s_{j_0} \right) \nonumber \\
\label{eq:case13} 
&= \sum_{\substack{i:x_i\in \mathcal{P}_k\\ i\neq i_0}} \tilde{a}_{k i} \tilde{s}_{i} + \sum_{\substack{s_j\in {\rm{Comp}}(\tilde{s}_k)\\ j\neq j_0}} b_{kj} s_j +\sum_{\substack{s_{j'}\in {\rm{Comp}}(\tilde{s}_{i_0})\\ j'\neq j_0}} \tilde{a}_{k i_0} b_{i_0j'} s_{j'} + (\tilde{a}_{k i_0} b_{i_0 j_0} +  b_{kj_0} ) s_{j_0}.
\end{align}
%Given the recovered mixing matrix $\tilde{\bW}$, $x_k$ can be written as a linear combination of sources. 
Suppose the total causal effects from all observed variables $x_i\in\mathcal{P}_k\setminus \{x_{i_0}\}$ are recovered as in \eqref{eq:case13}. Now, our task is to recover the total causal effect from $x_{i_0}$ to $x_k$. 

Let us consider the $s_{j_0}$ component in \eqref{eq:case13}. It consists of two parts: $\tilde{a}_{k i_0} b_{i_0 j_0}s_{j_0}$ which results from the causal connection/path from $x_{i_0}$, and $b_{k j_0} s_{j_0}$ which results from the exogenous connection from $s_{j_0}$. In order to learn the causal structure, we ought to be able to differentiate between these two parts from their mixture in \eqref{eq:case13}, which is not possible. More specifically, we can not distinguish the true model from the following two generating models:

{\it{Model 1:}}
\begin{align}
\nonumber&x_k^{(1)} = 
\sum_{\substack{i:x_i\in \mathcal{P}_k\\ i\neq i_0}} \tilde{a}_{k i} \tilde{s}_{i} + (\tilde{a}_{i_0}+ \frac{b_{kj_0}}{b_{i_0j_0}})\tilde{s}_{i_0} + \tilde{s}^{(1)}_k;\\
&\tilde{s}^{(1)}_k = \sum_{\substack{s_j\in {\rm{Comp}}(\tilde{s}_k)\\ j\neq j_0}} b_{kj} s_j - \sum_{\substack{s_{j'}\in {\rm{Comp}}(\tilde{s}_{i_0})\\ j'\neq j_0}} \frac{b_{kj_0}b_{i_0j'}}{b_{i_0j_0}} s_{j'},
\label{eq:case14}
\end{align}
which corresponds to the $s_{j_0}$ component in $x_k$ being caused only by the causal effect from $x_{i_0}$.

{\it{Model 2:}}
\begin{align}
\nonumber &x_k^{(2)} = \sum_{\substack{i:x_i\in \mathcal{P}_k\\ i\neq i_0}} \tilde{a}_{k i} \tilde{s}_{i} + \tilde{s}^{(2)}_k;\\
&\tilde{s}^{(2)}_k = \sum_{\substack{s_j\in {\rm{Comp}}(\tilde{s}_k)\\ j\neq j_0}} b_{kj} s_j + \sum_{\substack{s_{j'}\in {\rm{Comp}}(\tilde{s}_{i_0})\\ j'\neq j_0}}\tilde{a}_{k i_0} b_{i_0j'} s_{j'} + (\tilde{a}_{k i_0} b_{i_0 j_0} +  b_{kj_0} ) s_{j_0},
\label{eq:case15}
\end{align}
which corresponds to the $s_{j_0}$ component being caused only by the exogenous connection from $s_{j_0}$. 
In contrast, if the unique component $s_{j_0}$ is not exogenously connected to $x_k$, then we can use this unique component to compute the total causal effect from $x_{i_0}$ to $x_k$ (i.e., $\tilde{a}_{k i_0}$), which in turn can be used to determine if there exist exogenous connections from the remaining components in $\tilde{s}_{i_0}$ to $x_k$.

Notice that an observed variable $x_i$ may have multiple unique components in the unique component set $U_{k}(i)$. Suppose there exists another source $s_{l_0}$ which is a unique component of $x_{i_0}$ and is not exogenously connected to $x_k$, while the unique component $s_{j_0}$ of $x_{i_0}$ is exogenously connected to $x_k$. Since we have no prior information about which of the two unique components is not exogenously connected to $x_k$, there are two candidate models that we cannot distinguish between: In each of these two models, the total causal effect from $x_{i_0}$ to $x_k$ is calculated using one of the two unique components in $x_{i_0}$, and there is an extra exogenous connection from the other unique component. Thus none of the unique components in $U_k(i)$ can be exogenously connected to $x_k$.
Note that having multiple unique components for a single observed variable does not contradict the irreducibility assumption. This is because unique components are defined over the source mixtures of the possible parents of $x_k$, i.e., $\tilde{\setS}_k$, and irreducibility assumption is defined over the whole mixing matrix. 

%states that two sources will be merged if they only connect to one common source. 
% when we are recovering the causal structure of $x_k$, we are only considering the submatrix of $\tilde{\bW}$ corresponding to the observed variables in $\mathcal{P}_k$.

\textbf{Case (ii). } 
Suppose there exists an index $i_0$ such that ${\rm{Comp}}(\tilde{s}_{i_0})\cap {\rm{Comp}}(\tilde{s}_k) \neq \emptyset $. Let $s_{j_0}\in {\rm{Comp}}(\tilde{s}_{i_0})\cap {\rm{Comp}}(\tilde{s}_k)$, i.e., $s_{j_0}$ is exogenously connected to both $x_{i_0}$ and $x_k$, where $x_{i_0}$ does not have unique components.

Note that \eqref{eq:case11}-\eqref{eq:case13} hold for this case as well; \eqref{eq:case11}-\eqref{eq:case13} do not depend on whether $x_{i_0}$ has unique components. When $x_{i_0}$ has no unique components, and one of its components $s_{j_0}$ is exogenously connected to $x_k$, this results in a similar indistinguishably problem as in the previous case. In particular, we are not able to distinguish between the part of $s_{j_0}$ component in $x_k$ that results from the causal connection/path from $x_{i_0}$, and the part that results from the exogenous connection from $s_{j_0}$. Thus, given $\tilde{\bW}$, we can not distinguish the true model from the two models in \eqref{eq:case14}, \eqref{eq:case15}.

The only difference from case (i) is that $x_{i_0}$ does not have unique components. In this case, we cannot have any of the components in $\tilde{s}_{i_0}$ to be exogenously connected to $x_k$. In particular, all the shared (non-unique) components of $\tilde{s}_{i_0}$ are pivotal to recover the total causal effect from $x_{i_0}$ to $x_{k}$ and including any of them in $\tilde{s}_k$ will always result in indistinguishably of the generating model. 

% Notice that the analysis for the previous case applies here, with the only difference being thatsj0∈Comp( ̃si0)∩Comp( ̃sk).  In particular, (48)-(51) only depend on the fact thesj0∈Comp( ̃si0)∩Comp( ̃sk). That being said, we can only include a component which

\textbf{Case (iii). } 
Suppose the marriage condition is not satisfied for the collection $\tilde{\setS}_k$. Then as described in Appendix \ref{Matrix_representation}, the matrix $\bB_k$ is not of full row rank, where $\bB_k$ is defined as
% \begin{align}\nonumber
$\bB_k = [\bB_{ij}]_{i,j:\;x_i\in \mathcal{P}_k, \;s_j\in \mathcal{E}_k}$.
% \end{align}

Given the recovered mixing matrix $\tilde{\bW}$, suppose that the exogenous connections to $x_k$, i.e., $\tilde{s}_k$, are recovered as in \eqref{eq:form}. Then the remaining source components in $x_k$ that result from the causal connections/paths from the possible parents in $\mathcal{P}_k$ can be written as
\begin{align}\nonumber
\sum_{s_j\in {\rm{Comp}}(x_k)} \tilde{w}_{kj} s_j-\tilde{s}_k=\sum_{s_j\in \mathcal{E}_k} \tilde{w}'_{kj} s_j = \sum_{i: x_i\in \mathcal{P}_k}\tilde{a}_{k i} \tilde{s}_{i},
\end{align}
where $\{\tilde{w}_{kj}\}_{j:s_j\in {\rm{Comp}}(x_k)}$ are the entries corresponding to $x_k$ in $\tilde{\bW}$ (i.e., $x_k=\sum_{s_j\in {\rm{Comp}}(x_k)} \tilde{w}_{kj} s_j$). $\{\tilde{w}'_{kj}\}$ are also known, as they result from subtracting the components of $\tilde{s}_k$ (known by assumption) from $x_k$. Our task is to recover the total causal effect $\tilde{a}_{k i}$ from each possible parent $x_i\in \mathcal{P}_k$ to $x_k$, which translates to finding the solution of
\begin{align}
\sum_{i: x_i\in \mathcal{P}_k} \tilde{a}_{ki} \tilde{s}_{i} = \sum_{s_j\in \mathcal{E}_k} \tilde{w}'_{kj} s_j.
\label{eq:case31}
\end{align}
If we write each source mixture $\tilde{s}_i$ in terms of its source components, then \eqref{eq:case31} can be written as
\begin{align}
\sum_{i: x_i\in \mathcal{P}_k}\tilde{a}_{ki} \sum_{s_j\in \mathcal{E}_k} b_{ij} s_j = \sum_{ s_j\in \mathcal{E}_k} \tilde{w}'_{kj} s_j.
\label{eq:case32}
\end{align}
\eqref{eq:case32} must hold for each source component $s_j\in \mathcal{E}_k$, which means
\begin{align}
\sum_{i: x_i\in \mathcal{P}_k}\tilde{a}_{ki} b_{ij}  = \tilde{w}'_{kj},\quad \forall s_j\in \mathcal{E}_k \qquad 
\Longrightarrow \qquad \bB^{\top}_k \tilde{\mathbf{a}}_k = \tilde{\mathbf{w}}'_k,
\label{eq:case33}
\end{align}
where $\tilde{\mathbf{a}}_k = [\tilde{a}_{ki}]_{i:x_i\in\mathcal{P}_k}$, and $\tilde{\mathbf{w}}'_k = [\tilde{w}'_{kj}]_{j:s_j\in\mathcal{E}_k}$. Recall, $\bB^{\top}_k$ is not of full column rank, hence \eqref{eq:case33} has infinite number of non-zero solutions. Denote these solutions as $\{\tilde{\mathbf{a}}_k^{(l)}: l=1,2,\cdots\}$. For each $l$, we obtain a different generating model of $x_k$ as
\begin{align}
x_k = \sum_{i:x_i\in \mathcal{P}_k}  \tilde{a}_{ki}^{(l)} \tilde{s}_{i} + \tilde{s}_k.
\label{eq:case34}
\end{align}
We conclude that, given the mixing matrix, there exist infinite number of generating models that we cannot distinguish from one another, i.e., the true generating model cannot be uniquely identified.

\section{Other proofs}\label{app:other_proofs}
\subsection{Proof of equivalence between two representations of linear DS-P-SCM} \label{app:proof_equiv_nondeter}

In this subsection, we prove that linear DS-P-SCM can be equivalently written as a linear P-SCM or a linear DS-SCM. That is, the submodel of linear P-SCM under distinct source assumption is equivalent to the submodel of linear DS-SCM under linear latent confounding and jointly independent sources. 

First, we show that a linear DS-SCM with linear latent confounding and jointly independent sources can be written as a linear P-SCM with every observed variable associated with a distinct source. Denote the vector of observed variables as $X$. A linear DS-SCM with linear latent confounding and jointly independent sources can be written as 
\begin{equation}
X =\bA_{ol}X_{l} + \bA X + S_d = \bA X + [\bA_{ol}\quad \bI] 
\begin{bmatrix}
X_{l} \\
S_d
\end{bmatrix}.
\label{eq:scm_canonical}
\end{equation}
$X_l$ represents the vector of jointly independent latent confounders; $S_d$ represents the vector of jointly independent distinct sources; $\bA$ represents the causal connection among observed variables; $\bA_{ol}$ represents the linear latent confounding. Note that each column in $\bA_{ol}$ contains at least two non-zero entries because it is a confounder. Under acyclicity assumption, $\bA$ can be converted to a strictly lower triangular matrix following the causal order among observed variables.
Hence, \eqref{eq:scm_canonical} is equivalent to linear P-SCM in \eqref{eq:system_model_2}, where the adjacency matrix is $\bA$, the exogenous connection matrix is $\bB=[\bA_{ol}\;\; \bI]$, and $S=[X_l; S_d]$. Since $\bI$ is a submatrix of $\bB$, this means that each observed variable is associated with a distinct source, i.e., a source that is not shared with any other observed variables. 

%This means that, the canonical form of a linear SCM corresponds to a linear P-SCM, where the linear P-SCM includes the identity matrix $\bI$ as a submatrix of $\bB$. That is, each observed variable is associated with a distinct exogenous source that is not shared with any other observed variables in the system.

Next, we show by construction that a linear P-SCM with distinct sources can be written as a linear DS-SCM with linear latent confounding and jointly independent sources. Consider a linear P-SCM, $X = \bA X + \bB S$. A distinct source that is associated to only one observed variable corresponds to a column in $\bB$ with only one non-zero entry. Since each observed variable is associated with a distinct source, the number of these columns is exactly equal to the number of observed variables, and these columns can be permuted and scaled into an identity matrix. The remaining columns in $\bB$ must have two or more non-zero entries (same structure as $\bA_{ol}$), and hence each of these columns corresponds to a latent confounder where the confounding is linear. Thus, the linear P-SCM reduces to \eqref{eq:scm_canonical}. This concludes the proof. 
%Hence the linear P-SCM corresponds to the canonical form of a linear SCM with latent variables if and only if every column vector of $\bI$ can be found in the columns of $\bB$, which means that each observed variable of P-SCM has a distinct component.

% If each observed variable of P-SCM has an independent component, then the submatrix of $\bB$ that corresponds to these independent components must be identity. Thus we can set all the non-independent components as the exogenous sources of latent variables in linear SCM, and partition the matrix $\bB$ into the identity part and the remaining part which represents the connection from the latent variables to the observed variables, in the shape of \eqref{eq:scm_canonical}. Then the linear P-SCM corresponds to the canonical form of a linear SCM with latent variables. 

\subsection{Proof of equivalence between two representations of linear P-SCM} \label{app:proof_equiv_deter}
In this subsection, we prove that linear P-SCM (cf. \eqref{eq:system_model_2}) can be equivalently written as \eqref{eq:ds_scm_deterministic}, which is considered in most works studying linear DS-P-SCM. That is, instead of allowing an observed variable to have no distinct source, we can require the observed variable to have a distinct source but the source can have zero variance.

Notice that \eqref{eq:ds_scm_deterministic} can be equivalently written as
\begin{align}
X =\bA_{ol}X_{l} + \bA X + \bB_{oo}S_o,
\label{eq:scm_deterministic}
\end{align}
where $S_o$ is the vector of distinct sources with non-zero variance. $\bB_{oo}$ is composed of one-hot column vectors (i.e., binary vectors with exactly one non-zero entry) representing the correspondence between observed variables in $X$ and their associated non-zero distinct sources. If deterministic relation are not present (i.e., the distinct sources are all with non-zero variance), then $\bB_{oo}$ is identity, and \eqref{eq:ds_scm_deterministic} is the same as \eqref{eq:scm_deterministic}. If some of the distinct sources have zero variance, then there are less sources in $S_o$ than observed variables. In this case, we can always rewrite \eqref{eq:ds_scm_deterministic} as \eqref{eq:scm_deterministic} by removing the sources in $S_d$ with zero variance, and remove the corresponding column vectors in $\bI$. We can also rewrite \eqref{eq:scm_deterministic} as \eqref{eq:ds_scm_deterministic} by adding a distinct source with zero variance for each observed variable that is not associated with a non-zero distinct source in \eqref{eq:scm_deterministic}, and remove $\bB_{oo}$.

% Let us now consider the case when deterministic relations are present. From the canonical form of a linear DS-SCM, $\bA_{ll}$, $\bA_{lo}$ in \eqref{eq:scm_deterministic} are both zero matrices. Further, each latent variable must be associated with a source of non-zero variance, otherwise, the latent variable shall not exist, hence $\bB_{ll}$ is identity. In contrast, since deterministic relations are present, some observed variables are not associated with distinct sources: $\bB_{oo}$ has more rows than columns, and hence some of its rows are zero.

To show that \eqref{eq:scm_deterministic} reduces to a linear P-SCM, we have
\begin{equation}
X = \bA X + [\bA_{ol} \quad \bB_{oo}] 
\begin{bmatrix}
X_{l} \\
S_{o}
\end{bmatrix}.
\label{eq:scm_canonical_deterministic}
\end{equation}
Assuming the model is acyclic, $\bA$ can be permuted into a strictly lower triangular matrix, hence \eqref{eq:scm_canonical_deterministic} reduces to a linear P-SCM. %To conclude, a linear DS-P-SCM with zero-variance distinct sources can be written as a linear P-SCM.

Next, we show that a linear P-SCM can be written as \eqref{eq:scm_deterministic}. Consider the linear P-SCM, $X=\bA X+\bB S$, where some observed variables are not associated with distinct sources. Recall from Appendix \ref{app:proof_equiv_nondeter} that a distinct source that is associated with only one observed variable corresponds to a column in $\bB$ with one non-zero entry. However, the number of these columns is less than the number of observed variables, and hence these columns can not be permuted to identity. Thus, we can write $\bB$ as $[\bA_{ol} \;\; \bB_{oo}]$, where each column of $\bB_{oo}$ have exactly one non-zero entry, yet $\bB_{oo}$ have more rows than columns, and hence some of its rows are zero. Columns of $\bA_{ol}$ have two or more non-zero entries and still represent linear latent confounding. After we scale the non-zero entries in $\bB_{oo}$ to 1 (by changing the scales of the sources in $S$ corresponding to these columns), the linear P-SCM reduces to %a linear DS-P-SCM with deterministic relation as
\eqref{eq:scm_canonical_deterministic}, which can be further translated to \eqref{eq:scm_deterministic}.

%each column of $\bB$ with only one non-zero entry (a column of $\bB_{oo}$) corresponds to an exogenous source that is associated with only one observed variable, and each column of $\bB$ with two or more non-zero entries (a column of $\bA_{oo}$) correspond to a latent confounder. The observed variables without associated exogenous sources are considered as deterministic variables in the linear SCM.

\subsection{Proof of equivalence between two representations of the marriage condition} \label{app:proof_marriage}
% Remark \ref{rem:marriage_condition_equivalence}
We first show that the marriage condition, when defined over the collection $\{{\rm{Comp}}(x_i):x_i\in \mathcal{P}_k\}$, is equivalent to linear independence of the possible parents in $\mathcal{P}_k$ in the mixing matrix $\bW$ (cf. \eqref{eq:system_model_4}). Recall that, for each $x_i\in \mathcal{P}_k$, ${\rm{Comp}}(x_i)$ includes the source components with indices corresponding to the non-zero entries in $x_i$'s row of $\bW$. 

Consider only the rows of $\bW$ corresponding to possible parents in $\mathcal{P}_k$. The non-zero entries in these rows correspond to the existing source components, i.e., $\mathcal{E}_k=\bigcup_{x_i\in \mathcal{P}_k}{\rm{Comp}}(x_i)$ (cf. Definition \ref{def:existing_component}). Let $X_k= [x_i]_{x_i\in \mathcal{P}_k}$ and $S_k=[s_j]_{s_j\in \mathcal{E}_k}$. Then we have
\begin{align}
X_k = \bW_k S_k,\quad \text {where } \bW_k = \left[\bW_{ij}\right]_{i,j:\;x_i\in \mathcal{P}_k,\;s_j\in \mathcal{E}_k}.
\label{eq:marriage_condition_rank_1}
\end{align}

% Let $\bB_p$ be an arbitrarily permuted version of matrix $\bB$ (either row or column permutations). Any submatrix of $\bB_p$ (with non-zero rows or columns) is of full rank.  

The following lemma shows the connection between the marriage condition and the rank of the submatrix $\bW_k$.
\begin{lemma} [\citet{edmonds1967systems}] \label{prop:marriage_condition_rank}
Let $\bW$ be an $m\times n$ matrix, where $m\leq n$, such that any submatrix (with non-zero rows or columns) of an arbitrarily permuted version of $\bW$ is of full rank. 
%any submatrix of $\bW_p$ with non-zero rows or columns is of full rank, where $\bW_p$ be an arbitrarily permuted version of matrix $\bW$. 
%no non-zero entry can be written as a linear combination of the other non-zero entries in $\bW$.
Then, $\bW$ has rank $m$ if and only if for every subset of $C$ rows, the corresponding submatrix of $\bW$ has at least $C$ non-zero columns. 
\end{lemma}
According to Lemma \ref{prop:marriage_condition_rank}, the marriage condition on the collection $\{{\rm{Comp}}(x_i):x_i\in \mathcal{P}_k\}$ is equivalent to the matrix $\bW_k$ in \eqref{eq:marriage_condition_rank_1} being of full row rank. That is, by applying Lemma \ref{prop:marriage_condition_rank} to $\bW_k$, for $\bW_k$ to be of full rank, every subset of $C$ rows of $W_k$ (i.e., every $C$ possible parents) must have at least $C$ non-zero columns (i.e., $C$ different source components). $\bW_k$ is full rank means that each possible parent is linearly independent from the other possible parents in $\mathcal{P}_k$. 

Note that the recovered mixing matrix $\tilde{\bW}$ from BSS is a column-repermuted and rescaled version of $\bW$, which preserves the rank of the submatrices of $\bW$. Thus the marriage condition is also equivalent to the linear independency among possible parents in the representation from BSS (as stated in Example \ref{ex:marriage_condition}). 

Similarly, for each possible parent $x_i\in \mathcal{P}_k$, ${\rm{Comp}}(\tilde{s}_i)$ includes the source components with indices corresponding to the non-zero entries in $x_i$'s row of $\bB$. Since each $x_i\in\mathcal{P}_k$ can be written as a linear combination of the source mixtures $\tilde{s}$ in its parent set (plus $\tilde{s}_i$), all of which are included in $\tilde{\setS}_k=\{\tilde{s}_i: x_i\in\mathcal{P}_k\}$, we have $\mathcal{E}_k=\bigcup_{x_i\in \mathcal{P}_k}{\rm{Comp}}(x_i) =\bigcup_{\tilde{s}_i\in \tilde{\setS}_k}{\rm{Comp}}(\tilde{s}_i)$. Therefore, similar to \eqref{eq:marriage_condition_rank_1}, by considering only the rows of $\bB$ that correspond to the possible parents in $\mathcal{P}_k$, the non-zero entries in these rows correspond to $\mathcal{E}_k$.
%This means that the collection of the indices of the non-zero entries in the rows of $\bB$ corresponding to $\mathcal{P}_k$ also correspond to the existing source components in $\mathcal{E}_k$. Hence, we extract the submatrix $\bB_k$ of $\bB$ as:
\begin{align}
    \bB_k = \left[\bB_{ij}\right]_{i,j:\; x_i\in \mathcal{P}_k,\;s_j\in \mathcal{E}_k}.
\end{align}
Once again, applying Lemma \ref{prop:marriage_condition_rank} to matrix $\bB_k$, the marriage condition on the collection $\{{\rm{Comp}}(\tilde{s}_i):\tilde{s}_i\in \tilde{\setS}_k\}$ is equivalent to the full row rank of $\bB_k$.

Note that the parents of $x_i\in\mathcal{P}_k$ are also possible parents of $x_k$ (belong to $\mathcal{P}_k$). Thus, the rows of $\bA$ that correspond to possible parents of $x_k$ have non-zero entries only in the columns of $\bA$ that also correspond to the possible parents of $x_k$. Therefore, $X_k$ can be written as
%Thus, the indices of nonzero entries in the row of $\bA$ corresponding to $x_i$, are included in the columns corresponding to the possible parents in $\mathcal{P}_k$. Then, the generating model of $X_k$ can be written as
\begin{align}
X_k = \bA_k X_k + \bB_k S_k,
\label{eq:marriage_condition_rank_2}
\end{align}
where $\bA_k = \left[\bA_{ij}\right]_{i,j:\;x_i\in \mathcal{P}_k,\;x_j\in \mathcal{P}_k}$ is the submatrix of $\bA$ with both rows and columns corresponding to possible parents of $x_k$. Comparing \eqref{eq:marriage_condition_rank_1} and \eqref{eq:marriage_condition_rank_2}, we have
\begin{align}
\bW_k = (\bI - \bA_k)^{-1} \bB_k.
\label{eq:marriage_condition_rank_3}
\end{align}
Since $\bA_k$ can be permuted to a strictly lower triangular matrix, $\bI - \bA_k$ is of full rank. Thus $\bW_k$ is of full row rank if and only if $\bB_k$ is of full row rank. To conclude, the marriage condition defined over the collection of the component sets of possible parents, is equivalent to the condition over the collection of the component sets of their source mixtures.

\subsection{Proof of Theorem 6} \label{app:proof_simplify} %\ref{thm:simplify}
We have mentioned in Section \ref{sec:simplify} that if each observed variable in the linear P-SCM is associated with a distinct source, then for each observed variable $x_k$, the possible parent set is exactly the ancestor set of $x_k$, and every possible parent $x_i\in \mathcal{P}_k$ has a unique component (i.e., $U_k(i)\neq \emptyset$). Besides, for each observed variable $x_i$, the component set of its mixtures, i.e. ${\rm{Comp}}(\tilde{s}_i)$, includes a distinct source. Thus, for any subset $X_C$ of the observed variables in $\setX$, $ \left| \bigcup_{i:x_i\in X_C} {\rm{Comp}}(\tilde{s}_i) \right|\geq |X_C| $. This implies that the marriage condition (Condition \ref{condition:marriage}) holds for every observed variable $x_k$, where we select the collection $X_C$ as any subset of the possible parent set $\mathcal{P}_k$.

Therefore, a reduced linear P-SCM could be uniquely identified from the recovered mixing matrix $\tilde{\bW}$ if and only if for each observed variable $x_k$, the unique components condition (Condition \ref{condition:unique_component}) is satisfied: For every exogenous connection from $s_j$ to $x_k$, it must belong to one of the two following cases:
\begin{itemize}
\item $s_j$ is a new component that is not exogenously connected to any of the possible parents of $x_k$. This means that no observed variables containing $s_j$ are possible parents of $x_k$, thus there are no causal paths from any observed variable containing $s_j$ to $x_k$.

\item $s_j $ is connected to some possible parents in $\mathcal{P}_k$. Condition \ref{condition:unique_component} implies that $s_j$ cannot be a unique component of any $x_i\in\mathcal{P}_k$, and it must be exogenously connected to at least two observed variables $x_{i_1}$, $x_{i_2}$ in $\mathcal{P}_k$, where $i_1\neq i_2$. Since $\mathcal{P}_k$ is the ancestor set of $x_k$, there are at least two distinct causal paths $x_{i_1} \leadsto x_k $ and $x_{i_2} \leadsto x_k $ from the observed variables with $s_j$ in their source mixtures (i.e., exogenously connected to $s_j$) to $x_k$.

\end{itemize}

\begin{remark}
In a graphical representation of the reduced linear P-SCM, the condition in Theorem \ref{thm:simplify} can be expressed as follows. Let us define a causal cycle as an undirected cycle in the graph, where one directed edge on this cycle is from a source variable $s_j$ to an observed variable $x_k$, and the remaining edges of the cycle form a directed path from $s_j$ to $x_k$. We call the source $s_j$ and the observed variable $x_k$ as the source and sink of the causal cycle. Then the reduced linear P-SCM is uniquely identifiable if and only if for each causal cycle in the graph, there exists at least one other causal cycle that shares the same source and sink (including the exogenous connection between them).

A simpler representation of the condition (which is not generally necessary and sufficient) is that: The generating model is not uniquely identifiable if the graphical representation includes a causal cycle, and none of the edges on this cycle is shared with other cycles in the graph. 
% An example is shown in the middle and right plots of Figure \ref{fig:example_amir}, where in each plot, the triangle consisting of a source and two observed variables forms a causal cycle. Since the triangles are the only cycles in each plot, none of its edges is shared, hence the corresponding linear P-SCMs are not uniquely identifiable. 
\end{remark}

\subsection{Proof of Remark 7} \label{app:amirs_work} %\ref{remark:relation_to_amirs_work}
For the sake of completeness, we state the following theorem.
\begin{lemma} [Theorem 16, \citet{salehkaleybar2020learning}] \label{lemma:amir}
Let $des_o(x_i)$ be the observed descendant set of variable $x_i$ in a linear DS-P-SCM with latent confounders (including $x_i$ itself if it is observed). Then the linear DS-P-SCM is uniquely identifiable from observations (i.e., the total causal effect between any two observed variables can be identified), if for any observed variable $x_i$ and any latent variable $x_k$, $des_o(x_i)\neq des_o(x_k)$.
\end{lemma}

As we explained in Appendix \ref{app:proof_equiv_nondeter}, a latent confounder in the linear DS-P-SCM corresponds to a shared source that is connected to at least two observed variables in the representation of linear P-SCM (with distinct source). In the following, we first show that the condition in Theorem \ref{thm:simplify} implies the condition in Lemma \ref{lemma:amir}. Note that the observed descendant set of a latent confounder and that of an observed variable must be different if both variables are not causally connected. Therefore, it suffices to show that if the condition in Theorem \ref{thm:simplify} is satisfied, then for a source $s_j$ that is exogenously connected to at least two observed variables including $x_k$, $des_o(s_j)\neq des_o(x_k)$.

According to Theorem \ref{thm:simplify}, since $s_j$ is exogenously connected to $x_k$, one of the following situations must hold:
\begin{itemize}
    \item There are at least two distinct causal paths from observed variables (which are exogenously connected to $s_j$) to $x_k$. Suppose $x_i$ is an ancestor of $x_k$ and is exogenously connected to $s_j$. Then we have $x_i\in des_o(s_j)$, and $x_i\not\in des_o(x_k)$. Thus $des_o(s_j)\neq des_o(x_k)$.
    
    \item There are no causal paths from observed variables containing $s_j$ to $x_k$. Recall that $s_j$ is exogenously connected to at least one observed variable other than $x_k$. Denote $x_l$ as the observed variable exogenously connected to $s_j$ that has the smallest index in the causal order among observed variables (other than $x_k$). That is, all observed variables that are exogenously connected to $s_j$ must follow $x_l$ in the causal order except for $x_k$. In this case we have $x_l\in des_o(s_j)$. In the following we show that $x_l\not\in des_o(x_k)$, which means that $des_o(s_j)\neq des_o(x_k)$, and completes the proof.
    
    The proof follows by contradiction. Suppose $x_l \in des_o(x_k)$. This means that there is a causal path from $x_k$ to $x_l$. Now consider $x_l$. Since $s_j$ is connected to both $x_k$ and $x_l$, according to Theorem \ref{thm:simplify}, there must exist another causal path from an observed variable (exogenously connected to $s_j$) to $x_l$. However, such an observed variable must precede $x_l$ in the causal order, which contradicts the fact that $x_l$ has the smallest index among all observed variables connected to $s_j$ other than $x_k$. Therefore $x_l\not\in des_o(x_k)$.
    
\end{itemize}

To show the second part of the remark, i.e., the condition in Lemma \ref{lemma:amir} does not imply the condition in Theorem \ref{thm:simplify}, we provide an example which satisfies the condition in Lemma \ref{lemma:amir} but does not satisfy the condition in Theorem \ref{thm:simplify}. Hence, the generating model in this example is uniquely identifiable among all linear DS-P-SCMs, but is not uniquely identifiable among all linear P-SCMs.

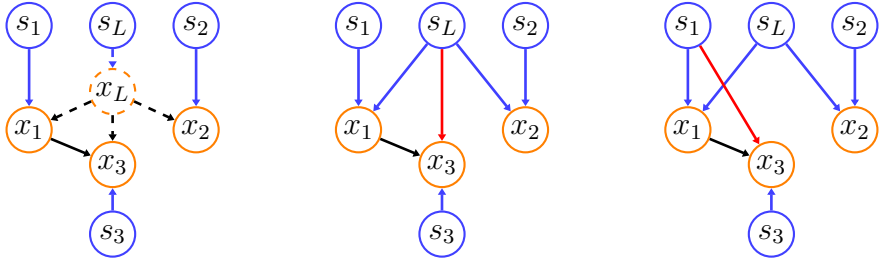
\begin{figure}[t]
\centering

\begin{tikzpicture}[thick, scale=.37]
\node[obsvar, dashed, label=center:$x_L$] (x_L) at (0,0.125) {};
\foreach \place/\name in {{(-3,-1.25)/x_1}, {(3,-1.25)/x_2}, {(0,-2.5)/x_3}}
    \node[obsvar, label=center:$\name$] (\name) at \place {};
  \foreach \source/\dest in {x_1/x_3}
    \path[causal] (\source) edge (\dest);
  \foreach \source/\dest in {x_L/x_1, x_L/x_2, x_L/x_3}
    \path[causal] (\source) edge[dashed] (\dest);

\foreach \place/\name in {{(-3,2.5)/s_1}, {(0,2.5)/s_L}, {(3,2.5)/s_2}, {(0,-5)/s_3}}
    \node[source, label=center:$\name$] (\name) at \place {};
  \foreach \source/\dest in {s_1/x_1, s_2/x_2, s_3/x_3}
    \path[exogenous] (\source) edge (\dest);
  \foreach \source/\dest in {s_L/x_L}
    \path[exogenous, dashed] (\source) edge (\dest);
%   \foreach \source/\dest in {s_3/x_4}
    % \path[difference] (\source) edge (\dest);

\end{tikzpicture}
\hspace{3em}
\begin{tikzpicture}[thick, scale=.37]
\foreach \place/\name in {{(-3,-1.25)/x_1}, {(3,-1.25)/x_2}, {(0,-2.5)/x_3}}
    \node[obsvar, label=center:$\name$] (\name) at \place {};
  \foreach \source/\dest in {x_1/x_3}
    \path[causal] (\source) edge (\dest);

\foreach \place/\name in {{(-3,2.5)/s_1}, {(0,2.5)/s_L}, {(3,2.5)/s_2}, {(0,-5)/s_3}}
    \node[source, label=center:$\name$] (\name) at \place {};
  \foreach \source/\dest in {s_1/x_1, s_2/x_2, s_3/x_3, s_L/x_1, s_L/x_2}
    \path[exogenous] (\source) edge (\dest);
  \foreach \source/\dest in {s_L/x_3}
    \path[difference] (\source) edge (\dest);

\end{tikzpicture}
\hspace{3em}
\begin{tikzpicture}[thick, scale=.37]
\foreach \place/\name in {{(-3,-1.25)/x_1}, {(3,-1.25)/x_2}, {(0,-2.5)/x_3}}
    \node[obsvar, label=center:$\name$] (\name) at \place {};
  \foreach \source/\dest in {x_1/x_3}
    \path[causal] (\source) edge (\dest);

\foreach \place/\name in {{(-3,2.5)/s_1}, {(0,2.5)/s_L}, {(3,2.5)/s_2}, {(0,-5)/s_3}}
    \node[source, label=center:$\name$] (\name) at \place {};
  \foreach \source/\dest in {s_1/x_1, s_2/x_2, s_3/x_3, s_L/x_1, s_L/x_2}
    \path[exogenous] (\source) edge (\dest);
  \foreach \source/\dest in {s_1/x_3}
    \path[difference] (\source) edge (\dest);

\end{tikzpicture}
\caption{(Left) The ground truth linear DS-P-SCM for Example \ref{example_amir}. This DS-P-SCM satisfies the condition in Lemma \ref{lemma:amir}. (Middle) The ground truth can be equivalently modeled as a linear P-SCM. This P-SCM does not satisfy the condition in Theorem \ref{thm:simplify}. (Right) An observationally equivalent linear P-SCM, which can not be modeled as a linear DS-P-SCM. Red arrows represent the differences.}
\label{fig:example_amir}
\end{figure}

\begin{example} \label{example_amir}
Consider the following linear DS-P-SCM with three observed variables $x_1$, $x_2$, $x_3$ and a latent confounder $x_L$:
\begin{equation}
\begin{aligned}
&x_L = s_L; \quad x_1 = a_{1L}x_L + s_1; \\
&x_2=a_{2L}x_L + s_2;\quad x_3 = a_{3L} x_L + a_{31} x_1 + s_3.
\end{aligned}
\label{eq:example_amir_scm}
\end{equation}
The graphical representation of this DS-P-SCM is shown in the plot on the left in Figure \ref{fig:example_amir}. The observed descendant sets of all four variables are
\begin{align}\nonumber
&des_o(x_L)=\{x_1, x_2, x_3\}; \quad des_o(x_1)=\{x_1, x_3\}; \\
&des_o(x_2)=\{x_2\}; \quad des_o(x_3)=\{x_3\}. \nonumber
\end{align}
Therefore, all four variables have distinct observed descendant sets, which satisfies the condition in Lemma \ref{lemma:amir}. Hence this linear DS-P-SCM is uniquely identifiable.

The linear DS-P-SCM in \eqref{eq:example_amir_scm} can be equivalently written into the following linear P-SCM:
\begin{equation}
\begin{aligned}
&x_1 = \tilde{s}_1 = a_{1L}s_L + s_1;\quad x_2=\tilde{s}_2=a_{2L}s_L + s_2;\\
&x_3 = a_{31} x_1 + \tilde{s}_3,\; \tilde{s}_3 = a_{3L} s_L + s_3.
\end{aligned}
\label{eq:example_amir_CPM}
\end{equation}
The graphical representation of this linear P-SCM is shown in the plot in the middle in Figure \ref{fig:example_amir}. This model violates the condition in Theorem \ref{thm:simplify}: $s_L$ is exogenously connected to $x_3$, but there is only one other path from an observed variable containing $s_L$ to $x_3$ (i.e., $x_1\rightarrow x_3$). Thus, this linear P-SCM is not uniquely identifiable, and an alternative linear P-SCM is shown in the plot on the right in Figure \ref{fig:example_amir}. The generating models of $x_1$ and $x_2$ in this alternative P-SCM are the same as in \eqref{eq:example_amir_CPM}, but the generating model of $x_3$ is
\begin{equation*}
x_3 = (a_{31}+a_{3L}/a_{1L}) x_1 + \tilde{s}_3;\quad \tilde{s}_3 = -(a_{3L}/a_{1L})s_1 + s_3.
\end{equation*}
Notice that this linear P-SCM cannot be represented as a linear DS-P-SCM, since there is no distinct source associated with $x_1$. Hence it does not affect the unique identifiability of the generating model in the class of linear DS-P-SCMs.

\end{example}

\section{Numerical experiments details} \label{app:simulations}
% We first provide estimates for the likelihood of the satisfiability of the necessary and sufficient conditions in Theorem \ref{thm:main}. Specifically, we randomly generate linear P-SCMs with fixed average degrees among the observed variables and source variables. We demonstrate that our conditions are likely to be satisfied when the number of source variables is large and the average degrees are small. Next, we evaluate the performance of our algorithm, Algorithm \ref{alg:recovery}, for P-SCM recovery. We compare our method with ICA-LiNGAM, Direct LiNGAM, ParceLiNGAM and Pairwise lvLiNGAM. Our algorithm outperforms these methods in recovering linear P-SCMs, even when the necessary and sufficient conditions are not satisfied. Moreover, when the model is separable (the recovered mixing matrix $\tilde{\bW}$ from BSS is highly accurate), our algorithm learns both the causal effects and the exogenous connections with high accuracy. For real data simulations, we consider the closing prices of five world stock indices, and model the corresponding returns as a linear P-SCM. In this case our algorithm can return a reasonable model with respect to the common belief and the previous works.

We generate synthetic models according to a linear P-SCM with $p$ observed variables and $rp$ source variables, where $r$ is the ratio of source to observed variables.\footnote{We use the words ``variable'' and ``node'' interchangeably, to refer to an observed/source variable.} Each pair of observed variables is connected with probability $d_e/(p-1)$, and each pair of source and observed variable is connected with probability $d_o/p$. $d_e$ (resp. $d_o$) is the average number of causal (resp. exogenous) connections for an observed (resp. a source) variable. That is, on average, each observed variable is causally connected to $d_e$ other observed variables, and each source variable is exogenously connected to $d_o$ observed variables. We refer to $r$, $d_e$, and $d_o$ as the source to observed node ratio, average observed node degree, and average source node degree, respectively. The strength of each connection is drawn independently from a uniform distribution over the interval $[-1, -0.5]\cup [0.5, 1]$. We randomly permute both the observed and source variables to hide the true causal order after model generation.

\subsection{Satisfiability of the conditions}\label{app:simulation_satisfiability}

To test the restrictiveness of Conditions \ref{condition:unique_component} and \ref{condition:marriage}, i.e., the unique component and marriage conditions in Sections \ref{sec:unique_component} and \ref{sec:marriage_condition}, we propose Algorithm \ref{alg:satisfy} which verifies whether the generating model satisfies these conditions. Given matrices $\bA$ and $\bB$ of a linear P-SCM, Algorithm \ref{alg:satisfy} first computes the true mixing matrix $\bW$, and use it to retrieve the possible parent set for each observed variable $x_k$. Subsequently, the algorithm implements the same iterative procedure as in Definition \ref{def:unique} to verify that the unique components condition is satisfied. 

\begin{algorithm}[t]
\SetAlgoVlined
\LinesNumbered
% \DontPrintSemicolon
\KwIn{Adjacency matrix $\bA$, and exogenous connection matrix $\bB$ of a linear P-SCM}
% Repermute $\bA$, $\bB$ following the causal order among observed variables \;
Compute $\bW = (\bI-\bA)^{-1} \bB$\;
Repermute $\bW$ such that the number of non-zero entries in each row is in an increasing order (rows with equal number of non-zero entries are permuted at random) \label{alg:s_permute}\;
Repermute $\bA$ and $\bB$ according to the order derived in Step \ref{alg:s_permute}\;
\For{$k=1:p$}{
Find the possible parent set $\mathcal{P}_k$ using $\bW$\;
Using $\bA[k,:]$, verify that all parents of $x_k$ are included in $\mathcal{P}_k$, otherwise abort the algorithm since the model does not satisfy Assumption \ref{assumption:P-SCM_connection}\;
Initialize $\bar{\mathcal{P}}_k = \mathcal{P}_k$\;
Find the set $\mathcal{U}$ of possible parents in $\bar{\mathcal{P}}_k$ that have unique components \label{alg:s_uniquecomponent}\;
\eIf{$|\mathcal{U}| \neq 0$}{
	 Using $\bB[k,:]$; verify that $x_k$ is not connected to any unique components of $x_i$ for each $x_i\in\mathcal{U}$, otherwise abort the algorithm \label{alg:s_check1}\;
	% Remove the causal effect of $\tilde{s}_i$ from $x_k$:
	Update $\bar{\mathcal{P}}_k$ as $\bar{\mathcal{P}}_k \leftarrow \bar{\mathcal{P}}_k \setminus \mathcal{U}$;~ 
	% Remove the possible parents in $U$ from possible parent set:  $. 
	Go back to Step \ref{alg:s_uniquecomponent}, using the updated $\bar{\mathcal{P}}_k$ \label{alg:s_subtraction} \;
}{Set $\mathcal{I}_k = \bar{\mathcal{P}}_k$, i.e., $\mathcal{I}_k$ is the set of possible parents with no unique components after any number of iterations\;
Select the rows of $\bB$ that correspond to the possible parents in $\mathcal{I}_k$, denote these rows by $\bB_I$ \; 
Verify that the number of non-zero columns in $\bB_I$ is greater than or equal the number of its rows, otherwise abort the algorithm \label{alg:s_total_effect_nonunique} \;
Using $\bB[k,:]$, verify that $x_k$ is not connected to any sources corresponding to the non-zero columns of $\bB_I$, otherwise abort the algorithm \label{alg:s_check3} \;
}
}
\KwOut{If verification steps \ref{alg:s_check1}, \ref{alg:s_total_effect_nonunique}, \ref{alg:s_check3} yield true for all $k$, the model satisfies our conditions.}
\caption{Verifying satisfiability of Conditions \ref{condition:unique_component} and \ref{condition:marriage} for a linear P-SCM} 
\label{alg:satisfy} 
\end{algorithm}

From Appendix \ref{Matrix_representation}, the marriage condition can be equivalently applied to the set of possible parents with no unique components, i.e., $\mathcal{I}_k$. The marriage condition then translates to: For any
$X_C \subseteq \mathcal{I}_k$, $|X_C| \leq | \bigcup_{i:x_i\in X_C} {\rm{Comp}}(\tilde{s}_i) |$.
Note that, Algorithm \ref{alg:satisfy} only verifies the satisfiability of this condition for the subset $X_C = \mathcal{I}_k$. Thus, the condition verified in Algorithm \ref{alg:satisfy} is weaker than the marriage condition; yet it is much more efficient to verify. Indeed, we do not observe much difference in the results when we only verify the marriage condition for $X_C=\mathcal{I}_k$, as opposed to verifying the condition for all $X_C \subseteq \mathcal{I}_k$. 
% (i.e., $\mathbf{Z}_N$ being of full row rank, see Appendix \ref{Matrix_representation}), 
%the algorithm verifies that the source mixtures of the possible parents in $\mathcal{I}_k$ include at least $|\mathcal{I}_k|$ different source components; let us denote this as Condition 3. The marriage condition condition however is equivalent to Condition 3 when checked for all subsets of $\mathcal{I}_k$. Thus, Condition 3 is weaker than the marriage condition (i.e., is implied by the marriage condition),

%Much like the recovery algorithm (Algorithm \ref{alg:recovery}), the satisfiability algorithm begins with retrieving the possible parent set for each observed variable using the true mixing matrix, and it implements the same iterative structure to check for the unique components condition. For the marriage condition in this algorithm, we replace it by the overdeterminacy of the linear system in Step \ref{alg:total_effect_nonunique} of Algorithm \ref{alg:recovery}, which is weaker yet is more efficient. We do not observe many differences in the result after this replacement.

\begin{figure}[t]
  \centering
  \includegraphics[height=4.5cm]{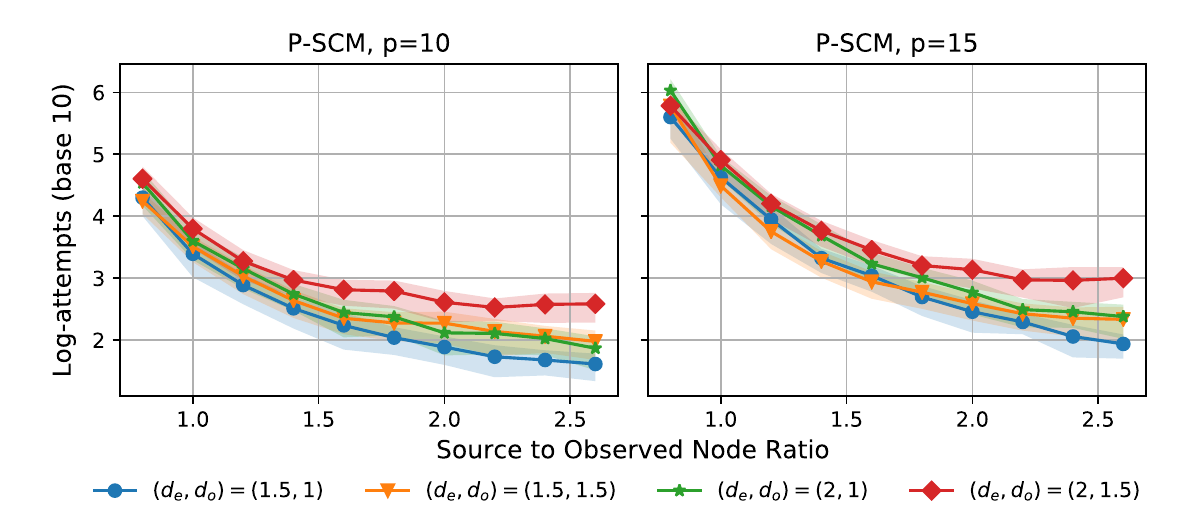} % [height=4.2cm]
  \caption{Satisfiability of our conditions for models generated according to a linear P-SCM. $d_e$ is the average observed node degree, and $d_o$ is the average source node degree.}
  \label{fig:satis1}
\end{figure}

\textbf{Satisfiability of P-SCMs. }
We first test the satisfiability of the conditions with different source to observed node ratios and different average degrees. We select $p=10, 15$ observed variables, average observed node degree $d_e= 1.5, 2$, and average source node degree $d_o= 1, 1.5$. Under each setting, we set the source to observed node ratio $r$ to be ranging from 0.8 to 2.6, with a step size of 0.2. We report the number of generated models (attempts) required to obtain a model that satisfies our conditions. We repeat the experiment 100 times, and calculate the average number of attempts (to find a satisfying model) and the standard deviations.
The common logarithms (base 10) of average \# attempts are shown in Figure \ref{fig:satis1}. The shaded area around each curve represents the logarithm of average \# attempts plus/minus half the standard deviation.

We observe that when there are less or equal number of sources than observed variables, our conditions are unlikely to be satisfied. This is due to the high probability of overlap among the exogenous connections, i.e., the same source is connected to multiple observed variables. This leads to a high probability of the model violating the unique component condition. When the ratio $r$ increases, the sources are less likely to be connected to multiple observed variables, hence the probability of satisfiability increases, converging to a fixed value. 

Now, let us compare the \# attempts for different average degrees within each subplot ($p=10$ and $p=15$). We observe when $r$ is large (greater than 2), increasing the observed node degree, $d_e$, is more likely to result in complex structures (among observed nodes). Further, increasing the source node degree, $d_o$, implies that more sources are shared among observed variables. Thus, increasing $d_e$ and $d_o$ increases the \# attempts required to obtain a satisfying model. When $r$ is small (less than 1), increasing $d_e$ again results in a complex structure. However, since the number of source variables is small, a small $d_o$ implies that observed variables are more likely to have either no exogenous connections, or a few that are shared among them. Subsequently, when $r$ is small, the average \# attempts required to generate a satisfying model is large for large $d_e$ and small $d_o$. This is shown by the green line in both subplots. We conclude that the satisfiability of our conditions depends on: 
\begin{enumerate}
\item The existence and similarity of exogenous connections among observed variables,
\item The ratio of overlap among exogenous connections (i.e., how many sources are shared among observed variables),
\item The structure of the causal graph among observed variables.
\end{enumerate}

\begin{figure}[t]
  \centering
  \includegraphics[height=4cm]{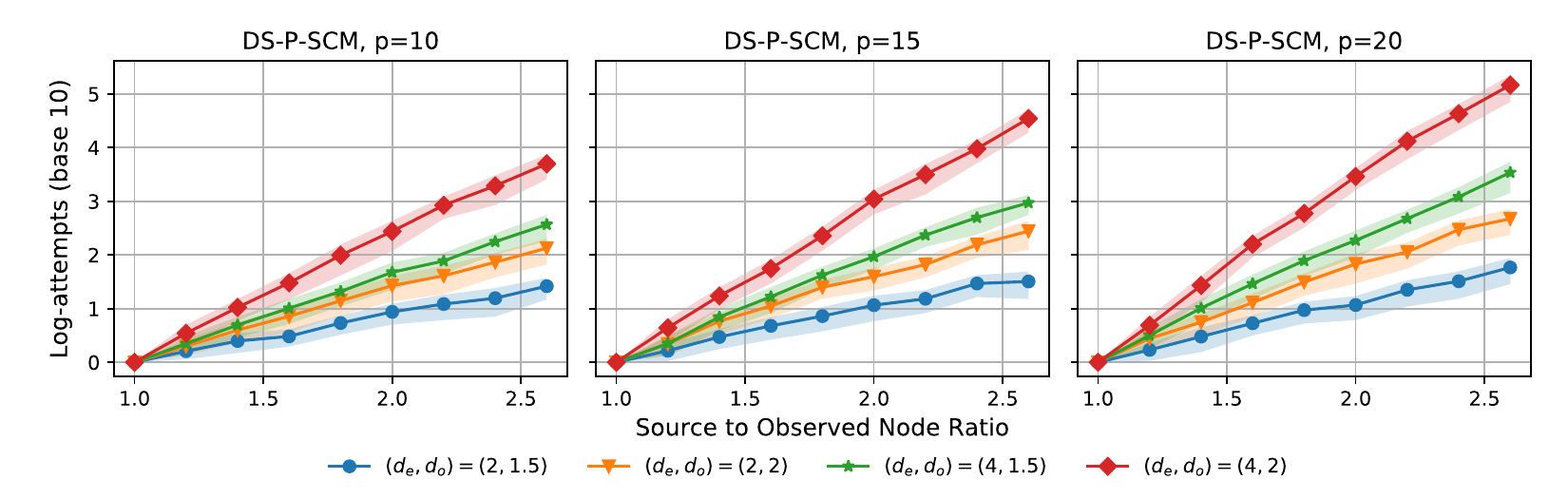}
  \caption{Satisfiability of our conditions for models generated according to a linear DS-P-SCM. 
  %The pair $(d_e,d_o)$ in the legend has the same meaning as in Figure \ref{fig:satis1}.
  }
  \label{fig:satis2}
\end{figure}

\textbf{Satisfiability of DS-P-SCMs.}
To further demonstrate our ideas, we test the satisfiability of the conditions when each observed variable in the generating model has a distinct source that is not shared with any other observed variable. As discussed in Section \ref{sec:comparison}, in this case, the model is a linear DS-P-SCM. We select $p=10, 15, 20$ observed nodes, $d_e= 2, 4$, and $d_o= 1.5, 2$. We report the average and the standard deviation of the \# attempts to generate 100 satisfying models. Figure \ref{fig:satis2} shows the logarithm of the \# attempts for the ratio $r$ ranging from 1 to 2.6. When $r=1$, the model can be exactly represented as a linear DS-P-SCM without latent confounding. Indeed, all models generated under this setting satisfy our conditions (satisfiability probability is equal to 1). Further, we observe that the \# attempts needed to obtain satisfying models increases as $r$ increases. Also, it is more likely for sparser graphs (small $d_e$ and $d_o$) to satisfy our conditions than denser graphs (large $d_e$ and/or $d_o$). These observations are aligned with our aforementioned conclusions.

\subsection{Performance of recovery}\label{app:simulation_recovery}
% \textbf{Implementation of baseline methods. }
% We compare the performance of P-SCM recovery algorithm with the following methods:
% \begin{itemize}
% \item LiNGAM: 
% \end{itemize}

\textbf{Uniquely identifiable cases. }
We test the performance of our P-SCM recovery algorithm (Algorithm \ref{alg:recovery}) for models generated according to each of the following settings:
\begin{enumerate}[(1)]
\item P-SCM with equal number of observed variables and sources, and $(d_e, d_o)=(1.5, 1.5)$. We refer to this setting as \texttt{P-SCM\_Equal}. 
\item P-SCM with $2$ less sources than observed variables, and $(d_e, d_o)=(1.5, 1.5)$. We refer to this setting as \texttt{P-SCM\_Fewer}. 

\item DS-P-SCM with $3$ more sources than observed variables, and $(d_e, d_o)=(2, 1.5)$. We refer to this setting as \texttt{DS-P-SCM}. 
\end{enumerate}

We select only the generated models that satisfy our conditions, and hence are uniquely identifiable. The sources are independent and identically distributed according to a uniform distribution over the interval $[-0.5, 0.5]$. We generate $n=1000$ samples of observations. For \texttt{P-SCM\_Equal} and \texttt{P-SCM\_Fewer}, we use FastICA \citep{hyvarinen1999fast} for blind source separation, and for \texttt{SCM}, we use ReconstructionICA \citep{le2011ica}.

We prune the mixing matrix using bootstrapping method \citep{efron1994introduction}. We first generate $n_{boot}=50$ bootstrap samples by sampling from the observed data. Then, for each bootstrap sample, we use ICA method to deduce estimates $\tilde{\bW}^{(i)}$, $i=1,\cdots,n_{boot}$ of the mixing matrix. Recall that each estimate suffers from permutation and scaling indeterminacies. For each estimate $\tilde{\bW}^{(i)}$, we normalize each column by the entry with the largest absolute value on that column. Further, we permute the columns of each $\tilde{\bW}^{(i)}$, $i=2,3,\cdots,n_{boot}$ such that the Frobenius distance between the permuted $\tilde{\bW}^{(i)}$ and $\tilde{\bW}^{(1)}$ is minimized. Lastly, we compute the average of the bootstrap estimates (samples) of the mixing matrix, and prune this average by applying a Student t-test element-wise in order to reject the entries with small variance around zero. In particular, we keep each entry with 95\% confidence level. The recovered mixing matrix $\tilde{\bW}$ is the resulting matrix after pruning the average of the bootstrap estimates. Notice that we assume that the number of sources is known for ICA recovery. 

We evaluate the performance of ICA recovery by comparing the recovered mixing matrix $\tilde{\bW}$ and the actual mixing matrix $\bW$, and call the ICA recovery successful when $\tilde{\bW}$ has the same non-zero entries (support) as $\mathbf{W}$ (after column permutations). That is, ICA learns the exact structure of $\mathbf{W}$ from observed data.

% divide the recovery of ICA into three cases: 1) $\tilde{\mathbf{W}}$ can learn the exact structure of $\mathbf{W}$, and the coefficients can be learned such that the Frobenius norm $||\tilde{\mathbf{W}} - \mathbf{W}||_F$ is controllable after permutation and scaling; 2) $\tilde{\mathbf{W}}$ can only learn the structure of $\mathbf{W}$, but the coefficients cannot be learned accurately enough; 3) $\tilde{\mathbf{W}}$ cannot learn the exact structure of $\tilde{\mathbf{W}}$. 

We evaluate the performance of our P-SCM Recovery algorithm using the recovered mixing matrix $\tilde{\bW} $ from ICA, and compare the performance with four other algorithms: ICA-LiNGAM\footnote{In this subsection, we refer to the LiNGAM algorithm proposed by \citep{shimizu2006linear} as ICA-LiNGAM, in order to distinguish this method from others. Note that the other three baseline methods do not use ICA.} \citep{shimizu2006linear}, DirectLiNGAM \citep{shimizu2011directlingam}, Pairwise lvLiNGAM \citep{entner2010discovering} and ParceLiNGAM \citep{tashiro2014parcelingam}, whenever applicable. These baseline methods (i) are all based on LiNGAM (or lvLiNGAM) and (ii) can recover the strengths of causal connections between observed variables, i.e., matrix $\bA$. We implement all baseline methods using the codes released by their corresponding authors, and select default hyperparameters throughout the simulation.

A well-known class of methods for causal structure learning in the presence of latent confounders is FCI algorithm \citep{spirtes2000causation} and its variants. However, we do not compare our approach with these methods since these methods returns a PAG, which is a class of Maximal Ancestral Graphs (MAGs), and our approach mostly focuses on unique identification of the ground truth (i.e., DAG). First of all, since our hypothesis space of possible graphical models (linear P-SCMs) is different from the class considered by FCI (DS-SCMs), the returned PAG may not contain the ground truth even if FCI does not make any mistakes in the estimation, and hence we do not believe that comparison is fair. Secondly, we note that in a MAG, two variables can be adjacent even if in the ground truth DAG they are not. Therefore, it is not straightforward to define a measure of performance between a MAG and the ground truth DAG. Third, we note that the size of the PAG returned by FCI can be exponentially large and hence, even if we have an evaluation method for each MAG, still computation time can make the comparison infeasible.

We replace the step of solving an overdetermined linear system (Step \ref{alg:total_effect_nonunique} in Algorithm \ref{alg:recovery}) by finding its least-squares approximation, in order to improve stability. We prune the recovered matrices $\bA$, by removing the edges with causal strength $|a_{ij}|< 0.1$ for all methods including ours. We compare the differences between the recovered adjacency matrix, and the true adjacency matrix $\bA$, using the following four test metrics: 
\begin{enumerate}[(1)]
    \item SHD / Edge; the normalized structural Hamming distance divided by the number of edges in the true model;
    \item Frobenius Norm;
    \item ``Precision'': The fraction of edges in the recovered model that appear in the true model;
    \item ``Recall'' (true positive rate): The fraction of edges in the true model that are correctly recovered.
\end{enumerate}
Note that for SHD / Edge and Frobenius Norm, lower values mean better performance. For Precision and Recall, higher values mean better performance.
% (1) normalized Structural Hamming Distance over the number of edges in the true model (SHD / Edge); (2) Frobenius norm; (3) ``Precision'', defined as the fraction of edges in the recovered model that appear in the true model; (4) ``Recall'', defined as the fraction of edges in the true model that are revealed (correctly recovered) in recovered model.

We also evaluate the performance of our P-SCM Recovery algorithm for recovering the matrix $\bB$, up to permutation and scaling indeterminacies, i.e., $\tilde{\bB}$.
%where the recovered exogenous connection matrix $\tilde{\bB}$ can only recover $\bB$ up to permutation and scaling indeterminacies. 
We normalize both $\bB$ and $\tilde{\bB}$ such that the entry with the maximum absolute value on each column is one. We then permute the columns of $\tilde{\bB}$ by minimizing the Frobenius distance:
\begin{align*}
    \tilde{\bB}^* = \argmin_{\mathbf{C}=\tilde{\bB}\mathbf{P}} ||\bB - \mathbf{C} ||_F,\quad \text{where } \mathbf{P}  \text{ is a permutation matrix.}
\end{align*}
We compare the difference between $\tilde{\bB}^*$ and $\bB$ using the same four metrics, and compare these results with the recovery of $\bA$. We use P-SCMR\_A and P-SCMR\_B to represent the recovery of matrices $\bA$ and $\bB$ by our algorithm.

\begin{figure}[t]
  \centering
  \includegraphics[width=\textwidth]{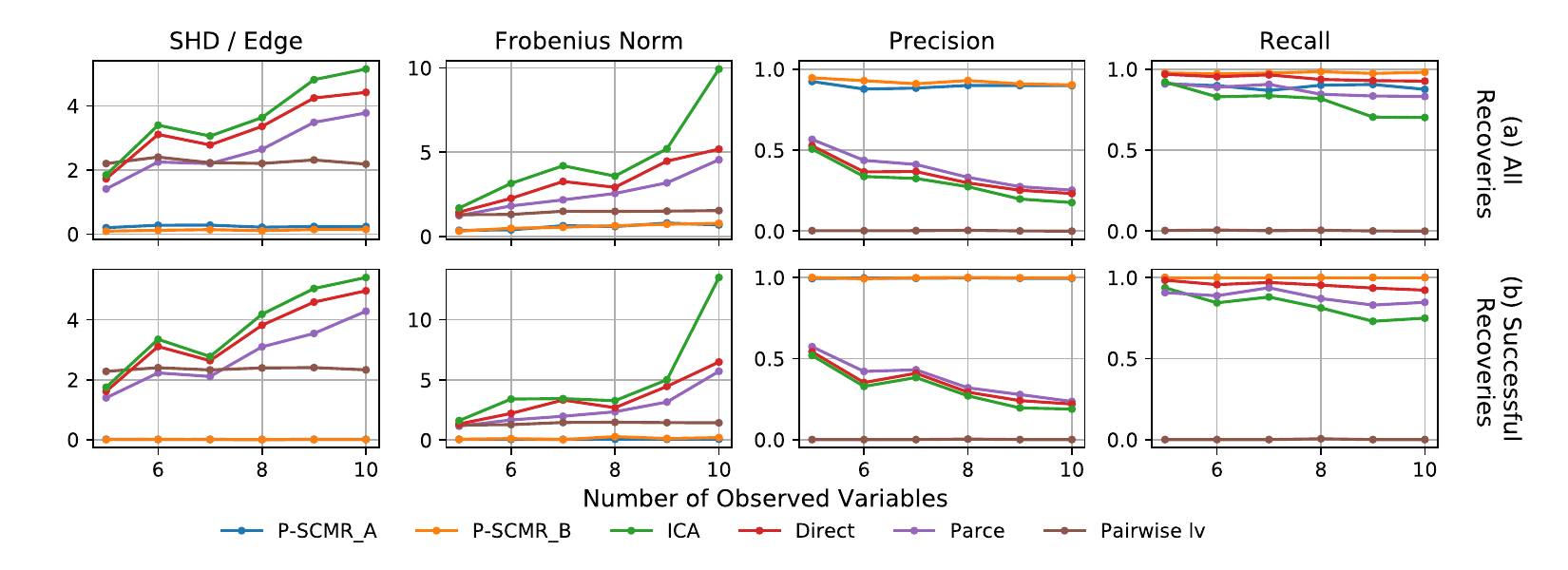}
  \caption{Performance of algorithms under \texttt{P-SCM\_Equal} setting. For the successful recoveries case, the P-SCMR\_A (blue line) overlaps with the P-SCMR\_B (orange line) in all four metrics.}
  \label{fig:recovery1}
\end{figure}

\begin{figure}[t]
  \centering
  \includegraphics[width=\textwidth]{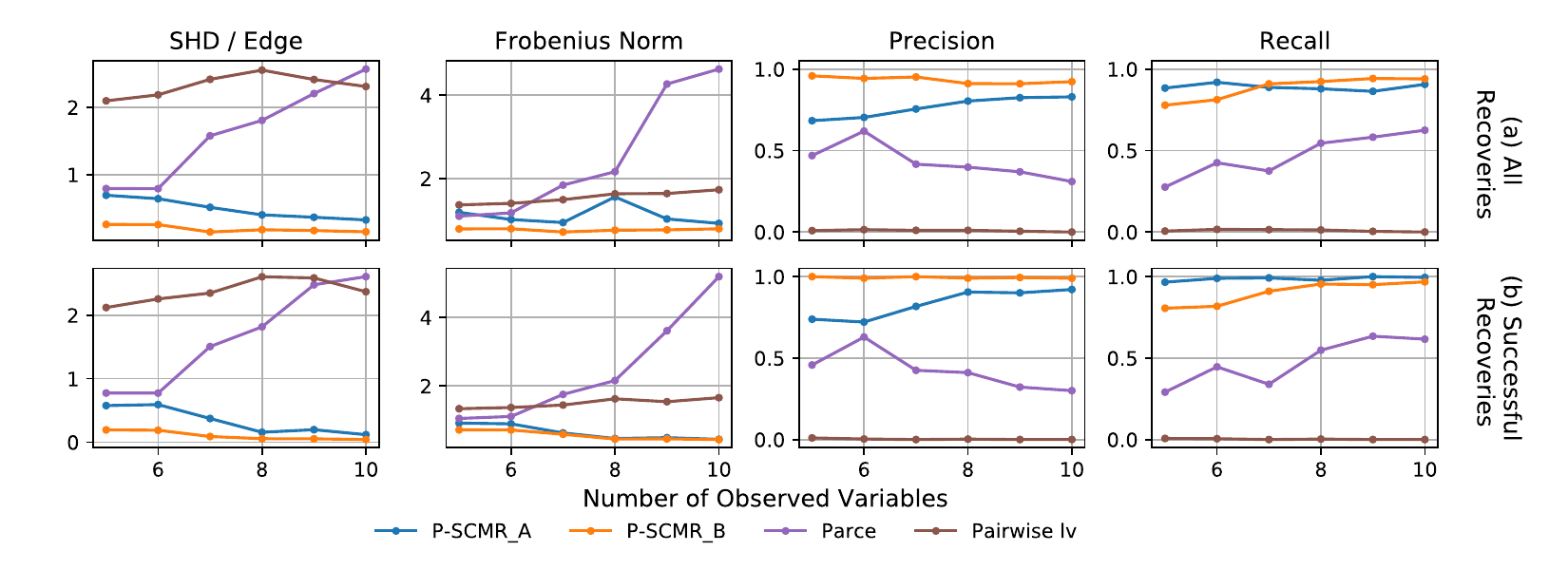}
  \caption{Performance of algorithms under \texttt{P-SCM\_Fewer} setting. ICA-LiNGAM and DirectLiNGAM cannot be applied to this case, since we have fewer sources than observed variables.}
  \label{fig:recovery2}
\end{figure}

\begin{figure}[t]
  \centering
  \includegraphics[width=\textwidth]{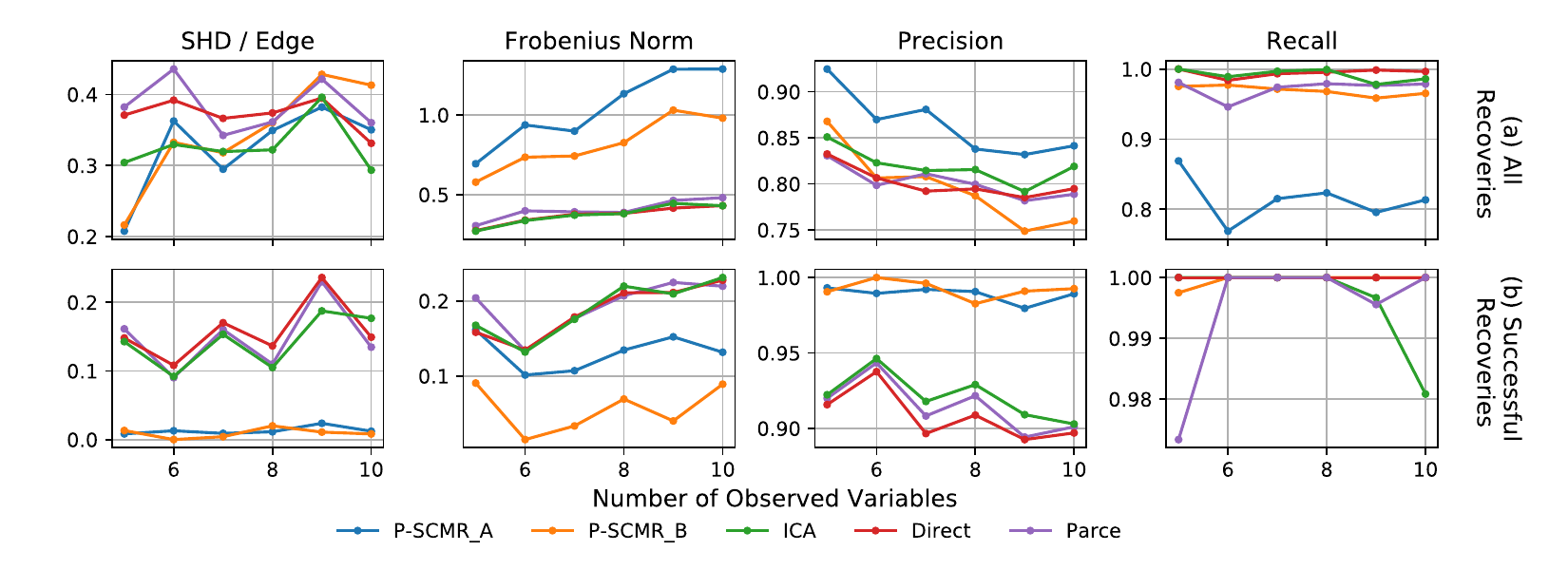}
  \caption{Performance of algorithms under \texttt{DS-P-SCM} setting. We ignore Pairwise lvLiNGAM since its performance is significantly worse than other methods, where it outputs empty graphs.}
  \label{fig:recovery3}
\end{figure}

We repeat the simulation until 50 models can be successfully recovered by ICA, and report the average of the aforementioned four metrics. Figures \ref{fig:recovery1} - \ref{fig:recovery3} show the performance of recoveries for all algorithms under Settings (1)-(3) for model generation, with the number of observed variables ranging from 5 to 10. We compare the performance for (i) all ICA recoveries, and (ii) only the successful recoveries by ICA. We observe that for \texttt{P-SCM\_Equal} and \texttt{P-SCM\_Fewer}, our algorithm can learn the model more accurately than the existing algorithms. This is because existing algorithms may misinterpret the shared sources among observed variables as confounding or direct causal connections. We observe from the plots for SHD/Edge and Precision that this misinterpretation may add additional edges to the recovered graph, which do not exist in the true model. For the third setting, \texttt{DS-P-SCM}, the performance of our method is comparable to existing algorithms. 

Besides, we observe that our algorithm performs significantly better than other algorithms when ICA recovery is successful, both in learning the structure of matrices $\bA$, $\bB$, and in learning the causal strengths (the coefficients). As discussed earlier, our algorithm assumes that the true mixing matrix $\bW$ can be correctly recovered up to permutation and scaling of its columns. Thus, our algorithm depends on the accuracy of BSS recovery. 
% We believe that the performance of our algorithm will be more accurate if better BSS methods are proposed in the future.

\begin{figure}[t]
  \centering
  \includegraphics[width=\textwidth]{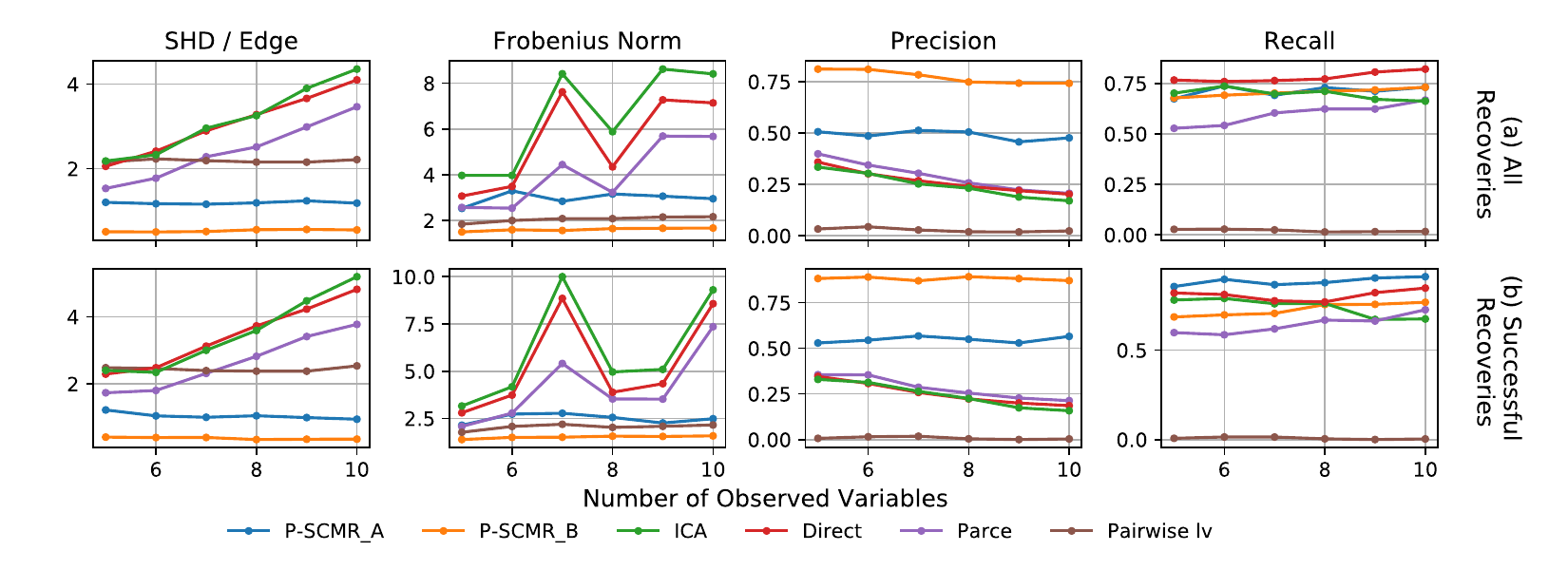}
  \caption{Performance of algorithms under \texttt{P-SCM\_NonUniq} setting.}
\label{fig:nonuniq}
\end{figure}

\textbf{Non-uniquely identifiable cases. }
We evaluate the performance of our algorithm when the generating model does not satisfy the necessary and sufficient conditions for unique identifiability, i.e., Conditions \ref{condition:unique_component} and \ref{condition:marriage}. Since the generating model cannot be uniquely recovered, there exist multiple generating models that correspond to the same mixing matrix $\bW$. However, since our conditions are imposed separately on every observed variable $x_k$, our algorithm can recover part of the generating model where the conditions are satisfied. Here, we consider \texttt{P-SCM\_Equal} setting (the number of observed variables is equal to the number of sources). We only select the generating models that do not satisfy our conditions, and refer to the modified setting as \texttt{P-SCM\_NonUniq}. Figure \ref{fig:nonuniq} shows the recovery performance for our algorithm, compared with the baseline algorithms. Not surprisingly, for \texttt{P-SCM\_NonUniq}, our algorithm does not perfectly recover the generating model, even when ICA recovery is successful (cf. Figure \ref{fig:nonuniq}). 
%This is due to the fundamental limitations developed in this work. 
However, our method outperforms the other algorithms in learning the structure and the causal strengths.

\subsection{Performance on real data}
We test the performance of our recovery algorithm on the daily closing prices of the following five world stock indices from 01/01/2015 to 06/30/2020: (1) Dow Jones Industrial Average (DJI) in USA, (2) Nikkei 225 (N225) in Japan, (3) Euronext 100 (N100) in Europe, (4) Hang Seng Index (HSI) in Hong Kong, and (5) the Shanghai Stock Exchange Composite Index (SSEC) in China. We only select the days where all five stock markets are active; we have $T=1192$ days for our observations. This data is obtained from Yahoo finance database\footnote{\url{https://finance.yahoo.com}}. The same setting, but with different data (different dates) has been considered in \citep[Section 5.2]{salehkaleybar2020learning}.

Let $c_i(t)$ be the closing price of the $i$-th index on day $t$. We define the corresponding return of an index $i\in\{\text{DJI, N225, N100, HSI, SSEC}\}$, at day $t$ as $R_{i}(t)=\left(c_{i}(t)-c_{i}(t-1)\right)/c_{i}(t-1)$, for all $t=2,3,\cdots,T$. In this setting, we consider the corresponding return of an index $i$ to be an observed variable. We suppose these observed variables were generated according to a linear P-SCM with five sources, where the sources are non-Gaussian random variables. 

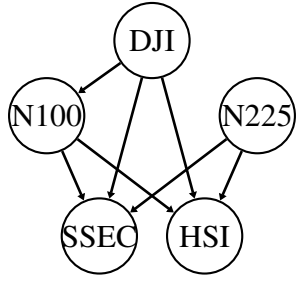
\begin{figure}[tbp]
\centering
\begin{tikzpicture}[thick, scale=0.4]
% \tikzstyle{every node}=[font=\small]
\foreach \place/\name in {{(-3.5,0)/N100}, {(3.5,0)/N225}, {(-1.75,-4)/SSEC}, {(1.75,-4)/HSI}, {(0,2.5)/DJI}}
    \node[real_appendix, label=center:\name] (\name) at \place {};
\foreach \source/\dest in {DJI/N100, DJI/SSEC, DJI/HSI, N100/SSEC, N100/HSI, N225/SSEC, N225/HSI}
    \path[causal] (\source) edge (\dest);
\end{tikzpicture}
  \caption{Recovered causal relations among five world stock indices using P-SCM Recovery algorithm.}
\label{fig:stock}
\end{figure}

We apply FastICA  with bootstrapping for source separation from observations. The recovered mixing matrix $\tilde{\bW}$ is
\begin{align}
\begin{bmatrix}
0.9096 & 0.2761 & 0 & 0 & 0 \\
0 & 0.7993 & 0 & 0.7414 & 0.2048 \\
0.4412 & 0.7738 & 0.1805 & -0.2962\ ~ & 0 \\
0.1537 & 0.4141 & 0.2902 & 0.1992 & 0.9398 \\
0.1480 & 0.2048 & 1.0000 & 0.4624 & 0.3513 \\
\end{bmatrix}
,
\label{eq:stock}
\end{align}
where the five rows correspond to DJI, N225, N100, HSI and SSEC respectively. Using P-SCM Recovery algorithm, the directed graph among these observed variables can be recovered as in Figure \ref{fig:stock}.
% the plot on the right in Figure \ref{fig:nips2021_exp1} in Section \ref{sec:simulations}.

For a directed acyclic graph (DAG), let the DAG-source node be the node with no parents, and the DAG-sink node be the node with no children. We observe from both $\tilde{\bW}$ and Figure \ref{fig:stock} that DJI is a DAG-source, with the fewest number of source components among all other indices. Further, we observe that DJI, N225 and N100 all have causal effects on HSI. Both observations are known to be true from common belief in economy, and from previous results in \citep{hyvarinen2010estimation,salehkaleybar2020learning}. One main difference in our result from the result in \citep{hyvarinen2010estimation} is that our algorithm cannot detect the directed edge from SSEC to HSI. This is because $\tilde{\bW}$ in \eqref{eq:stock} has the same components for HSI and SSEC, and our algorithm concludes that there are no causal relation between these two observed variables. This may happen when the exogenous connections for these two variables are fully overlapped, as in this example. %However, it might happen that one of these two observed variables have an exogenous source component which \it{cannot} be detected by BSS methods, such as an additional Gaussian exogenous noise. 

Recall that our linear P-SCM is advantageous when there are more sources than observed variables. However, in this example, we are assuming a number of sources equal to the number of observed variables. This is mainly so that we are able to use FastICA, which provides a highly accurate estimate for the mixing matrix (when a ground truth is available). The existing overcomplete ICA methods on the other hand are not as accurate.
%and our one shortage of our proposed algorithm is its sensitivity to noisy estimates of the mixing matrix.

\end{document}